\documentclass[11pt,reqno]{amsart}
\usepackage[left=1in,right=1in,top=1in,bottom=1in]{geometry}
\usepackage[section]{placeins}
\usepackage{float}
\usepackage{tcolorbox}
\usepackage{enumerate}
\usepackage{mathtools,amssymb}
\usepackage{subfig}
\usepackage{dcolumn}
\newcolumntype{d}[1]{D{.}{.}{#1}}

\usepackage[font=small]{caption}

\PassOptionsToPackage{hyphens}{url}\usepackage{hyperref}
\usepackage{xcolor}
\hypersetup{
    colorlinks,
    linkcolor={red!50!black},
    citecolor={blue!50!black},
    urlcolor={blue!80!black}
}

\DeclareMathOperator{\float}{fl}
\DeclareMathOperator{\im}{im}
\DeclareMathOperator{\rank}{rank}
\DeclareMathOperator{\nullity}{nullity}

\DeclareMathOperator{\VR}{VR}
\newcommand{\TC}{\omega}
\newcommand{\ccirc}{\mathbin{\mathchoice
{\xcirc\scriptstyle}
{\xcirc\scriptstyle}
{\xcirc\scriptscriptstyle}
{\xcirc\scriptscriptstyle}
}}
\newcommand{\xcirc}[1]{\vcenter{\hbox{$#1\circ$}}}
\usepackage{multirow}
\usepackage{tikz}
\usepackage{graphicx}

\newtheorem{thm}{Theorem}[section]

\newtheorem{prop}[thm]{Proposition}

\numberwithin{equation}{section}

\usepackage{subfig}
\usepackage[percent]{overpic}

\begin{document}
\graphicspath{{compressed_pictures/}}

\title{Topology of deep neural networks}
\author[G.~Naitzat]{Gregory Naitzat}
\address{Computational and Applied Mathematics Initiative, Department of Statistics,
University of Chicago, Chicago, IL 60637-1514.}
\email{gregn@galton.uchicago.edu, lekheng@uchicago.edu}
\author[A.~Zhitnikov]{Andrey Zhitnikov}
\address{Faculty of Electrical Engineering, Technion -- Israel Institute of Technology, Haifa 32000, Israel}
\email{andreyz@technion.ac.il}
\author[L.-H.~Lim]{Lek-Heng~Lim}

\begin{abstract}
We study how the topology of a data set $M = M_a \cup M_b \subseteq \mathbb{R}^d$, representing two classes of objects $a$ and $b$ in a binary classification problem, changes as it passes through the layers of a well-trained neural network, i.e., one with perfect accuracy on its training set and a near-zero generalization error ($\approx 0.01\%$). The goal is to shed light on two well-known mysteries in deep neural networks: (i) a nonsmooth activation function like ReLU outperforms a smooth one like hyperbolic tangent; (ii) successful neural network architectures rely on having many layers, despite the fact that a shallow network is able to approximate any function arbitrary well. We performed extensive experiments on the persistent homology of a wide range of point cloud data sets, both real and simulated. The results consistently demonstrate the following: (1) Neural networks operate by changing topology, transforming a topologically complicated data set into a topologically simple one as it passes through the layers. No matter how complicated the topology of $M$ we begin with, when passed through a well-trained neural network $f : \mathbb{R}^d \to \mathbb{R}^p$, there is invariably a vast reduction in the Betti numbers of both components $M_a$ and $M_b$; in fact they nearly always reduce to their lowest possible values: $\beta_k\bigl(f(M_i)\bigr) = 0$ for $k \ge 1$ and $\beta_0\bigl(f(M_i)\bigr) = 1$, $i =a, b$. Furthermore, (2)  the reduction in Betti numbers is significantly faster for ReLU activation compared to hyperbolic tangent activation as the former defines nonhomeomorphic maps that change topology, whereas the latter defines homeomorphic maps that preserve topology. Lastly, (3) shallow and deep networks transform the same data set somewhat differently --- a shallow network operates mainly through changing geometry and changes topology only in its final layers, a deep one spreads topological changes more evenly across all layers.
\end{abstract}


\maketitle

\section{Overview}

A key insight of topological data analysis is  that ``\emph{data has shape}'' \cite{focm,acta}. That data sets often have nontrivial topologies, which may be exploited in their analysis, is now a widely accepted principle with abundant examples across multiple disciplines: dynamical systems \cite{khasawneh2016chatter},  medicine \cite{li2015identification, nielson2015topological}, genomics \cite{perea2015sw1pers}, neuroscience \cite{giusti2015clique}, time series \cite{perea2015sliding}, etc. An early striking example came from computer vision, where  \cite{carlsson2008local} showed that naturally occurring image patches reside on a low-dimensional manifold that has the topology of a Klein bottle.

We will study how modern deep neural networks transform topologies of data sets, with the goal of shedding light on their breathtaking yet somewhat mysterious effectiveness.  Indeed, we seek to show that neural networks operate by changing the topology (i.e., shape) of data. The relative efficacy of ReLU activation over traditional sigmoidal activations can be explained by the different speeds with which they change topology --- a ReLU-activated neural network (which is not a homeomorphism) is able to sharply reduce Betti numbers but not a sigmoidal-activated one (which is a homeomorphism). Also, the higher the topological complexity of the data, the greater the depth of the network required to reduce it, explaining the need to have an adequate number of layers.

We would like to point out that the idea of changing the topology of space to facilitate a machine learning goal is not as esoteric as one might imagine. For example, it is implicit in kernel methods \cite{kernel} --- a data set with two components inseparable by a hyperplane is embedded in a higher-dimensional space where the embedded images of the components are separable by a hyperplane. Note that \emph{dimension} is a topological invariant, changing dimension is changing topology. We will see that a ReLU-activated neural network with many layers effects topological changes primarily through changing Betti numbers, another topological invariant.

Our study differs from current approaches in two important ways. Many existing studies either (i) analyze neural networks in an asymptotic or extreme regime, where the number of neurons in each layer or the number of layers becomes unbounded or infinite, leading to conclusions that really pertain to neural networks of somewhat unrealistic architectures; or (ii) they focus on what a neural network does to a single object, e.g., an image of a cat, and examine how that object changes as it passes through the layers. While we do not dispute the value of such approaches, we would like to contrast them with ours: We study what a neural network with a realistic architecture does to an entire class of objects. 
It is common to find expositions (especially in the mass media) of deep neural networks that purport to show their workings by showing how an image of a cat is transformed as it passes through the layers. We think this is misguided --- one should be looking at the entire manifold of cat images, not a single point on that manifold (i.e., a single cat image). This is the approach we undertake in our article.

\begin{figure}[hbt]
\fcolorbox{lightgray}{white}{\includegraphics[trim={0 0 00mm 0mm}, clip,  width=0.55\linewidth]{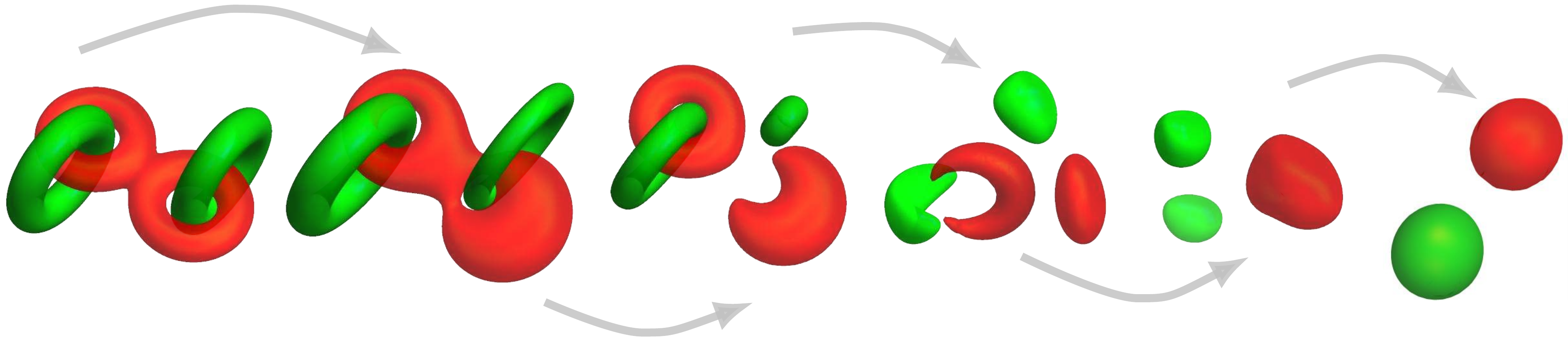}}\hspace{0.2cm}
\fcolorbox{lightgray}{white}{\includegraphics[width=0.39\linewidth]{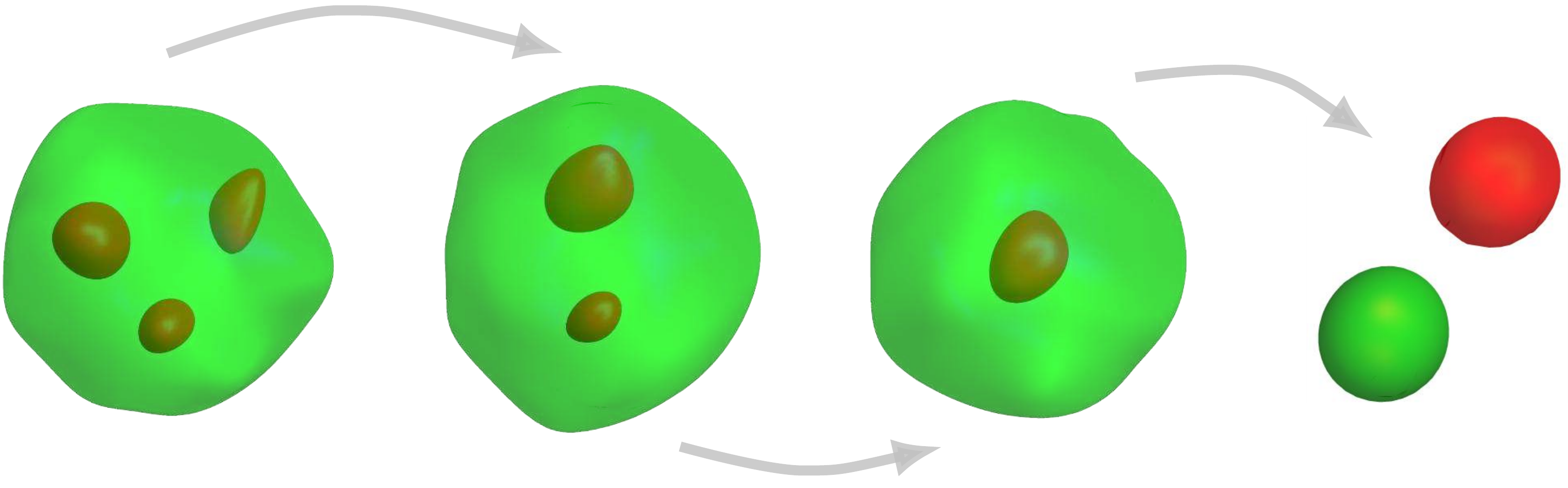}}
\caption{Progression of Betti numbers $\beta(X) = (\beta_0(X), \beta_1(X), \beta_2(X))$.
\emph{Left}: $\beta($red$)$: $(1, 2, 0) \to (1, 2, 0)\to (2, 1, 0)\to (2, 0, 0)\to (1, 0, 0)\to (1, 0, 0)$;
$\beta($green$)$: $(2, 2, 0)\to (2, 2, 0)\to (2, 1, 0)\to (2, 0, 0)\to (2, 0, 0)\to (1, 0, 0)$.
\emph{Right}: $\beta ($red$)$: $(3, 0, 0)\to (2, 0, 0)\to (1, 0, 0)\to (1, 0, 0)$;
$\beta($green$)$: $(1, 0, 3)\to (1, 0, 2)\to (1, 0, 1)\to (1, 0, 0)$.}
\label{fig:combined_plot_1c}
\end{figure}

Figure~\ref{fig:combined_plot_1c} illustrates what we mean by `changing topology'. The two subfigures are caricatures of real results (see Figures~\ref{fig:network_folding}, \ref{fig:2d_dataset},  \ref{fig:dataset_ii}, \ref{fig:dataset_iii}, for the true versions obtained via actual Betti numbers computations and projections to the principal components.) 

In both subfigures, we begin with a three-dimensional manifold $M = M_a \cup M_b$, comprising two disjoint submanifolds $M_a$ (green) and $M_b$ (red) entangled in a topologically nontrivial manner, and track its progressive transformation into a topologically simple manifold comprising a green ball and a red ball. In the left box, $M$ is initially the union of  the two green solid tori $M_a$,  interlocked with the red solid figure-eight $M_b$. In the right box, $M$ is initially a union of  $M_a$, the green solid ball with three voids inside, and $M_b$, three red balls each placed within one of the three voids of $M_a$. The topological simplification in both boxes are achieved via a reduction in the Betti numbers of both $M_a$ and $M_b$ so that eventually we have $\beta_k(M_i) = 0$ for $k \ge 1$ and $\beta_0(M_i) = 1$, $i =a,b$. Our main goal is to provide (what we hope is) incontrovertible evidence that this picture captures the actual workings of a  \emph{well-trained}\footnote{One with near zero generalization error.} neural network in a binary classification problem where $M_a$ and $M_b$ represent the two classes. 

In reality, the manifold $M = M_a \cup M_b$ would have to be replaced by a point cloud data set, i.e., a finite set of points sampled with noise from $M$. The notion of \emph{persistent homology} allows us to give meaning to the topology of point cloud data and estimate the Betti numbers of its underlying manifold.

\subsection{Key findings.} This work is primarily an empirical study --- we performed more than 10,000 experiments on real and simulated data sets of varying topological complexities and have made our codes available for reader's further experimentations.\footnote{\url{https://github.com/topnn/topnn_framework}.} We summarize our most salient observations and discuss their implications:
\begin{enumerate}[\upshape (i)]
\item  For a fixed data set and fixed network architecture, topological changes effected by a well-trained network are robust across different training instances and follow a similar profile.

\item\label{it:activation} Using smooth activations like hyperbolic tangent results in a slow down of topological simplification compared to nonsmooth activations like ReLU or Leaky ReLU.

\item\label{it:layer} The initial layers mostly induce only \emph{geometric} changes, it is  in  deeper layers that \emph{topological} changes take place. Moreover, as we reduce network depth,  the burden of producing topological simplification is not spread uniformly across layers but remains concentrated in the last layers. The last layers see a greater reduction in topological complexity than the initial layers.
\end{enumerate} 

Observation~\eqref{it:activation} provides a plausible answer to a widely asked question \cite{nair2010rectified, maas2013rectifier, glorot2011deep}: What makes rectified activations such as ReLU and its variants perform better than smooth sigmoidal activations? We posit that it is not a matter of smooth versus nonsmooth but that a neural network with sigmoid activation is a \emph{homeomorphic} map that preserves topology whereas one with ReLU activation is a \emph{nonhomeomorphic} map that can change topology.
It is much harder to change topology with homeomorphisms; in fact, mathematically it is impossible; but maps like the hyperbolic tangent achieve it in practice via rounding errors. Note that in IEEE finite-precision arithmetic, the hyperbolic tangent is effectively a piecewise linear step function:
\[
\tanh_\delta (x) = 
\begin{cases}
+1   & \text{if }  \float(\tanh(x)) > 1 - \delta,\\
\float(\tanh(x)) & \text{if }  -1 + \delta \le \float(\tanh(x)) \le 1 - \delta,\\
-1   & \text{if }  \float(\tanh(x)) < -1 + \delta, 
\end{cases}
\]
where $\float(x)$ denotes floating point representation of $x \in \mathbb{R}$, and $\delta > 0$ is the \emph{unit roundoff}, i.e., $\delta = \epsilon/2$ with $\epsilon = \inf \{ x > 0 :  \float(1 + x) \ne 1\}$ the \emph{machine epsilon} \cite{Overton}. Applied coordinatewise to a vector,
$\tanh : \mathbb{R}^n \to (-1,1)^n$  is a homeomorphism of $\mathbb{R}^n$ to $(-1,1)^n$ and necessarily preserves topology; but $\tanh_\delta : \mathbb{R}^n \to [-1,1]^n$ is not a homeomorphism and thus has the ability to change topology. We also observe that lowering the floating point precision increases the value of $\delta$ (e.g., for double precision $\delta = 2^{-54}$, for half precision\footnote{Assuming the {\scriptsize\textsf{BFloat16}}   floating-point format used in \textsf{TensorFlow}.} $\delta = 2^{-9}$), which has the effect of coarsening $\tanh_\delta$, making it even further from a homeomorphism and thus more effective at changing topology. We suspect that this may account for the paradoxical superior performance of lower precision arithmetic in deep neural networks \cite{courbariaux2014training, gupta2015deep, hubara2017quantized}.

\begin{figure}[h]
\centering
\fcolorbox{lightgray}{white}{\includegraphics[trim={0 2mm 00mm 0mm}, clip, width=1\linewidth]{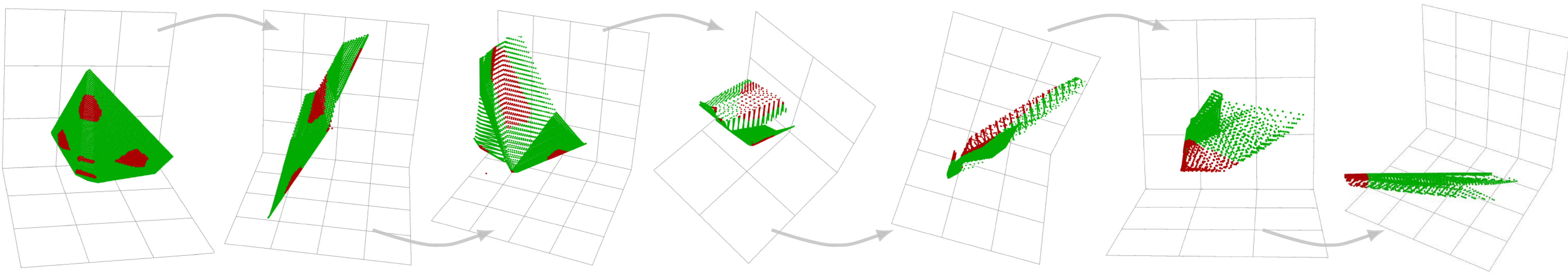}}
\caption{We see how the data set is transformed after passing through layers $2,3,\dots,8$ of a ReLU network with three neurons in each layer, well-trained to detect five disks in a square. $\beta($red$)$: $(5, 0)\to (4, 0)\to (4, 0)\to (4, 0)\to (2, 0)\to (1,0)\to (1, 0)$.}
\label{fig:network_folding}
\end{figure}

The ReLU activation, on the other hand, is far from a homeomorphism (for starters, it is not injective)  even in exact arithmetic. Indeed, if changing topology is the goal, then a composition of an affine map with ReLU activation, $\nu : \mathbb{R}^n \to \mathbb{R}^n$, $x \mapsto \max(Ax +b, 0)$, is a quintessential tool for achieving it --- any topologically complicated part of $M \subseteq \mathbb{R}^n$ can be affinely moved outside the nonnegative orthant and collapsed to a single point by the rectifier. We see this in action in Figure~\ref{fig:network_folding}, which unlike Figure~\ref{fig:combined_plot_1c}, is a genuine example of a  ReLU neural network trained to perfect accuracy on a two-dimensional manifold data set, where $M_a$ comprises five red disks in a square $M$ and $M_b = M \setminus M_a$ is the remaining green portion with the five disks removed. The `folding' transformations in Figure~\ref{fig:network_folding} clearly require  many-to-one maps and can never be achieved by any homeomorphism.

The effectiveness of ReLU activation over sigmoidal activations is often attributed to the former's avoidance of the vanishing/exploding gradient problem. Our results in Section~\ref{sec:results} indicate that this does not give the full explanation. Leaky ReLU and ReLU both avoid vanishing/exploding gradients, yet they transform data sets in markedly different manners --- for one, ReLU reduces topology faster than Leaky ReLU. The sharpness of the gradients is clearly not what matters most; on the other hand, the topological perspective perfectly explains why.

Observation~\eqref{it:layer} addresses another perennial paradox \cite{krizhevsky2012imagenet, eigen2013understanding, seide2011conversational}: Why does a neural network with more layers work better, despite the well-known universal approximation property that any function can be approximated arbitrarily well by a two-layer one? We posit that the traditional approximation-theoretic view of neural networks is insufficient here; instead, the proper perspective is that of a topologically complicated input getting progressively simplified as it passes through the layers of a neural network. Observation~\eqref{it:layer} accounts for the role of the additional layers ---  topological changes are minor in the first few layers and occur mainly in the later layers, thus a complicated data set requires many more layers to simplify.

We emphasize that our goal is to explain the mechanics of what happens from one layer to the next, and to see what role each attribute of the network's architecture --- depth, width, activation --- serves in changing topology. Note that we are not merely stating that a neural network is a blackbox that collapses each class to a component but how that is achieved, i.e., what goes on inside the blackbox.

\subsection{Relations with and departures from prior works.} 
As in topological data analysis, we make use of persistent homology and quantify topological complexity in terms of  Betti numbers; we track how these numbers change as a point cloud data set passes through the layers of a neural network.  But that is the full extent of any similarity with topological data analysis. In fact, from our perspective, topological data analysis and neural networks have opposite goals --- the former is largely concerned with \emph{reading} the shape of data, whereas the latter is about \emph{writing} the shape of data; not unlike the relation between computer vision and computer graphics, wherein one is interested the inverse problems of the other. Incidentally, this shows that a well-trained neural network applied in reverse can be used as a tool for labeling components of a complex data set and their interrelation, serving a role similar to \emph{mapper} \cite{SinghMC07} in topological data analysis. This idea has been explored in \cite{Pearson13, naitzat2018m}.

To the best of our knowledge, our approach towards elucidating the inner workings of a neural network by studying how the topology, as quantified by persistent Betti numbers, of a point cloud data set changes as it passes through the layers has never been done before. The key conclusion of these studies, namely, that the role of a neural network is primarily as a topology-changing map, is also novel as far as we know. 
Nevertheless, we would like to acknowledge a Google Brain blog post  \cite{Colah} that inspired our work --- it speculated on how neural networks may act as homeomorphisms that distort \emph{geometry}, but stopped short of making the leap to topology-changing maps.

There are other works that employ Betti numbers in the analysis of neural networks. \cite{bianchini2014complexity} did a purely theoretical study of upper bounds on the topological complexity (i.e., sum of Betti numbers) of the decision boundaries of neural networks with smooth sigmoidal activations; \cite{pmlr-v97-ramamurthy19a} did a similar study with a different measure of topological complexity. \cite{Guss2018} studied the empirical relation between the topological complexity of a data set and the minimal network capacity required to classify it. \cite{Rieck2019} used persistent homology to monitor changes in the weights of neural network during training and proposed an early stopping criteria based on persistent homology.

\subsection{Outline.}  In Section~\ref{sec:tools1} we introduce, in an informal way, the main topological concepts used throughout this article. This is supplemented by a more careful and detailed treatment in Section~\ref{sec:background}, which provides a self-contained exposition of simplicial homology and persistent homology tailored to our needs. Section~\ref{sec:problem} contains a precise formulation of the problem  we study, specifies what is tracked empirically, and addresses some caveats.  Section~\ref{sec:methodology} introduces our methodology for tracking topological changes and  implementation details. We present the results from our empirical studies with discussions in Section~\ref{sec:results}, verified our findings on real-world data in Section~\ref{sec:real_data}, and conclude with some speculative discussions in Section~\ref{sec:final_remarks}.

\section{Quantifying topology}\label{sec:tools1}

\begin{figure}[h]
\centering
\begin{minipage}[h]{0.45\textwidth}
\includegraphics[width=0.9\textwidth]{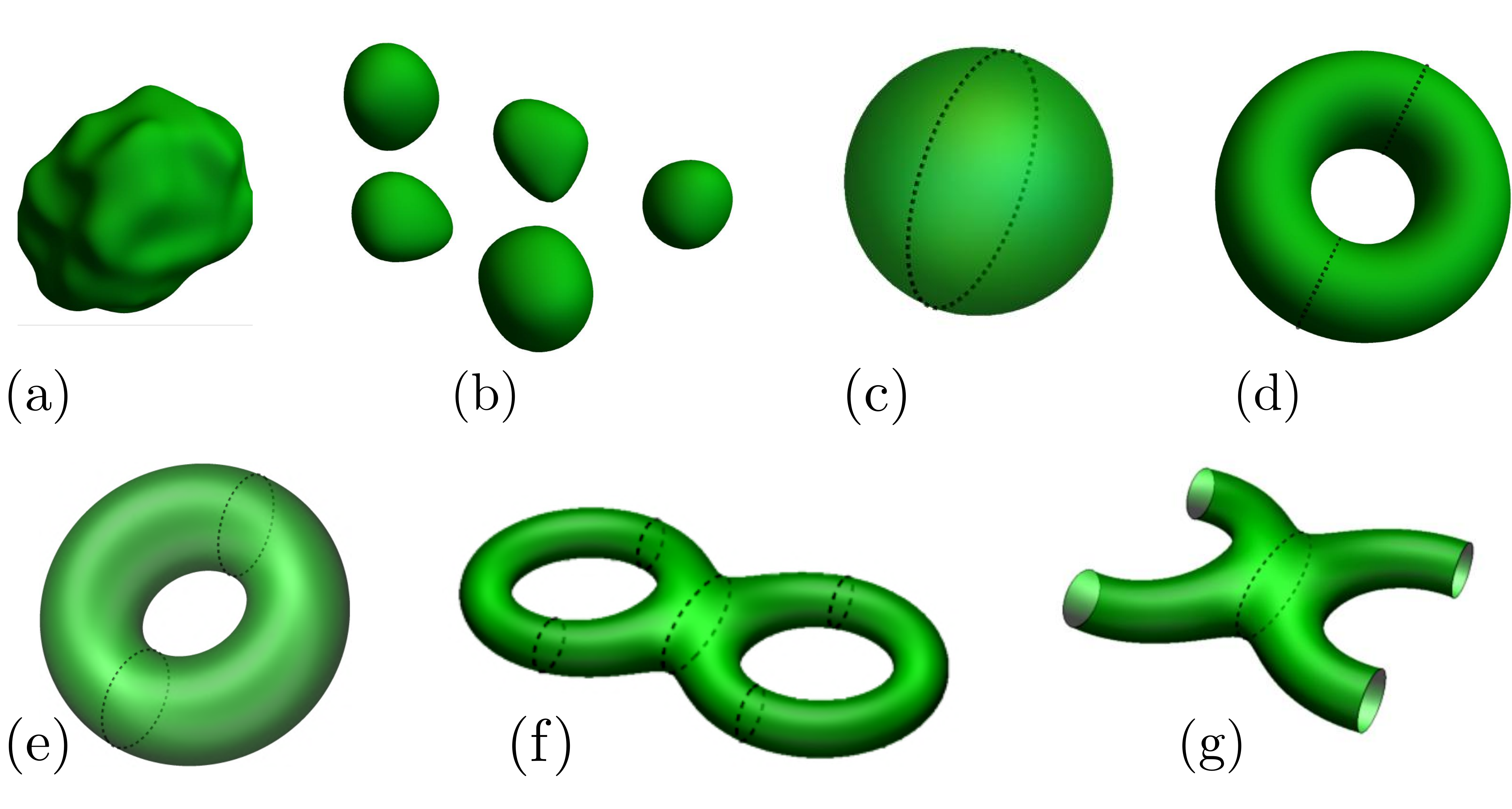}
\end{minipage}\hspace{2.5cm}\begin{minipage}[c]{0.3\textwidth}
\hspace{-15ex}
\centering
\footnotesize
\begin{tabular}[t]{llc}
\hline
\hline
& Manifold $M\subseteq \mathbb{R}^3$ & $\beta(M)$ \\
\hline
(a) & Single contractible manifold  & $(1, 0, 0)$\\
(b) & Five contractible manifolds  & $(5, 0, 0)$ \\
(c) & Sphere  & $(1,0,1)$\\
(d) & Solid torus (filled)  & {\color{black}$(1, 1, 0)$}\\
(e) & Surface of torus (hollow) & $(1, 2, 1)$ \\
(f) & Genus two surface (hollow)  &  {\color{black}$(1, 4, 1)$} \\
(g) & Torso surface (hollow) & $(1, 3, 0)$ \\
\hline
\hline
\end{tabular}
\end{minipage}
\caption{Manifolds in $\mathbb{R}^3$ and their Betti numbers.}
\label{fig:manifolds}
\end{figure}

In this article, we rely entirely on \emph{Betti numbers} $\beta_k(M)$ to quantify topology as they are the simplest topological invariants that capture the shape of a space $M \subseteq \mathbb{R}^d$, have intuitive interpretations, and are readily computable within the framework of persistent homology for a point cloud data set sampled from $M$. The zeroth Betti number, $\beta_0(M)$, counts the number of connected components in $M$; the $k$th Betti number, $\beta_k(M)$, $k \ge 1$, is informally the number of $k$-dimensional holes in $M$. In particular, $\beta_k(M)=0$ when $k > d$ as there is no $(d+1)$-dimensional holes in $d$-dimensional space. So for $M \subseteq \mathbb{R}^d$, we write $\beta(M) \coloneqq (\beta_0(M), \beta_1(M), \dots, \beta_d(M))$ --- these numbers capture the shape or topology of $M$, as  one can surmise from Figure~\ref{fig:manifolds}. So whenever we refer to `topology' in this article, we implicitly mean $\beta(M)$.

If $M$ has no holes and can be continuously (i.e., without tearing) deformed to a point, then $\beta_0(M) = 1$ and $\beta_k(M) = 0$ for all $k \ge 1$; such a space is called \emph{contractible}.  The simplest noncontractible space is a circle  $S^1 \subseteq \mathbb{R}^2$, which has a single connected component and a single one-dimensional hole, so $\beta_0 (S^1)= 1 = \beta_1(S^1)$ and $\beta_k(S^1) = 0$ for all $k \ge 2$. Figure~\ref{fig:manifolds} has a few more examples. 

Intuitively, the more holes a space has, the more complex its topology. In other words, the larger the numbers in $\beta(M) $, the more complicated the topology of $M$. 
As such, we define its \emph{topological complexity} by
\begin{equation}\label{eq:TC}
\TC(M) \coloneqq \beta_0(M) + \beta_1(M) + \dots + \beta_d(M).
\end{equation}
While not as well-known as the \emph{Euler characteristic} (which is an alternating signed sum of the Betti numbers), the topological complexity is also a classical notion in topology, appearing most notably in Morse theory  \cite{Milnor}; one of its best known result is that the topological complexity of $M$ gives a lower bound for the number of stationary points of a function $f : M \to \mathbb{R}$ with nondegenerate Hessians.
It also appears in many other contexts \cite{milnor1964betti, basu2018multi}, including neural networks. We highlight in particular the work of \cite{bianchini2014complexity} that we mentioned earlier, which studies the topological complexity of the decision boundary of neural networks with activations that are Pfaffian functions \cite{TZell, gabrielov2004betti}. These include sigmoidal activations but not the ReLU nor leaky  ReLU activations studied in this article. For piecewise linear activations like ReLU and leaky ReLU, the most appropriate theoretical upper bounds for topological complexity of decision boundaries are likely given by the number of linear regions \cite{Montufar,tropical18}.

The goal of our article is different, we are interested not in the shape of the decision boundary of  an $l$-layer neural network $\nu_l : \mathbb{R}^d \to \mathbb{R}^p$  but in the shapes of the input $M \subseteq \mathbb{R}^d$, output  $\nu_l(M) \subseteq \mathbb{R}^q$, and all its intermediate layers $\nu_k(M)$, $k =1,\dots,l-1$. By so doing, we may observe how the shape of $M$ is transformed as it passes through the layers of a well-trained neural network, thereby elucidating its workings.  In other  words, we would like to track the Betti numbers
\[
\beta(M) \to \beta(\nu_1(M)) \to  \beta(\nu_2(M)) \to \dots \to  \beta(\nu_{l-1}(M)) \to  \beta(\nu_l(M)).
\]
To do this in reality, we will have to estimate $\beta(M)$ from a point cloud data set, i.e., a finite set of points $T \subseteq M$ sampled from $M$, possibly with noise. The next section will describe the procedure to do this via persistent homology, which is by now a standard tool in topological data analysis. Readers who do not want to be bothered with the details just need to know that one may reliably estimate $\beta(M)$ by sampling points from $M$; those who like to know the details may consult the next section. The main idea is that the Betti numbers of $M$ may be estimated by constructing a simplicial complex from $T$ in one of several ways that depend on a `persistent parameter', and then using simplicial homology to compute the Betti numbers of this simplicial complex. Roughly speaking, the `persistent parameter' allows one to pick the right scale at which the point cloud $T$ should be sampled so as to give a faithful estimation of $\beta(M)$. Henceforth whenever we speak of $\beta(M)$, we mean the Betti numbers estimated in this fashion.

For simplicity of the preceding discussion, we have used $M$ as a placeholder for any manifold. Take say a handwritten digits classification problem (see Section~\ref{sec:real_data}), then $M = M_0 \cup M_1 \cup \dots \cup M_9$ has ten components, with $M_i$ the manifold of all possible handwritten digits $i \in \{0,1,\dots,9\}$. Here we are not interested in $\beta(M)$ per se but in  $\beta(\nu_k(M_i))$ for all $k =0,1,\dots, l$ and $i=0,1,\dots,9$ --- so that we may see how each component is transformed as $M$ passes through the layers, i.e., we will need to sample points from each of $\nu_k(M_i)$ to estimate its Betti numbers, for each component $i$ and at each layer $k$.

\section{Algebraic topology and persistent homology background}\label{sec:background}

This section may be skipped by readers who are already familiar with persistent homology or are willing to take on faith what we wrote in the last two paragraphs of the last section. Here we will introduce background knowledge in algebraic topology --- simplicial complex, homology, simplicial homology --- and provide a brief exposition on selected aspects of topological data analysis --- Vietoris--Rips complex, persistent homology, practical homology computations --- that we need for our purposes. 

\subsection{Simplicial complexes}\label{sec:simp_comp}

A $k$-dimensional \emph{simplex}, or $k$-simplex, $\sigma$ in $\mathbb{R}^d$, is the convex hull of $k + 1$  affinely independent points $v_0,\dots,v_k \in \mathbb{R}^d$. A $0$-simplex is a point, a $1$-simplex is a line segment, a $2$-simplex is a triangle, and a $3$-simplex is a tetrahedron. A $k$-simplex is represented by listing the set of its $k+1$ \emph{vertices} and denoted $\sigma = [ v_0, \dots, v_{k}]$. The \emph{faces} of a $k$-simplex are simplices of dimensions $0$ to $k-1$ formed by convex hulls of proper subsets of  its \emph{vertex set} $\{v_0, \dots, v_{k}\}$. For example, the faces of  a line segment/$1$-simplex are its end points, which  are $0$-simplices; the faces of a triangle/$2$-simplex are its three sides, which are $1$-simplices, and its three vertices, which are $0$-simplices.

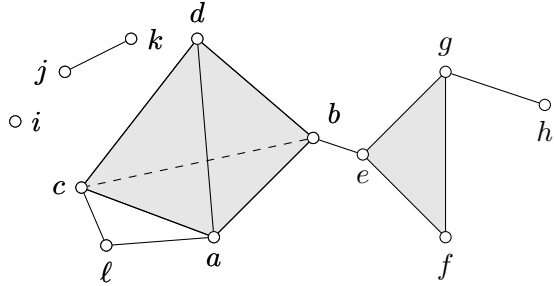
\begin{figure}[!h]
\centering
\begin{tikzpicture}[scale=2.2, vertices/.style={draw, fill=white,  shape =circle,circle, inner sep=1.5pt}]

\filldraw[fill=black!10, draw=black] (-0.5,0.5)--(0,0)--(0,1)--cycle;
\node[vertices, label=below:{$e$}] (a) at (-0.5,0.5) {};
\node[vertices, label=below:{$f$}] (b) at (0,0) {};
\node[vertices, label=above:{$g$}] (c) at (0,1) {};
\node[vertices, label=below:{$h$}] (d) at (0.6,0.8) {};

\draw (c)--(d);
\begin{scope}[shift={(-1.4,0)}]
\node[vertices, label=below:{$a$}] (a2) at (0,0) {};

\node[vertices, label=below:{$\ell$}] (e2) at (-0.65,-0.05) {};

\node[vertices, label=above right:{$b$}] (b2) at (0.6,0.6) {};
\node[vertices, label=left:{$c$}] (c2) at (-0.8,0.3) {};
\node[vertices, label=above:{$d$}] (d2) at (-0.1,1.2) {};

\node[vertices, label=right:{$i$}] (e) at (-1.2,0.7) {};

\node[vertices, label=left:{$j$}] (f) at (-0.9,1) {};
\node[vertices, label=right:{$k$}] (g) at (-0.5,1.2) {};
\draw (g)--(f);

\filldraw[fill=black!10, draw=black] (0,0)--(0.6,0.6)--(-0.1,1.2) --(-0.8,0.3)--cycle;

\filldraw[fill=black!10, draw=black] (a2)--(e2)--(c2) --cycle;

\draw (b2)--(a);

\foreach \to/\from in {a2/b2,a2/c2,a2/d2,b2/d2,c2/d2}
\draw [-] (\to)--(\from);
\draw [dashed] (b2)--(c2);

\node[vertices, label=below:{$a$}] (a2) at (0,0) {};

\node[vertices, label=below:{$\ell$}] (e2) at (-0.65,-0.05) {};

\node[vertices, label=above right:{$b$}] (b2) at (0.6,0.6) {};
\node[vertices, label=left:{$c$}] (c2) at (-0.8,0.3) {};
\node[vertices, label=above:{$d$}] (d2) at (-0.1,1.2) {};

\node[vertices, label=right:{$i$}] (e) at (-1.2,0.7) {};

\node[vertices, label=left:{$j$}] (f) at (-0.9,1) {};
\node[vertices, label=right:{$k$}] (g) at (-0.5,1.2) {};
\end{scope}
\end{tikzpicture}
\caption{A \emph{geometrical simplicial complex} in $\mathbb{R}^3$ that is a geometrical realization of an \emph{abstract simplicial complex} $K = \{ [a,b,c,d], [e,f,g], \dots, [e, b], \dots,  [a], [b], \dots, [\ell]\}$ comprising $32$ simplices: a single $3$-simplex $[a,b,c,d]$, five $2$-simplices such as $[a,c,d]$ and $[e,f,g]$, eighteen $1$-simplices such as $[e, b]$ and $[g,h]$, fourteen $0$-simplices $[a], [b], \dots, [\ell]$. Note that in the geometrical simplicial complex, the simplices intersect along faces.}
\label{fig:2}
\end{figure}

An $m$-dimensional \emph{geometrical simplicial complex} $K$ in $\mathbb{R}^d$ is a finite collection of simplices in $\mathbb{R}^d$ of dimensions at most $m$ that are (i) glued together along faces, i.e., any intersection between two simplices in $K$ is necessary a face of both of them; and (ii) include all faces of all its simplices, e.g., if the simplex $\sigma_1 = [v_0,v_1, v_2]$ is in $K$, then the simplices $[v_0, v_1], [v_1,v_2],[v_0, v_2], [v_0],[v_1],[v_2]$ must all also belong to $K$. Behind each geometrical simplicial complex is an \emph{abstract simplicial complex} --- a list of simplices $K =\{ \sigma_1, \sigma_2, \dots, \sigma_n \}$ with the property that if $\tau \subseteq \sigma \in K$, then $\tau \in K$. This combinatorial description of an abstract simplicial complex is exactly  how we describe a graph, i.e., $1$-dimensional simplicial complex, as an abstract collection of edges, i.e., $1$-simplices, comprising pairs of vertices. Conversely, any abstract simplicial complex can be realized geometrically as a geometrical simplicial complex in $\mathbb{R}^d$ like in Figure~\ref{fig:2}, an example of a $3$-dimensional simplicial complex in $\mathbb{R}^3$.  The abstract description of a simplicial complex allows us to treat its simplices as elements in a vector space, a key to define simplicial homology, as we will see in Section~\ref{sec:simp_hom}.

\subsection{Homology and Betti numbers}\label{sec:betti_homology}

Homology  is an abstract way to encode the topology of a space by means of a chain of vector spaces and linear maps. We refer readers to \cite{hodge} for an elementary treatment requiring nothing more than linear algebra and graph theory. Here we will give an even simpler treatment restricted to  $\mathbb{F}_2 =\{0,1\}$, the field of two elements with arithmetic performed modulo $2$, which is enough for this article.

Let $C_0, C_1,\dots, C_d$ be vector spaces over $\mathbb{F}_2$. Let  $\partial_k : C_k \to C_{k-1}$ be linear maps called \emph{boundary operators} that satisfy the condition that ``a boundary of a boundary is trivial,''  i.e.,
\begin{equation}\label{eq:bound}
\partial_k \ccirc \partial_{k+1} = 0
\end{equation} 
for all $k=0, \dots, d$. A \emph{chain complex} refers to the sequence
\[
0 \xrightarrow{\partial_{d+1} } C_d 
\xrightarrow{\partial_{d} } C_{d-1}
\xrightarrow{\partial_{d-1}} \cdots 
\xrightarrow{\partial_{k+1}} 
C_{k}
\xrightarrow{\partial_{k}}  
C_{k-1}
\xrightarrow{\partial_{k-1}} 
\cdots 
\xrightarrow{\partial_2 } 
C_{1}
\xrightarrow{\partial_1 } 
C_{0}
\xrightarrow{\partial_0} 0,
\]
where we set $C_{d+1} = C_{-1} = 0$, the trivial subspace. The elements in the image of $\partial_{k}$ are called \emph{boundaries} and elements in the kernel of $\partial_{k-1}$ are called \emph{cycles}. Clearly $\ker(\partial_k)$ and $\im(\partial_{k+1})$ are both subspaces of  $C_k$ and by \eqref{eq:bound},
\[
B_k \coloneqq \im(\partial_{k+1}) \subseteq  \ker(\partial_{k}) \eqqcolon Z_k.
\]
We may form the quotient vector space
\[
H_k \coloneqq  Z_k/ B_k = \ker(\partial_k)/ \im(\partial_{k+1}), \qquad k=0,1, \dots, d,
\]
and we will call it the $k$th \emph{homology group} --- the `group' here refers to the structure of $H_k$ as an abelian group under addition. The elements of $H_k$ are called \emph{homology classes}; note that these are cosets or equivalence classes of the form
\begin{equation}\label{eq:homcls}
[z] = z + B_k = \{ z + b \in Z_k : b \in B_k\}.
\end{equation}
In particular $[z + b ] = [z]$ for any $b \in B_k$. The dimension of $H_k$ as a vector space is denoted
\[
\beta_k \coloneqq  \dim(H_k),\qquad k=0, 1,\dots, d.
\]
This has special topological significance when $H_k$ is the homology group of a topological space like a simplicial complex $K$ and is called the $k$th \emph{Betti number} of $K$.
Intuitively $\beta_k$ counts the number of $k$-dimensional holes in $K$. Note that by definition, $H_k$ has a \emph{basis} comprising homology classes $[z_1],\dots,[z_{\beta_k}]$ for some $z_1,\dots,z_{\beta_k} \in Z_k \subseteq C_k$.

\subsection{Simplicial homology}\label{sec:simp_hom}

We present a very simple exposition of simplicial homology tailored to our purposes. The simplification stems partly from our working over a field of two elements  $\mathbb{F}_2 \coloneqq \{0,1\}$. In particular $-1 = +1$ and we do not need to concern with signs.

Given an abstract simplicial complex $K$, we define an $\mathbb{F}_2$-vector space $C_k(K)$ in the following way: Let $K^{(k)} = \{\sigma_1,\dots,\sigma_{m}\}$ be the set of all $k$-dimensional simplices in $K$. Then an element of $C_k(K)$ is a formal linear combination:
\[
\sum_{j=1}^{m}  n_j \sigma_j, \qquad n_j = 0 \text{ or } 1.
\]
In other words, $C_k(K)$ is a vector space over  $\mathbb{F}_2$ with $K^{(k)}$ as a basis.

The boundary operators $\partial_k : C_k(K) \to C_{k-1}(K)$ are defined on a $k$-simplex $\sigma = [v_0, \dots, v_k]$ by
\begin{equation}\label{eq:bdmaps}
\partial_k \sigma \coloneqq \sum_{j=0}^k [v_0, \dots, \hat{v}_j, \dots, v_k],
\end{equation}
where $\hat{v}_j$ indicates that $v_j$ is omitted from $\sigma$, and extended linearly to all of $C_k(K)$, i.e.,
\[
\partial_k \biggl(\sum_{j=1}^{m}  n_j \sigma_j\biggr) \coloneqq \sum_{j=1}^{m}  n_j \partial_k \sigma_j.
\]
For example, $\partial_1[a,b] = a + b$, $\partial_2 [a,b,c] = [a,b] + [b,c] + [c,a]$, $ \partial_2 ([a,b,c] + [d,e,f]) =  \partial_2 [a,b,c] +\partial_2  [d,e,f]$.

Working over $\mathbb{F}_2$ simplifies calculations enormously. In particular, it is easy to check that $\partial_k \ccirc  \partial_{k+1} = 0$ for all $k=0, \dots ,d$, as each $(k-2)$-simplex appears twice in the resulting sum and $2=0$ in $\mathbb{F}_2$. Thus \eqref{eq:bound} holds and $\partial_k : C_k(K) \to C_{k-1}(K)$, $k =0,\dots,d+1$ form a chain complex. The $k$th homology of the simplicial complex $K$ is then $H_k(K) = \ker(\partial_k)/ \im(\partial_{k+1})$ with $\partial_k$ as defined in \eqref{eq:bdmaps}. Working over $\mathbb{F}_2$ also guarantees that  $H_k(K)$ takes the simple form $\mathbb{F}_2^{\beta_k}$ where $\beta_k$ is the $k$th Betti number, i.e.,
\begin{equation}\label{eq:betti}
\beta_k(K) = \dim\bigl(H_k(K)\bigr) = \nullity(\partial_k) - \rank(\partial_{k+1}),
\end{equation}
for $k =0,1,\dots,d$. Let $m_k \coloneqq \lvert K^{(k)}\rvert$, the number of $k$-simplices in $K$.
To compute $\beta_k(K)$, note that with the $k$-simplices in $K^{(k)}$ as basis, $\partial_k$ is an $m_{k-1} \times m_k$ matrix  with entries in $\mathbb{F}_2$ and the problem in \eqref{eq:betti} reduces to linear algebra over $\{0,1\}$ with modulo $2$ arithmetic. While this seems innocuous, the cost of computing Betti numbers becomes prohibitive when the size of the simplicial complex $\lvert K\rvert = m_0 + m_1 + \dots + m_d$ is large. The number of simplices in a $d$-dimensional simplicial complex $K$ is bounded above by
\begin{equation}\label{eq:uppbd}
\lvert K\rvert \leq \sum_{i=0}^{d} {n \choose i+1}
\end{equation}
where $n$ is the size of the vertex set, i.e., $n = m_0$, and the bound is obtained by summing over the maximal number of simplices of each dimension. The cost of computing $\beta(K)$ is $\approx O(\lvert K\rvert^{2.38})$  \cite{Storjohann96}. %This rapidly growing upper bound,  more than anything else, is the main challenge in practical homology estimation and is the bottleneck of our experiments.

We conclude with a discussion of simplicial maps, which we will need in persistent homology. Let $K_1$ and $K_2$ be two abstract simplicial complexes. A \emph{simplicial map} is a map defined on their vertex sets $f:K_1^{(0)} \to K_2^{(0)}$ so that for each simplex  $\sigma = [v_0, \dots, v_k] \in K_1$,  we have that $[f(v_0), \dots, f(v_k)]$ is a simplex in $K_2$. Such a map induces a map between chain complexes that we will also denote by $f$, slightly abusing notation, defined by
\[
f : C_k(K_1) \to C_k(K_2), \qquad
\sum_{j=1}^{m}  n_j \sigma_j \mapsto \sum_{j=1}^{m}  n_j f(\sigma_j),
\]
that in turn induces a map between homologies
\begin{equation}\label{eq:fk}
H_k(f) :H_k(K_1) \to H_k(K_2), \qquad \biggl[\sum_{j=1}^{m}  n_j \sigma_j\biggr] \mapsto \biggl[\sum_{j=1}^{m}  n_j f(\sigma_j)\biggr]
\end{equation}
for all $k=0,1,\dots,d+1$.
Recall that $[z] \in H_k$ is a shorthand for homology class \eqref{eq:homcls}.

The composition of two simplicial maps $f:K_1^{(0)} \to K_2^{(0)}$ and $g:K_2^{(0)} \to K_3^{(0)}$ is also a simplicial map  $g \ccirc f : K_1^{(0)} \to K_3^{(0)}$  and thus induces a map between homologies
$H_k(g \ccirc f) :H_k(K_1) \to H_k(K_3)$ for any $k=0,1,\dots,d+1$. For the type of simplicial complex (Vietoris--Rips) and simplicial maps (inclusions) we consider in this article, we have that $H_k(g \ccirc f) = H_k(g) \ccirc H_k( f)$, a property known as functoriality.

\subsection{Vietoris--Rips complex}\label{sec:VR}

There are several ways to obtain a simplicial complex from a point cloud data set but one stands out for its simplicity and widespread adoption in topological data analysis. Note that a point cloud data set is simply a finite set of $n$ points $X  \subseteq \mathbb{R}^d$. We will build an abstract simplicial complex $K$ with vertex set $K^{(0)} = X$.

\begin{figure}[t]
%\centering
\begin{picture}(0,130)
\put(-230,15){\includegraphics[width=0.15\linewidth]{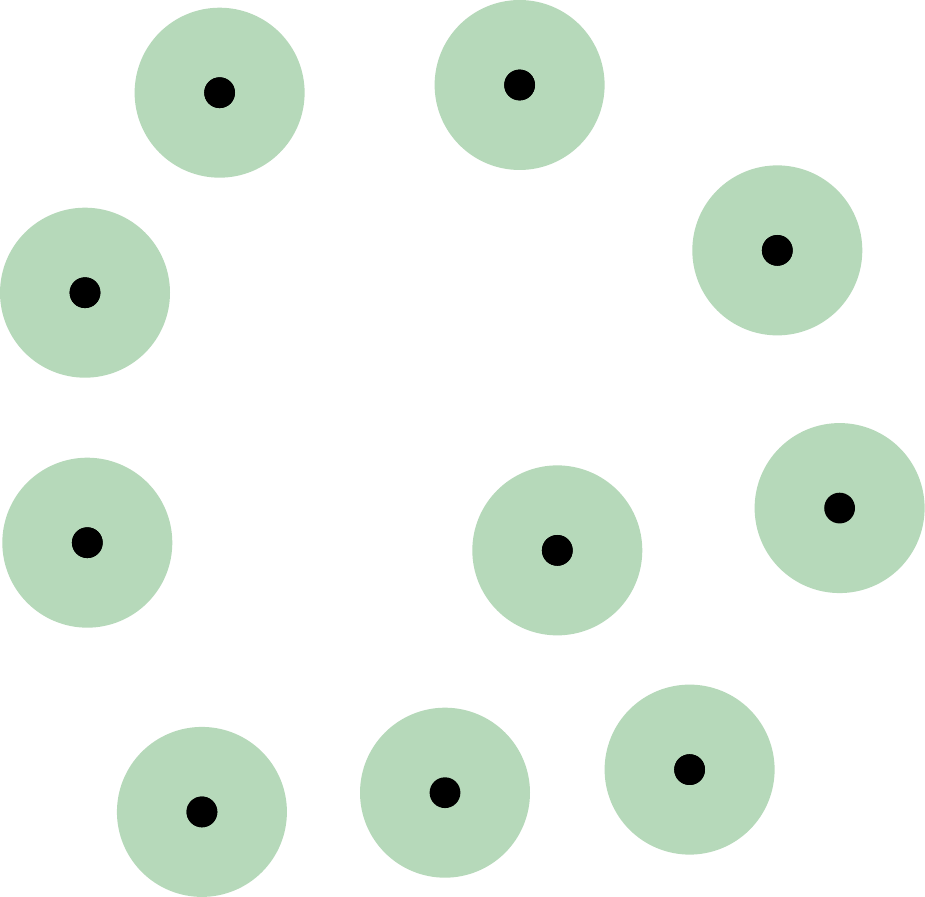}}
\put(-145,15){\includegraphics[width=0.15\linewidth]{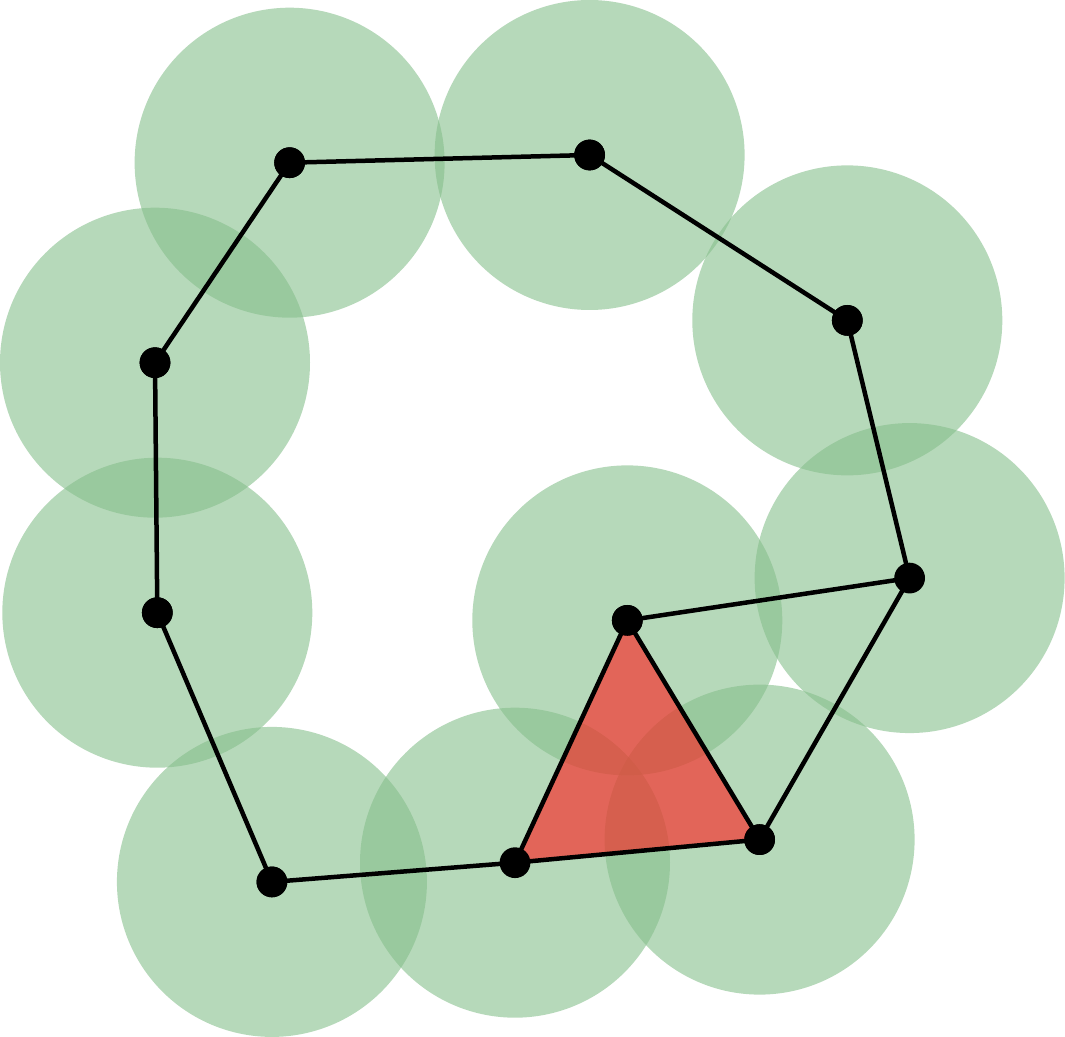}}
\put(-60,15){\includegraphics[width=0.15\linewidth]{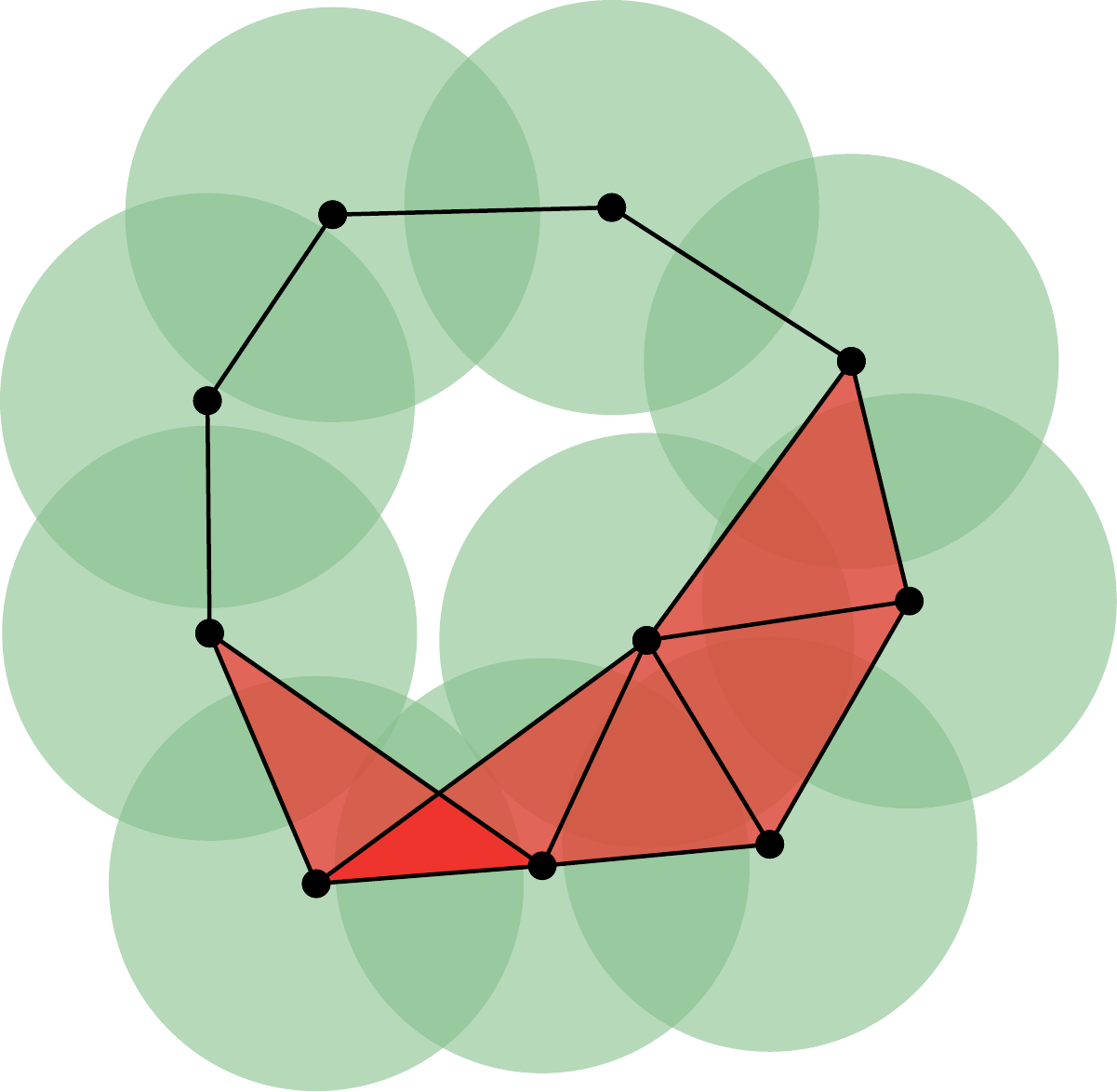}}
\put(20,5){\includegraphics[width=0.45\linewidth]{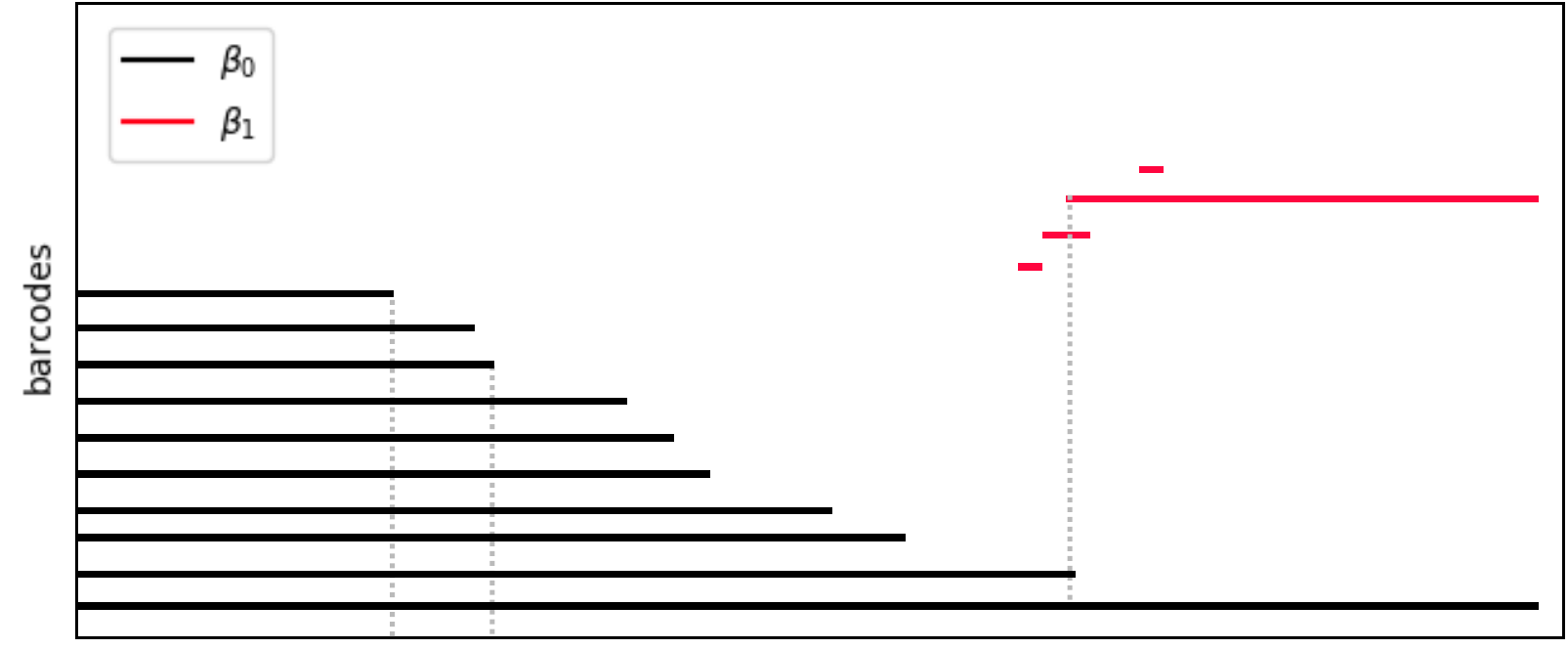}}
\put(52, -4){\scriptsize $\varepsilon= 0.3$}
\put(81, -4){\scriptsize $0.31$}
\put(157, -4){\scriptsize $0.39$}
\put(-208.5, 90){\scriptsize ${\varepsilon=0.15}$}
\put(-207.2, 80){\footnotesize ${}^{\to}{|\;\,}|^{\leftarrow}$}
\end{picture}  %}
\caption{\emph{Left}: Vietoris--Rips complex on ten points in $\mathbb{R}^2$ at scales  $\varepsilon=0.15$, $0.4$, $0.6$. \emph{Right}: Persistence barcodes diagram obtained from filtration of the Vietoris--Rips complex with scale $\varepsilon$ varying from $0$ to $4$. Barcodes show two most prominent topological features of the point cloud, the long black line at the bottom and the long red line near the top, revealing the topology of a circle, i.e., $\beta_0  = \beta_1 = 1$.  A $0$-homology class dies at times $\varepsilon = 0.3$, $0.31$, and $0.39$; a $1$-homology class is simultaneously born at time $\varepsilon = 0.39$.}
\label{fig:vr_complex}
\end{figure}
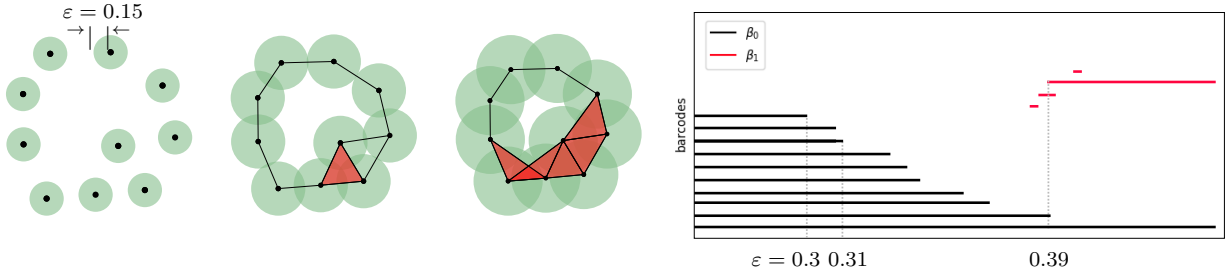

Let $\delta$ be a metric on $\mathbb{R}^d$. The \emph{Vietoris--Rips complex} at scale $\varepsilon\geq 0$ on $X$ is denoted by $\VR_\varepsilon(X)$ and defined  to be the simplicial complex whose vertex set is $X$ and whose $k$-simplices comprise all simplices $[x_0, \dots, x_k]$ satisfying  $\delta(x_i, x_j) \leq 2 \varepsilon$ for all $i, j =0,1,\dots,k$. In other words,
\[
\VR_\varepsilon (X) \coloneqq \bigl\{  [x_0, \dots, x_k] :  \delta(x_i, x_j) \leq 2 \varepsilon, \;
x_0,\dots,x_k \in X,  \; k = 0,1,\dots,n\bigr\}. 
\]
It follows immediately from definition that $\VR_\varepsilon ({X})$ is an abstract simplicial complex. Note that it depends on two things --- the scale $\varepsilon$ and the choice of metric $\delta$. Figure~\ref{fig:vr_complex} shows an example of Vietoris--Rips complex constructed from a point cloud data set of ten points in $\mathbb{R}^2$ at three different scales $\varepsilon = 0.15$, $0.4$, $0.6$ and with $\delta$ given by the Euclidean norm. 

For a point cloud $X \subseteq M \subseteq \mathbb{R}^d$ sampled from a manifold $M$ embedded in $\mathbb{R}^d$, the most appropriate metric $\delta$ is the geodesic distance on $M$ and not the Euclidean norm on $\mathbb{R}^d$. This is usually estimated from $X$ using the graph geodesic distance as we will see in Section~\ref{sec:comhom}.

When $X$ is sampled from a manifold $M \subseteq \mathbb{R}^d$, then for a dense enough sample, and at sufficiently small scale, the  topology of $\VR_\varepsilon (X)$  recovers the true topology $M$ in an appropriate sense, made precise in the following result in \cite{niyogi2008finding}:
\begin{prop}[Niyogi--Smale--Weinberger]\label{prop:NSW}
Let $X = \{x_1, \dots, x_n\} \subseteq \mathbb{R}^d$ be $(\varepsilon / 2)$-dense in a compact Riemannian manifold $M \subseteq \mathbb{R}^d$, i.e., for every $p\in M$, there exists $x \in X$ such that $\lVert p -x\rVert < \varepsilon / 2$. Let $\tau$ be the condition number of $M$. Then for any $\varepsilon< \sqrt{3\tau/5}$, the union of balls $V = \bigcup_{i=1}^n B_{\varepsilon}(x_i)$ deformation retracts to $M$. In particular, the homology of $V$ equals the homology of $M$.
\end{prop}
Roughly speaking the condition number of a manifold embedded in $\mathbb{R}^d$ encodes its local and global curvature properties but the details are too technical for us to go into here.

\subsection{Persistent homology}\label{sec:per_hom}

The Vietoris--Rips complex $\VR_\varepsilon (X)$  of a point cloud data set involves a parameter $\varepsilon$. Here we will discuss how this may be determined.

The homology classes, are very sensitive to small changes. For example, punching a small hole in a sphere has little effect on its geometry but has large consequence on its topology --- even a very small hole would kill the $H_2$ homology class, turning a sphere into a topological disk. This also affects the estimation of Betti numbers of a manifold from a subset of sampled point cloud data: there are many scenarios where moving a single point can significantly change the homology estimates.  \emph{Persistent homology} \cite{EdelsbrunnerPH} addresses this problem by blending geometry and topology. It allows one to reliably estimate the Betti numbers of a manifold from a point cloud data set, and to a large extent avoids the problem of the extreme sensitivity of topology to perturbations. In machine learning lingo, 
Betti numbers are features associated with the point cloud, and persistent homology enables one to identify the features that are robust to noise.

Informally, the idea of persistent homology is to introduce a geometric scale $\varepsilon$ that varies from $0$ to $\infty$ into homology calculations. At a scale of zero, $\VR_0 (X) = \{[x] : x\in X\}$ is a collection of $0$-dimensional simplices with $\beta_0=\lvert X\rvert$ and all other Betti numbers zero. In machine learning lingo the simplicial $\VR_0 (X)$ `overfits' the data $X$, giving us a discrete topological space. As $\varepsilon$ increases, more and more distant points come together to form higher and higher-dimensional simplices in  $\VR_\varepsilon (X)$  and its topology becomes richer. But as $\varepsilon \to \infty$,  eventually all points in $X$ become vertices of a single $\lvert X \rvert$-dimensional simplex, giving us a contractible topological space. So at the extreme ends $\varepsilon = 0$ and $\varepsilon \to \infty$, we have trivial (discrete and indiscrete) topologies  and the true answer we seek lies somewhere in between --- to obtain a `right' scale $\varepsilon_*$, we use the so-called \emph{persistence barcodes}. Figure~\ref{fig:vr_complex} shows an example of a persistence barcode diagram. This is the standard output of persistent homology calculations and it provides a summary of the evolution of topology across all scales. Generally speaking, a persistence barcode is an interval $[\varepsilon, \varepsilon']$ where its left-end point $\varepsilon$ is the scale at which the new feature appears or \emph{born}; and its right-end point $\varepsilon'$ is the scale at which that feature disappears or \emph{die}. The length of the interval $\varepsilon'-\varepsilon$ is the \emph{persistence} of that feature. Features that are non-robust to perturbations will produce short intervals; conversely, features that persist long enough, i.e., produce long intervals, are thought to be prominent features of the underlying manifold. For our purpose, the feature in question will always be a homology class in $k$th homology group. The collection of all persistence barcodes over $k =0,1,\dots,d$ then gives us our persistence barcode diagram.  If we sample a point cloud satisfying Proposition~\ref{prop:NSW} from a sphere with a small punctured hole, we expect to see a single prominent interval corresponding to $\beta_2 = 1$, and a short interval corresponding to the small hole. The persistence barcode would allow us to identify a scale $\varepsilon_*$ at which all prominent topological features of $M$ are represented, assuming that such a scale exists. In the following we will assume that we are interested in selecting  $\varepsilon_*$ from a list of finitely many scales $\varepsilon_0 < \varepsilon_1 < \dots < \varepsilon_m$ but that they could go as fine as we want. For our purpose, the simplicial complex below are taken to be $K_j = \VR(X, \varepsilon_j)$, $j =0,1,\dots,m$, but the following discussion  holds more generally.

We provide the details for computing persistence barcodes for homology groups, or \emph{persistent homology} in short. This essentially tracks the evolution of homology in a \emph{filtration} of simplicial complexes, which is chain of a nested simplicial complexes
\begin{equation}\label{eq:fil}
K_0 \subseteq K_1 \subseteq K_2 \subseteq \dots \subseteq K_m.
\end{equation}
We let $f_j:K_j \hookrightarrow K_{j+1}$, $j=1, \dots, m-1$ denote the inclusion maps where each simplex of $K_j$ is sent to the same simplex in $K_j \subseteq K_{j+1}$ and regarded as a simplex in $K_{j+1}$.  As $f_j$ is obviously a simplicial map and induces a linear map $H_k(f_j)$ between the homologies of $H_k(K_j)$ and $H_k(K_{j+1})$ as discussed in Section~\ref{sec:simp_hom}, composing inclusions $f_{j+p}\ccirc \dots \ccirc f_j$ gives us a linear map between any two complexes in a filtration $H_k(K_j)$ and $H_k(K_{j+p})$, $j=0, 1, \dots, m$, $p=1, 2, \dots, m-j$. The index $j$ is often referred to as `time' in this context. As such, for any $i<j$, one can tell whether two simplices belonging to two different homology classes in $H_k(K_i)$ are mapped to the same homology class in $H_k(K_j)$ --- if this happens, one of the homology class is said to have \emph{died} while the other has \emph{persisted} from time $i$ to $j$. If a homology class in $H_k(K_{j+1})$ is not in the image of $H_k(f_j)$, we say that its homology class is \emph{born} at time $j+1$. The  persistence barcodes simply keep track of  the birth and death times of the homology classes. 

\begin{figure}[ht]
\centering
\includegraphics[trim={70 0 0 0}, clip, width=1.05\linewidth]{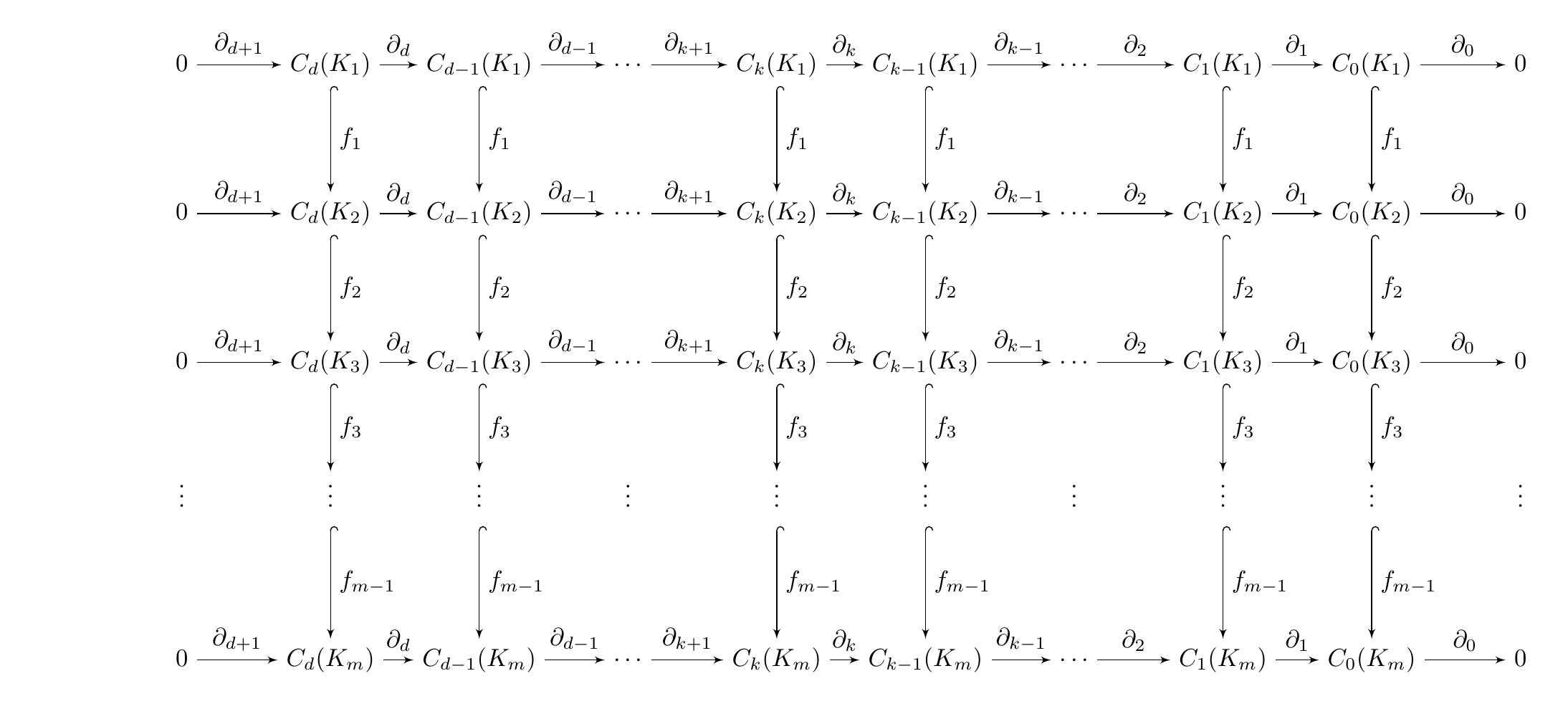}
\caption{Persistence complex of the filtration $K_0 \subseteq K_1 \subseteq  \dots \subseteq K_m$.}
\label{fig:ph_diagram_b}
\end{figure}

To be completely formal, we have the two-dimensional complex called a \emph{persistent complex} shown in Figure~\ref{fig:ph_diagram_b} with horizontal maps given by boundary maps $\partial_k : C_k(K_j) \to C_{k-1}(K_j) $ and vertical maps given by simplicial maps $f_j :C_k(K_j) \to C_k(K_{j+1}) $. 
Thanks to a well-known structure theorem \cite{MR2121296} which guarantees that a barcodes diagram completely describes the structure of a persistent complex in an appropriate sense, we may avoid persistent complexes like Figure~\ref{fig:ph_diagram_b} and work entirely with persistence barcodes diagram like the one on the right of Figure~\ref{fig:vr_complex}.

Henceforth we let  $K_j = \VR(X, \varepsilon_j)$, $j =0,1,\dots,m$, be the Vietoris--Rips complex of our point cloud data at scales $\varepsilon_0 < \varepsilon_1 < \dots < \varepsilon_m$. An important fact to note is that persistence barcodes may be computed without having to compute homology at every scale $\varepsilon_j$, or, equivalently, at every time $j$. To identify the homology classes in $H_k(K_j)$ that persist from time $j$ to time $j+p$, there is no need to compute $H_k(K_{j+1}),\dots, H_k(K_{j+p})$ individually as one might think. Rather, one considers the \emph{$p$-persistent $k$th homology group}  
\[
H_k^{j,p} = Z_k^j / (B_k^{j+p}  \cap Z_k^j ),
\]
where $p=1,2\dots,m-j$. This captures the cycles in $C_k(K_j)$ that contribute to homology in $C_k(K_{j+p})$. One may consistently choose a basis for each $H_k^{j,p}$ so that the basis elements are compatible for homologies across $H_k(K_{j+1}),\dots, H_k(K_{j+p})$ for all possible values of $k$ and $p$. This allows one to track the persistence of each homology class throughout the filtration \eqref{eq:fil} and thus obtain the persistence barcodes: roughly speaking, with the right basis, we may simultaneously represent the boundary maps on $C_k(K_j)$ as matrices in an column-echelon form and read-off the dimension of $H_k^{j,p}$, known as the \emph{$p$-persistent $k$th Betti number} $\beta^{j,p}_k$, from the  pivot entries in these matrices. For details we refer readers to \cite{EdelsbrunnerPH, MR2121296}.

\subsection{Homology computations in practice}\label{sec:practice}

Actual computation of homology from a point cloud data set is more involved than what one might surmise from the description in the last few sections. We will briefly discuss some of the issues involved.

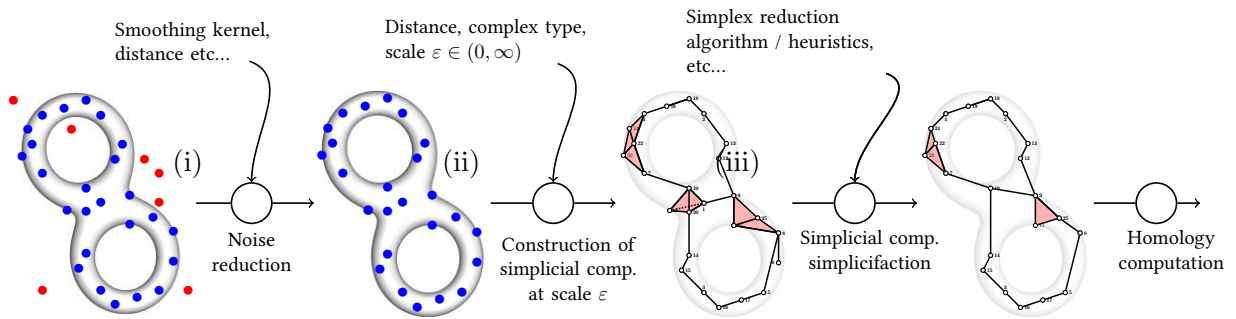
\begin{figure}[ht]
\begin{picture}(-50,130)
\put(-238,0){\includegraphics[width=1\linewidth]{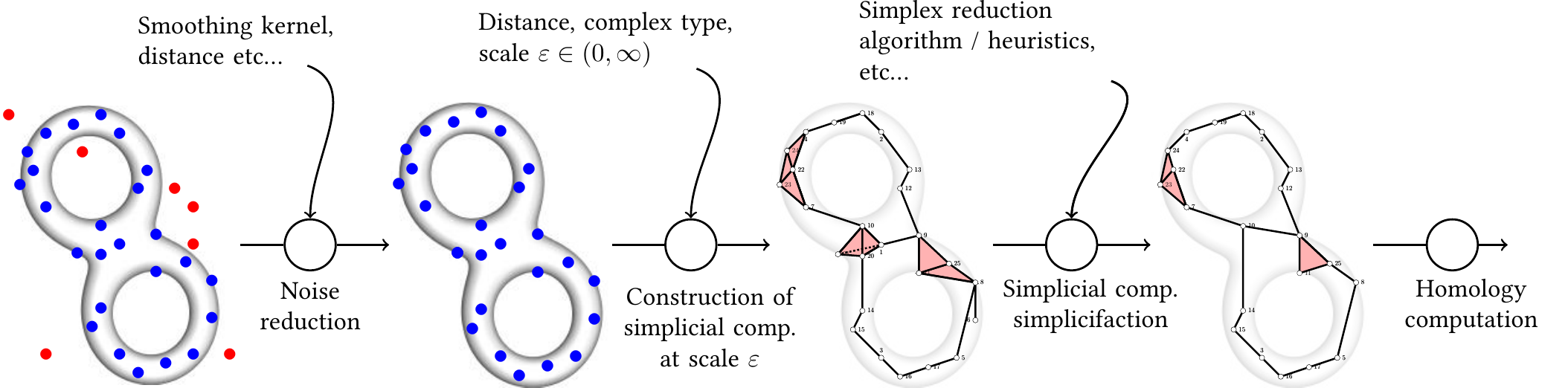}}
\put(-175, 55){(i)}
\put(-73, 55){(ii)}
\put( 30, 55){(iii)}
\end{picture}  %}
\caption{Pipeline for computation of homology from a point cloud data.}
\label{fig:homology_computation}
\end{figure}

Before we even begin to compute the homology of the point cloud data $X\subseteq M \subseteq \mathbb{R}^d$, we will need to perform a few preprocessing steps, as depicted in Figure~\ref{fig:homology_computation}. These steps are standard practice in topological data analysis: (i) We  smooth out  $X$ and discard outliers to reduce noise. (ii) We then select the scale $\varepsilon$ and constructing the corresponding Vietoris--Rips complex $\VR_\varepsilon(X)$. (iii) We  simplify $\VR_\varepsilon(X)$ in a way that reduces its size but leaving its topology unchanged. All processing operations that can have an effect on the homology are in steps (i) and (ii), and the homology of the simplicial complex $\VR_\varepsilon(X)$ is assumed to closely approximate that of the underlying manifold $M$.  The simplification in step (iii) is done to accelerate computations without altering homology.

Note that the size of the final simplicial complex on which we perform homology calculations is the most important factor in computational cost. While increasing  the number of points sampled from a manifold, i.e., the size of $X$, up to the point in Proposition~\ref{prop:NSW} improves the accuracy of our homology estimates, it also results in a simplicial complex $\VR_\varepsilon(X)$ that is prohibitively large for carrying out computations, as we saw in \eqref{eq:uppbd}.  But since we are not concerned with the geometry of the underlaying manifold, only its topology, it is desirable to construct a small simplicial complex with minimal topology distortion. A well-known simplification is the \emph{Witness complex} \cite{witness2004}, which gives results of nearly the same quality with a simplicial complex of a smaller size constructed from so-called landmark points.  Many other methods have been proposed for this \cite{boissonnat, MR3171260, MR3090522}, and while we will take advantage of these techniques in our calculations, we will not discuss them here.

The takeaway is that persistence barcodes are considerably more expensive to compute than homology  at a single fixed scale $\varepsilon$. Therefore, running full persistent homology in the context of modern deep neural network poses some big challenges: modern deep neural networks operate on very high dimensional big data sets, a setting in which persistent homology cannot be used directly due to computation and memory complexity. This situation is exacerbated by the fact that neural networks are randomly trained (with  potentially big variation in the learned decision boundaries), therefore one needs to run many computations to obtain reliable results. Furthermore, an automated statistical analysis of persistent homology is still an active area of research and often requires additional large computational effort. It seems therefore largely beyond the reach of current technology to try to analyze topology of many of the standard deep learning data sets (such as \texttt{SVHN},  \texttt{CIFAR-10}, \texttt{ImageNet}, see \cite{SVHN, cifar10, imagenet_cvpr09}). We will return to this point later when we introduce our methodology for monitoring topology transformations in a neural network. In particular, we will see in Section~\ref{sec:comhom} that our experiments are designed in such a way that although we will compute homology at every layer, we only need to compute persistence barcodes once, before the data set is passed through the layers.

\section{Overview of problem and methodology}\label{sec:problem}

We will use \emph{binary classification}, the most basic and fundamental problem in supervised learning, as our platform for studying how neural networks change topology. More precisely, we seek to classify two different probability distributions supported on two disjoint manifolds $M_a$, $M_b \subseteq \mathbb{R}^d$. The distance $\inf\{\lVert x - y \rVert : x \in M_a, \; y \in M_b\}$ can be arbitrarily small but not zero. So there exists an ideal classifier with zero prediction error. Here and henceforth, $\lVert \, \cdot \, \rVert$ will denote the Euclidean norm in $\mathbb{R}^d$.

We sample a large but finite set of points  $T \subseteq M_a \cup M_b$ uniformly and densely, so that the Betti numbers of $M_a$ and $M_b$ can be faithfully obtained from the point cloud data sets $T \cap M_a$ and $T \cap M_b$ as described in Section~\ref{sec:background}. Our training set is a labeled point cloud data set, i.e., $x \in T$  is labeled to indicate whether $x\in M_a$ or $M_b$. We will use $T_a \coloneqq T \cap M_a$ and $T_b \coloneqq T \cap M_b$, or rather, their Vietoris--Rips complex as described in Section~\ref{sec:VR}, as finite proxies for $M_a$ and $M_b$.

Our feedforward neural network $\nu:\mathbb{R}^d \to [0,1]$ is given by the usual composition 

\begin{equation}\label{network}
\nu  = s \ccirc f_l \ccirc f_{l-1} \ccirc  \dots \ccirc f_2 \ccirc f_1,
\end{equation}
where each \emph{layer} of the network $f_j : \mathbb{R}^{n_j} \to \mathbb{R}^{n_{j+1}}$, $j=1, \dots, l$, is the composition of an affine map $\rho_j : \mathbb{R}^{n_j} \to \mathbb{R}^{n_{j+1}}$, $x \mapsto A_j x + b_j$, with an \emph{activation} function $\sigma : \mathbb{R}^{n_{j+1}} \to  \mathbb{R}^{n_{j+1}}$; and $s:\mathbb{R}^{n_l} \to [0,1]$ is the \emph{score} function. The \emph{width} $n_j$ is the number of nodes in the $j$th layer and we set $n_1 = d$ and $n_l = p$. For $j =1,\dots,l$, the composition of the first through $j$th layers is denoted
\[
\nu_j\coloneqq f_j \ccirc \dots \ccirc f_2  \ccirc f_1\quad \text{and} \quad \nu = s \ccirc \nu_l.
\]
We assume that $s$ is a \emph{linear classifier} and thus the decision boundary of $s$ is a hyperplane in $\mathbb{R}^p$. For notational convenience later, we define the `$(l+1)$th layer' $\nu_{l+1}\coloneqq s$ to be the score function and the `$0$th layer' $\nu_0$ to be the identity function on $\mathbb{R}^d$.

We train an $l$-layer neural network $\nu :\mathbb{R}^d \to [0,1]$ on a training set $T \subseteq M_a \cup M_b$ to classify samples into class $a$ or $b$. As usual, the network's output for a sample $x\in T$ is interpreted to be the probability of $x \in M_a$. In all our experiments, we train $\nu$ until it correctly classifies all $x \in T$ --- we will call such a network $\nu$ \emph{well-trained}. In fact, we sampled $T$ so densely that in reality $\nu$ also has near zero misclassification error on any test set $S \subseteq (M_a \cup M_b) \setminus T$; and  we trained $\nu$ so thoroughly that its output is concentrated near $0$ and $1$. For all  intents and purposes, we may treat $\nu$ as an ideal classifier.

We deliberately choose $M_a \cup M_b$ to have doubly complicated topologies in the following sense:
\begin{enumerate}[\upshape (i)]
\item For each  $i=a,b$, the component $M_i$ itself will have complicated topologies, with multiple components, i.e., large $\beta_0(M_i)$, as well as multiple $k$-dimensional  holes, i.e., large $\beta_k (M_i)$.

\item In addition, $M_a$ and $M_b$ will be entangled in a topologically complicated manner. See Figures~\ref{fig:data_sets} and \ref{fig:ds2_ds3} for example. They not only cannot be separated by a hyperplane but any decision boundary $D \subseteq \mathbb{R}^d$ that separates them will necessarily have complicated topology itself.
\end{enumerate}
In terms of the topological complexity in \eqref{eq:TC}, $\omega(M_a)$, $\omega(M_b)$, $\omega(D)$ are all large.

Our experiments are intended to show the topologies of $\nu_j(M_a)$ and $ \nu_j(M_b)$ evolve as $j$ runs from $1$ through $l$, for different manifolds $M_a, M_b$ entangled in different ways, for different number of layers $l$ and choices of widths $n_1,\dots,n_d$, and different activations $\sigma$. Getting ahead of ourselves, the results will show that a well-trained neural network $\nu : \mathbb{R}^d \to [0,1]$ reduces the topological complexity of $M_a$ and $M_b$ on a layer-by-layer basis until, at the output, we see a simple disentangled arrangement where the point cloud $T$ gets mapped into two clusters of points $\nu(T_a)$ and $\nu(T_b)$  on opposite ends of $[0, 1]$. This indicates that an initial decision boundary $D \subseteq \mathbb{R}^d$ of complicated topology ultimately gets transformed into a hyperplane in $\mathbb{R}^p$ by the time it reaches the final layer. We measure and track the topologies of $\nu_j(M_a)$ and $ \nu_j(M_b)$ directly, but our approach only permits us to indirectly observe the topology of the decision boundary separating them.

\subsection{Real versus simulated data}

We perform our experiments on a range of both real-world and simulated data sets to validate our premise that a neural network operates by simplifying topology.  We explain why each is indispensable to our goal.

Unlike real-world data, simulated data may be generated in a controlled manner with well-defined topological features that are known in advance (crucial for finding a single scale for all homology computations). Moreover, with simulated data we have clean samples and may skip the denoising step mentioned in the previous section. We can generate samples that are uniformly distributed on the underlaying manifold, and ensure that the assumptions of Section~\ref{sec:problem} are satisfied. In addition, we may always simulate a data set with a perfect classifier, whereas such a classifier for a real-wold data set may not exist when the probability distributions of different categories overlap. For convincing results, we train our neural network to perfect accuracy on training set and near-zero generalization error --- this may be impossible for real-world data. Evidently if there is no complete separation of one category $M_a$ from the other $M_b$, i.e., $M_a \cap M_b \ne \varnothing$, the manifold $M = M_a \cup M_b$ will be impossible to disentangle. Such is often the case with real-world data sets, which means that they may not fit our required setup in Section~\ref{sec:problem}. 

Nevertheless, the biggest issue with real-world data sets is that they have \emph{vastly} more complicated topologies that are nearly impossible to determine in advance. Even something as basic as the Mumford data set \cite{LeePM03}, a mere collection of $3 \times 3$-pixels of high contrast patches of natural images, took many years to have its topology determined \cite{carlsson2008local} and whether the conclusion (that it has the topology type of a Klein bottle) is correct is still a matter of debate. Figuring out, say, the topology of the manifold of cat images within the space of all possible images is well-beyond our capabilities for the foreseeable future.

Since our experiments on simulated data allow us to pick the right scale to compute homology, we only need to compute homology at one single scale. On the other hand, for real data we will need to find the persistence barcodes, i.e., determine homology over a full range of scales.  Consequently, our experiments on simulated data are  extensive --- we repeat our experiments for each simulated data set over a large number of neural networks of different architectures to examine their effects on topology changes. In all we ran more than 10,000 homology computations on our simulated data sets since we can do them fast and accurate. In comparison, our experiments on real-world data are more limited in scope as it is significantly more expensive to compute persistence barcodes then to compute homology at a single scale.
As such, we use simulated data to fully explore and investigate the effects of depth, width, shapes, activation functions, and various combinations of these factors on the topology-changing power of neural networks. Thereafter we use real-world data to validate the findings we draw from the simulated data sets.

\section{Methodology}\label{sec:methodology}

In this section, we will describe the full details of our methodology for (i) simulating topologically nontrivial data sets in a binary classification problem; (ii) training a variety of neural networks to near-perfect accuracy for such a problem; (iii) determining the homology of the data set as it passes through the layers of such a neural network.
For real data sets, step (i) is of course irrelevant, but steps (ii) and (iii) will apply with minor modifications; these discussions will be deferred to Section~\ref{sec:real_data}.

The key underlying reason for designing our experiments in the way we did is relative computational costs:
\begin{itemize}
\item multidimensional persistent homology is much more costly than persistent homology;
\item persistent homology is much more costly than homology;
\item homology is much more costly than training neural networks.
\end{itemize}
As such, we train tens of thousands of neural networks to near zero generalization error; for each neural network, we compute homology at every layer but we compute persistent homology only once; and we avoid multidimensional persistent homology altogether.

\subsection{Generating data sets}\label{sec:data}
 
We generate three point cloud data sets D-I, D-II, D-III in a controlled manner to have complicated but manageable topologies that we \emph{know in advance}. 

\begin{figure}[h]
\centering
\includegraphics[width=0.3\linewidth]{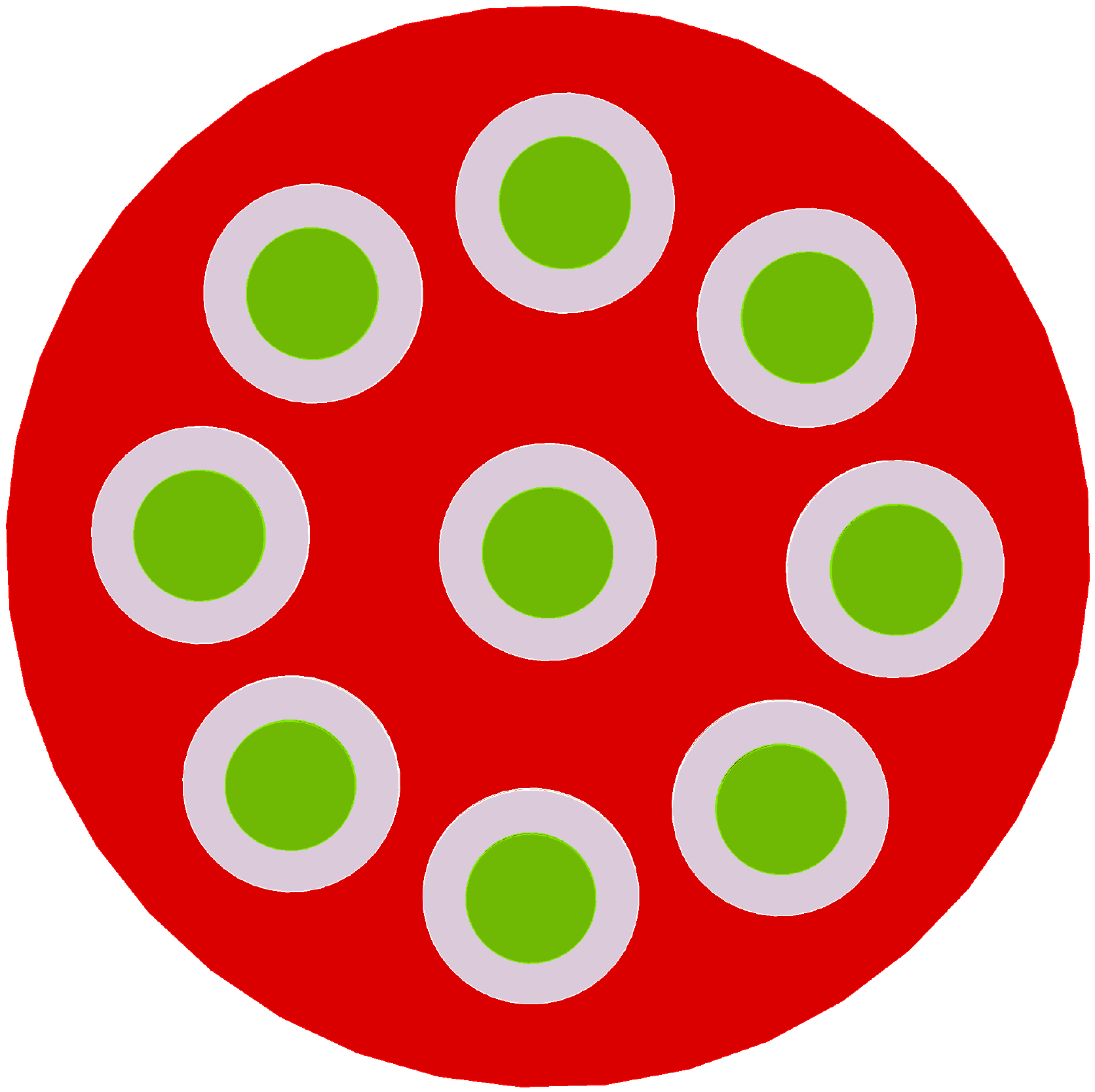}
\includegraphics[width=0.32\linewidth]{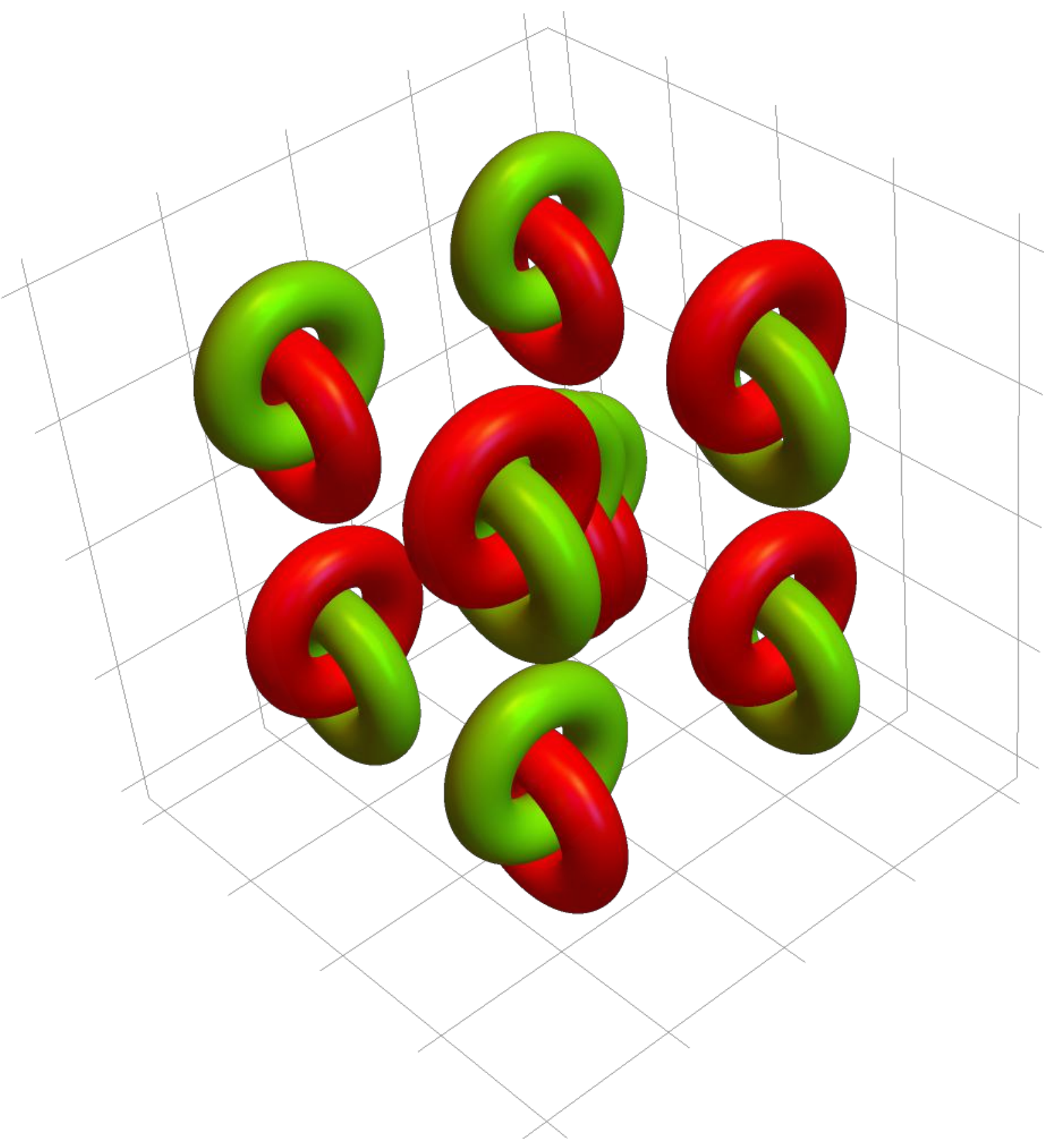}~~
\includegraphics[width=0.32\linewidth]{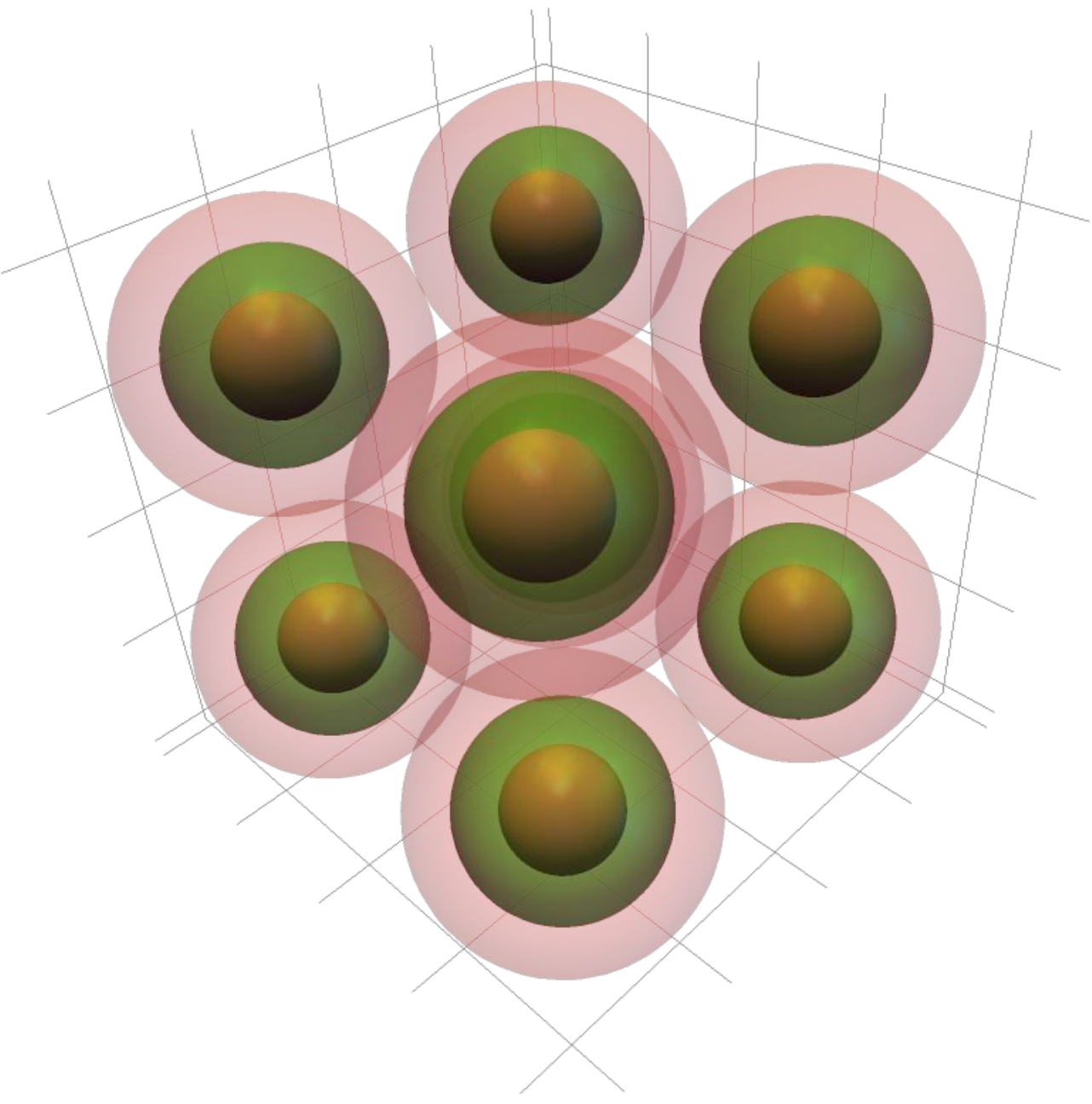}
\caption{The manifolds underlying data sets D-I, D-II, D-III (\emph{left to right}). The green $M_a$ represents category $a$, the red $M_b$ represents category $b$.}
\label{fig:data_sets}
\end{figure}
\vspace*{-3ex}
\begin{figure}[h]
\begin{minipage}[b]{0.99\textwidth}
\centering
\includegraphics[scale=0.25]{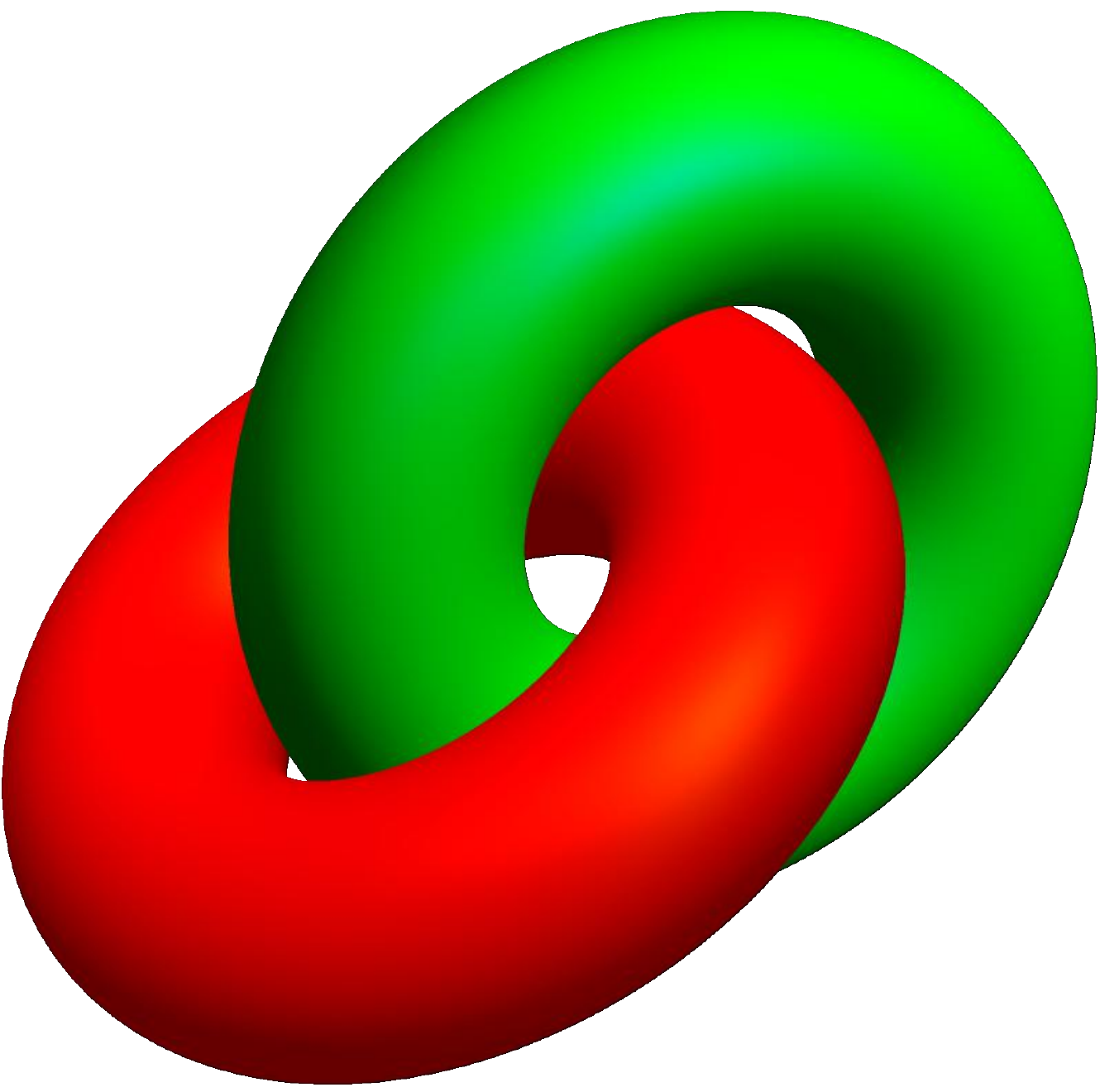}
\hspace{2.4cm}
\includegraphics[scale=0.25]{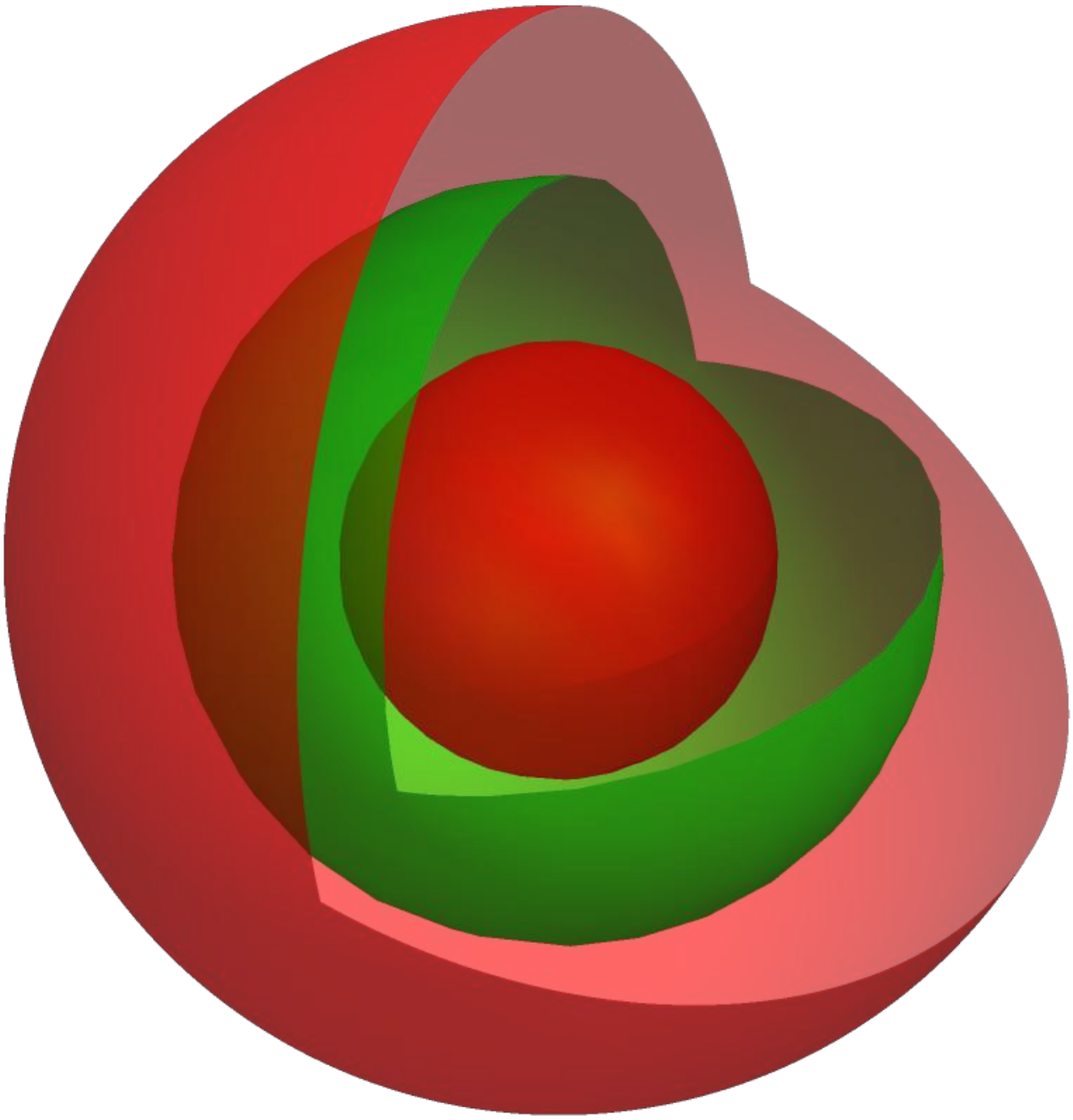}
\caption{\emph{Left}: D-II comprises nine pairs of such interlocking rings. \emph{Right}: D-III comprises nine units of such doubly concentric spheres.}
\label{fig:ds2_ds3}
\end{minipage}
\end{figure}

\textbf{D-I} is sampled from a two-dimensional manifold consisting of $M_a$, nine green disks, positioned in $M_b$, a larger disk with nine holes as on the left in Figure~\ref{fig:data_sets}. We clearly have $\beta(M_a) = (9, 0)$ and $\beta(M_b) =(1, 9)$ (one connected component, nine holes). \textbf{D-II} is sampled from a three-dimensional manifold comprising nine disjoint pairs of red solid torus interlocked with a green solid torus (a single pair is shown in Figure~\ref{fig:ds2_ds3}). $M_a$ (resp.\ $M_b$) is the union of all  nine green (resp.\ red) tori.  So $\beta(M_a) = \beta(M_b) =  (9, 9, 0)$. \textbf{D-III} is sampled from a three-dimensional manifold comprising nine disjoint units of the following --- a large red sphere enclosing a smaller green sphere enclosing a red ball; the green sphere is trapped between the red sphere and the red ball. $M_a$ is the union of all nine green spheres and $M_b$ is the union of the nine spheres and nine balls. So we have $\beta(M_a)  =  (9, 0, 9)$ and = $\beta(M_b)= (18, 0, 9)$ (see Figures~\ref{fig:data_sets} and \ref{fig:ds2_ds3} for more details, but note on Figure~\ref{fig:ds2_ds3} the spheres are shown with portions omitted). In all cases, the two categories $a$ and $b$ are entangled in such a way that any decision boundary separating the two categories necessarily has highly complex topology.

The point cloud data sets D-I and D-III are sampled on a grid whereas D-II is sampled uniformly from the solid tori. The difference in sampling schemes is inconsequential for all intents and purposes in this article,  as the samples are sufficiently dense that there is no difference in training and testing behaviors.

\subsection{Training neural networks}\label{sec:nn}

Our goal is to examine the topology changing effects of (i) different \emph{activations}: hyperbolic tangent, leaky ReLU set to be $\max(x, 0.2x)$, and ReLU; (ii) different \emph{depths} of four to ten layers; (iii) different \emph{widths} of six to fifty neurons. So for any given data set (D-I, D-II, D-III) and any given architecture (depth, width, activation), we tracked the Betti numbers through all layers for at least $30$ well-trained neural networks. The repetition is necessary --- given that neural network training involves a fair amount of randomization in initialization, batching, optimization, etc --- to ensure that what we observe is not a fluke.

To train these neural networks to our requirements --- recall that this means zero training error and a near-zero ($\approx 0.01\%$) generalization error --- we relied on \textsf{TensorFlow} (version 1.12.0 on Ubuntu 16.04.1). Training is done on cross-entropy categorical loss with standard Adam optimizer \cite{kingma2014adam}  for up to 18,000 training epochs. Learning rate is set to $0.02$--$0.04$ with an exponential decay, i.e., $\eta^{t/ {d}}$ where $t$ is the training epoch normalized by $d=2500$. For the `bottleneck architectures' where the widths narrow down in the middle layers (see Table~\ref{tab:experiments}), the decay is set to $4000$ and $\eta=0.5$. We use the $\operatorname{softmax}$ function as the score function in all of our networks, i.e., $s : \mathbb{R}^p \to \mathbb{R}^p$ whose $i$th coordinate is
\[
s_i(x) = e^{x_i}/( e^{x_1}+\dots+e^{x_p}),\qquad i =1, \dots, p,
\]
where $p$ is the number of categories. In our case, $p=2$ and $i =a,b$.

Table~\ref{tab:experiments} summarizes our results: the data set used, the activation type, the widths of each layer, and the number of successfully trained neural networks of that architecture obtained. The first number in the sequence of the third column gives the dimension of the input, which is two for the two-dimensional D-I and three for the three-dimensional D-II and D-III. The last number in that sequence is always two since they are all binary classification problems. To give readers an idea, training any one of these neural networks to near zero generalization error takes at most 10 minutes, often much less.

\begin{table}[!h]
\centering
\scriptsize
\begin{tabular}[c]{cclccc}
\hline
\hline
%\rowcolor{black!5}
data set & activation & \multicolumn{1}{c}{neurons in each layer} & \# \\
\hline
D-I & $\tanh$ &  2-15-15-15-15-15-15-15-15-15-15-2 & 30\\ 
D-I & leaky ReLU & 2-15-15-15-15-15-15-15-15-15-15-2 & 30\\ 
D-I & leaky ReLU &  2-05-05-05-05-03-05-05-05-05-05-2 & 30\\ 
D-I & leaky ReLU &  2-15-15-15-15-03-15-15-15-15-2 & 30\\ 
D-I & leaky ReLU &  2-50-50-50-50-50-50-50-50-50-50-2 & 30\\ 
D-I & ReLU  & 2-15-15-15-15-15-15-15-15-15-15-2 & 30\\
\hline
D-II & $\tanh$ & 3-15-15-15-15-15-15-15-15-15-15-2 & 32\\
D-II & leaky ReLU & 3-15-15-15-15-15-15-15-15-15-15-2 & 36\\
D-II & ReLU & 3-15-15-15-15-15-15-15-15-15-15-2 & 31\\
\hline
D-II & $\tanh$ &  3-25-25-25-25-25-25-25-25-25-25-2 & 30\\ 
D-II & leaky ReLU & 3-25-25-25-25-25-25-25-25-25-25-2 & 30\\
D-II & ReLU  & 3-25-25-25-25-25-25-25-25-25-25-2 & 30\\
\hline
D-III & $\tanh$ & 3-15-15-15-15-15-15-15-15-15-15-2 & 30 \\
D-III & leaky ReLU & 3-15-15-15-15-15-15-15-15-15-15-2 & 46 \\ 
D-III & ReLU & 3-15-15-15-15-15-15-15-15-15-15-2 & 30 \\ 
\hline 
D-III & $\tanh$ & 3-50-50-50-50-50-50-50-50-50-50-2 & 30 \\  
D-III & leaky ReLU & 3-50-50-50-50-50-50-50-50-50-50-2 & 30 \\
D-III & ReLU & 3-50-50-50-50-50-50-50-50-50-50-2 & 34 \\
\hline
D-I & $\tanh$ & 2-15-15-15-15-2 & 30 \\
D-I & $\tanh$ &2-15-15-15-15-15-15-15-15-2& 30 \\
\hline
D-I & leaky ReLU & 2-15-15-15-15-2& 30\\
D-I & leaky ReLU &2-15-15-15-15-15-15-15-15-2& 30 \\
\hline
D-I & ReLU & 2-15-15-15-15-2 & 30\\
D-I & ReLU &2-15-15-15-15-15-15-15-15-2& 30 \\
\hline
D-II & $\tanh$ & 3-15-15-15-15-2 & 31 \\
D-II & $\tanh$ & 3-15-15-15-15-15-2 & 31 \\
D-II & $\tanh$ & 3-15-15-15-15-15-15-15-2 & 30 \\
\hline
D-II & leaky ReLU & 3-15-15-15-15-2 & 31 \\
D-II & leaky ReLU & 3-15-15-15-15-15-2 & 30 \\
D-II & leaky ReLU & 3-15-15-15-15-15-15-2 & 30 \\
D-II & leaky ReLU & 3-15-15-15-15-15-15-15-2 & 31 \\
D-II & leaky ReLU & 3-15-15-15-15-15-15-15-15-2 & 42 \\ 
\hline
D-II & ReLU & 3-15-15-15-15-2 & 32 \\
D-II & ReLU & 3-15-15-15-15-15-2 & 32 \\
D-II & ReLU & 3-15-15-15-15-15-15-15-15-2& 31 \\
\hline
D-III & $\tanh$ & 3-15-15-15-15-15-15-2 & 30 \\
D-III & $\tanh$ &  3-15-15-15-15-15-15-15-15-2& 31 \\
\hline
D-III & leaky ReLU & 3-15-15-15-15-15-15-2 & 30 \\
D-III & leaky ReLU &  3-15-15-15-15-15-15-15-15-2& 30 \\
\hline
D-III & ReLU & 3-15-15-15-15-15-15-2 & 33 \\
D-III & ReLU &  3-15-15-15-15-15-15-15-15-2& 32 \\
\hline
\hline
\end{tabular}
\bigskip
\caption{First column specifies the data set on which we train the networks. Next two columns give the activation used and a sequence giving the number of neurons in each layer. Last column gives the number of well-trained networks obtained.}
\label{tab:experiments}
\vspace*{-3ex}
\end{table}

\subsection{Computing homology}\label{sec:comhom}

For each of the neural networks obtained in Section~\ref{sec:nn}, we track how the topology of the respective point cloud data set changes as it passed through the layers. This represents the bulk of the computational effort, way beyond that required for training neural networks in Section~\ref{sec:nn}. 
With simulated data, we are essentially assured of a perfectly clean data set and the preprocessing step  in Figure~\ref{fig:homology_computation} of Section~\ref{sec:practice} may be omitted. We describe the rest of the  work involved below.

The metric $\delta$ used to form our Vietoris--Rips complex is given by the graph geodesic distance on the $k$-nearest neighbors graph determined by the point cloud $X \subseteq \mathbb{R}^d$. As this depends on $k$, a positive integer specifying the number of neighbors used in the graph construction, we denote the metric by $\delta_k$. In other words, the Euclidean distance on $\mathbb{R}^d$ is used only to form the $k$-nearest neighbors graph and do not play a role thereafter. For any $x_i,x_j \in X$, the distance $\delta_k(x_i,x_j)$ is given by the minimal number of edges between them in the $k$-nearest neighbors graph. Each edge, regardless of its Euclidean length, has the same length of one when measured in $\delta_k$.

The metric $\delta_k$ has the effect of normalizing distances across layers of a neural network while preserving connectivity of nearest neighbors. This is important for us as the local densities of a point cloud can vary enormously as it passes through a layer of a well-trained neural network --- each layer stretches and shrinks different regions, dramatically altering geometry as one can see in  the bottom halves of Figures~\ref{fig:2d_dataset}, \ref{fig:dataset_ii}, and \ref{fig:dataset_iii}. Using an intrinsic metric like $\delta_k$ ameliorates  this variation in densities; it is robust to geometric changes and yet reveals topological ones.
Furthermore, our choice of $\delta_k$ allows for comparison across layers with different numbers of neurons. Note that if $d \ne p$, the Euclidean norms on $\mathbb{R}^d$ and $\mathbb{R}^p$ are two different metrics on two different spaces with no basis for comparison. Had we used Euclidean norms, two Vietoris--Rips complexes of the same scale $\varepsilon$ in  two different layers cannot be directly compared --- the scale needs to be calibrated somehow to reflect that they live in different spaces. Using $\delta_k$ avoids  this problem. 

This leaves us with two parameters to set: $k$, the number of neighbors in the nearest neighbors graph and $\varepsilon$, the scale at which to build our Vietoris--Rips complex. This is where persistent homology, described at length in Section~\ref{sec:per_hom}, comes into play. Informed readers may think that we should be using \emph{multidimensional persistence} since there are two parameters but this is prohibitively expensive as the problem is EXPSPACE-complete \cite{mult2} and its results are not quite what we need, for one, there is no multidimensional analogue of persistence barcodes \cite{mult1}. To choose an appropriate $(k_*,\varepsilon_*)$ for a point cloud $X \subseteq M \subseteq \mathbb{R}^d$, we construct a filtered complex over the two parameters: Let $\VR_{k,\varepsilon}(X)$ be the Vietoris--Rips complex of $X$ with respect to the  metric $\delta_k$ at scale $\varepsilon$. In our case, we know the topology of the underlying manifold $M$ completely as we generated it in Section~\ref{sec:data} as part of our data sets. Thus we may ascertain whether our chosen value $(k_*,\varepsilon_*)$ gives a Vietoris--Rips complex $\VR_{k_*,\varepsilon_*}(X)$ with the same homology as $M$.

Set $\varepsilon = 1$, we determine a value of $k_*$ with persistent homology on the $k$-filtered complex in the metric $\delta_k$ with correct zeroth homology, i.e., $k_*$ is chosen so that
\[
\beta_0\bigl(\VR_{k_*,1}(X)\bigr) = \beta_0(M).
\]
Set $k=k_*$, we determine a value of $\varepsilon_*$ with persistent homology on the $\varepsilon$-filtered complex in the metric $\delta_{k_*}$ with correct first and second homologies, i.e., $\varepsilon_*$ is chosen so that
\[
\beta_1\bigl(\VR_{k_*,\varepsilon_*}(X)\bigr) = \beta_1(M) \qquad \text{and}\qquad
\beta_2\bigl(\VR_{k_*,\varepsilon_*}(X)\bigr) = \beta_2(M).
\]
If there is a range of parameters that  all recover the correct homology we pick our $(k_*,\varepsilon_*)$ closest to the middle of the range. Once these parameters are set, we keep them fixed for our homology computations across all layers of the network. 

\begin{figure}[ht]
\centering
\includegraphics[trim={2.0cm 0 0.3cm 0},clip, width=0.48\linewidth]{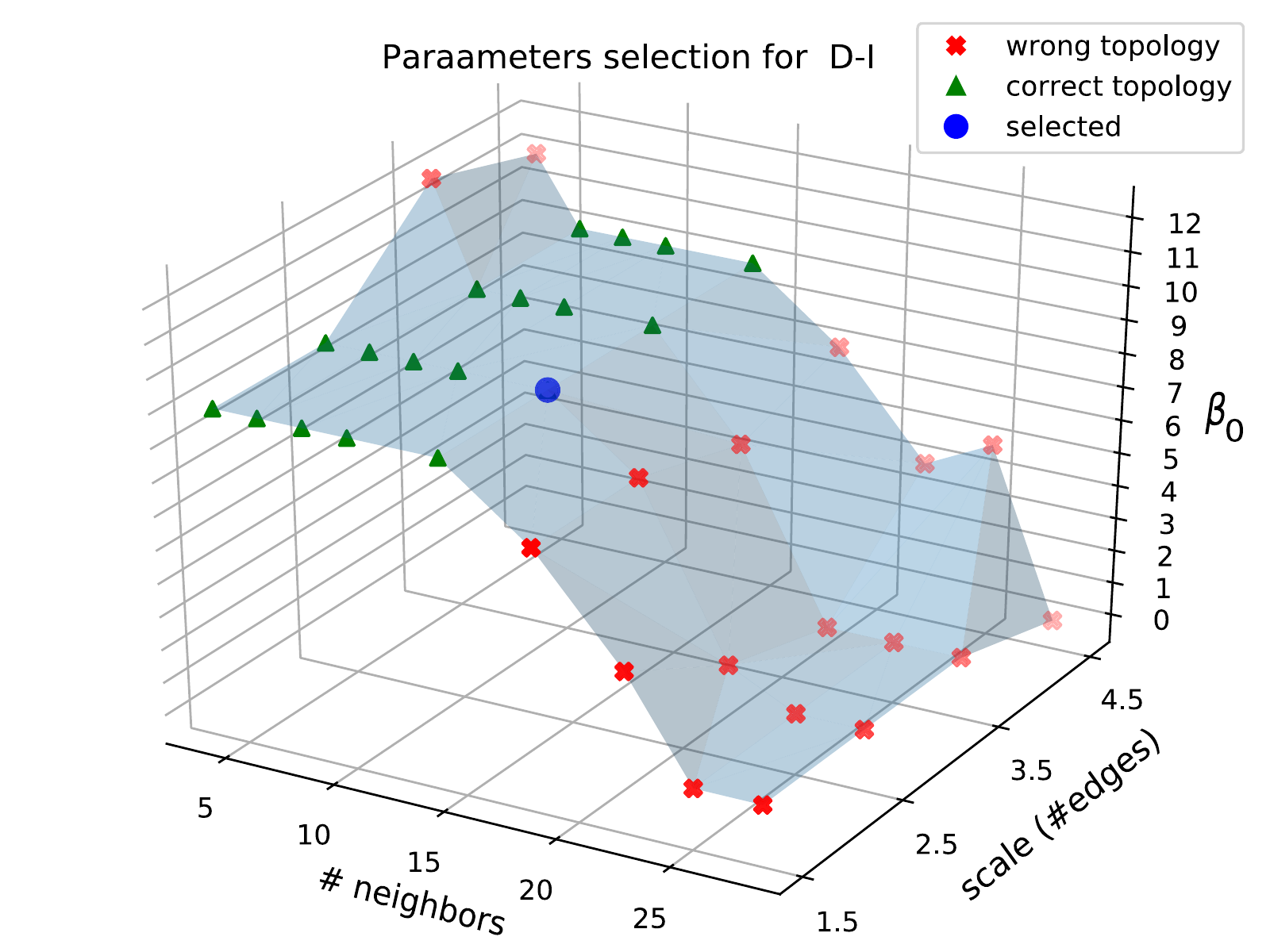}\hfill
\includegraphics[trim={2.0cm 0 0.3cm 0},clip, width=0.48\linewidth]{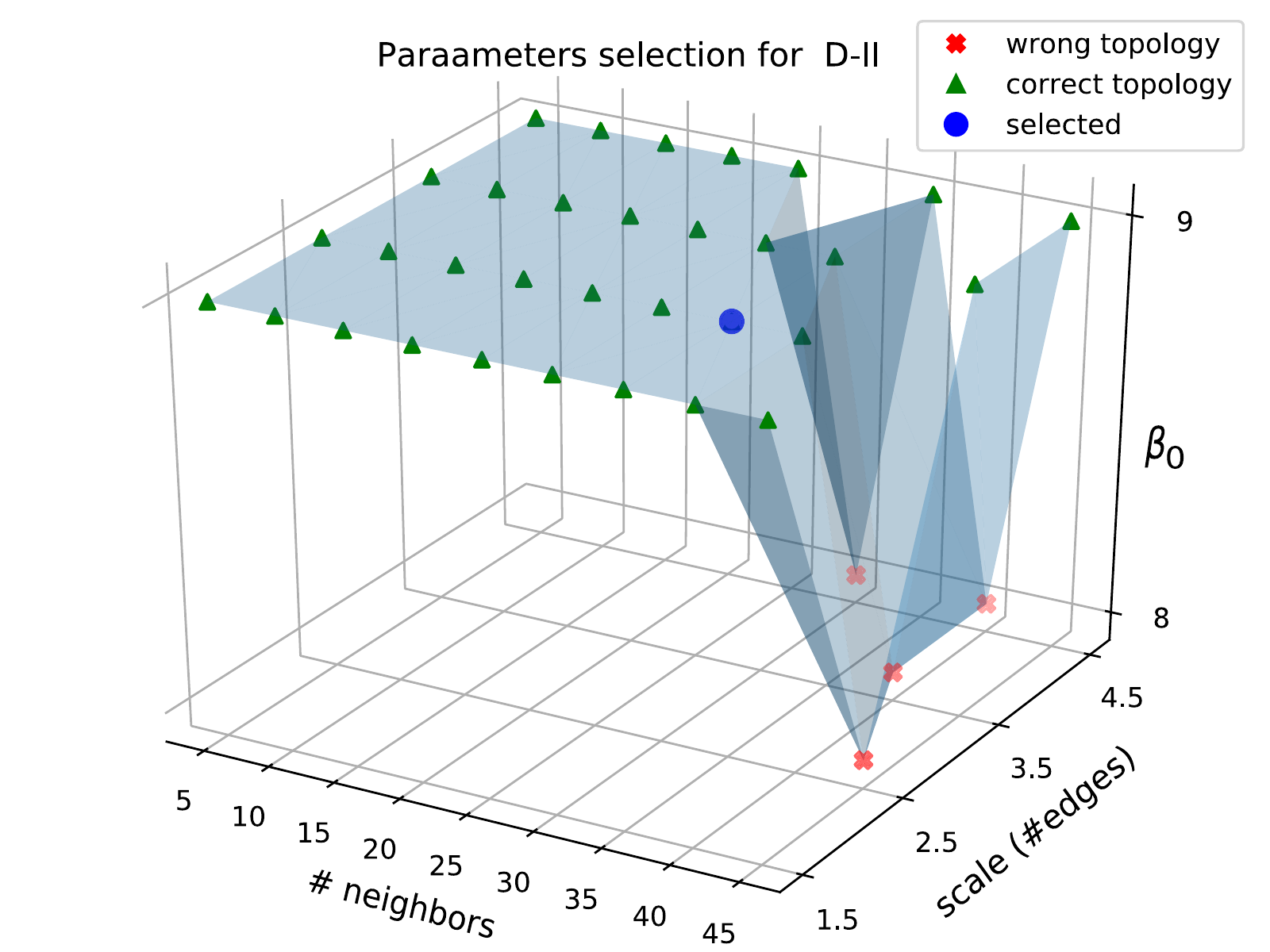}
\includegraphics[trim={2.0cm 0 0.3cm 0},clip,width=0.48\linewidth]{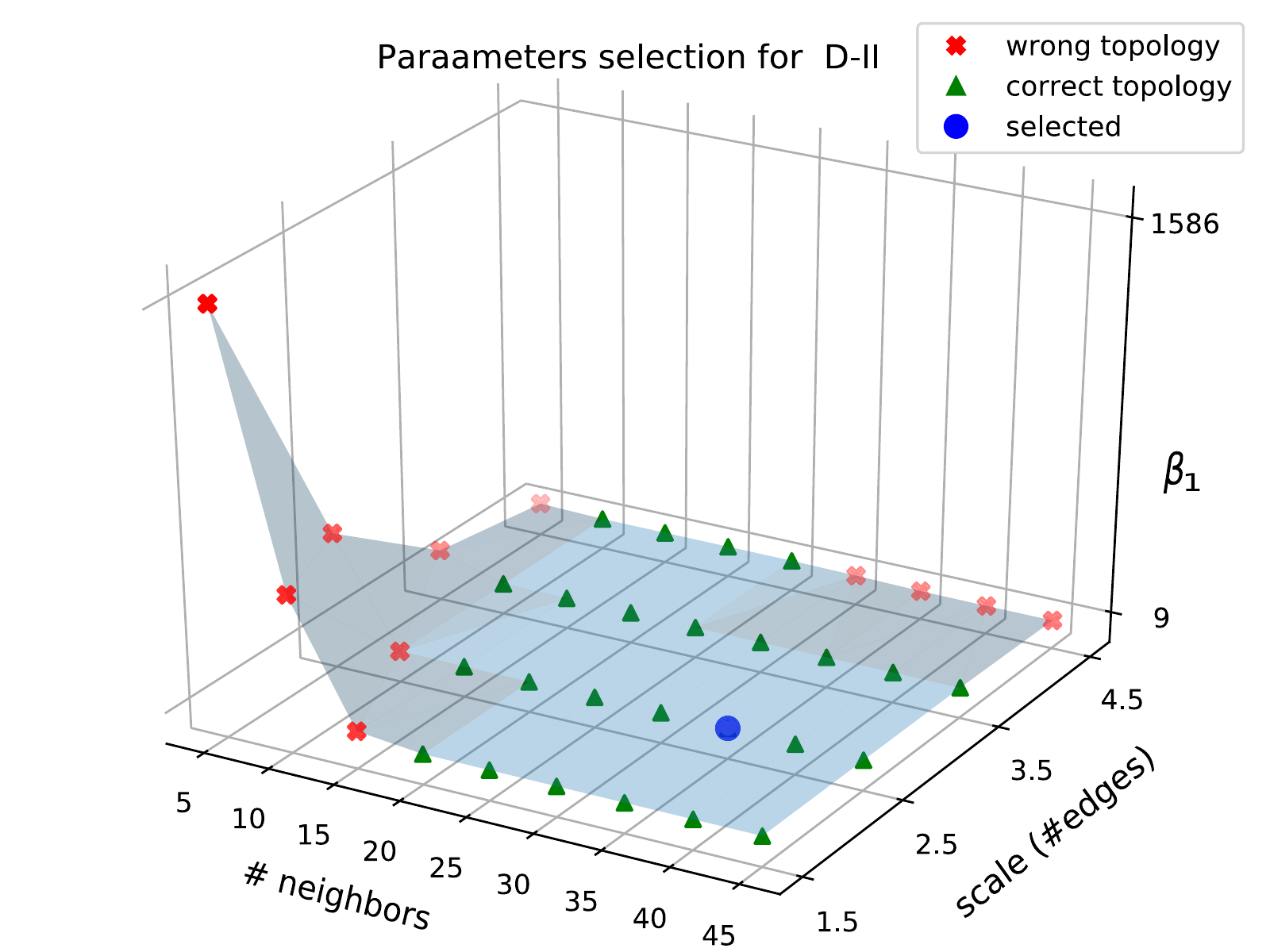}\hfill
\includegraphics[trim={2.0cm 0 0.3cm 0},clip,width=0.48\linewidth]{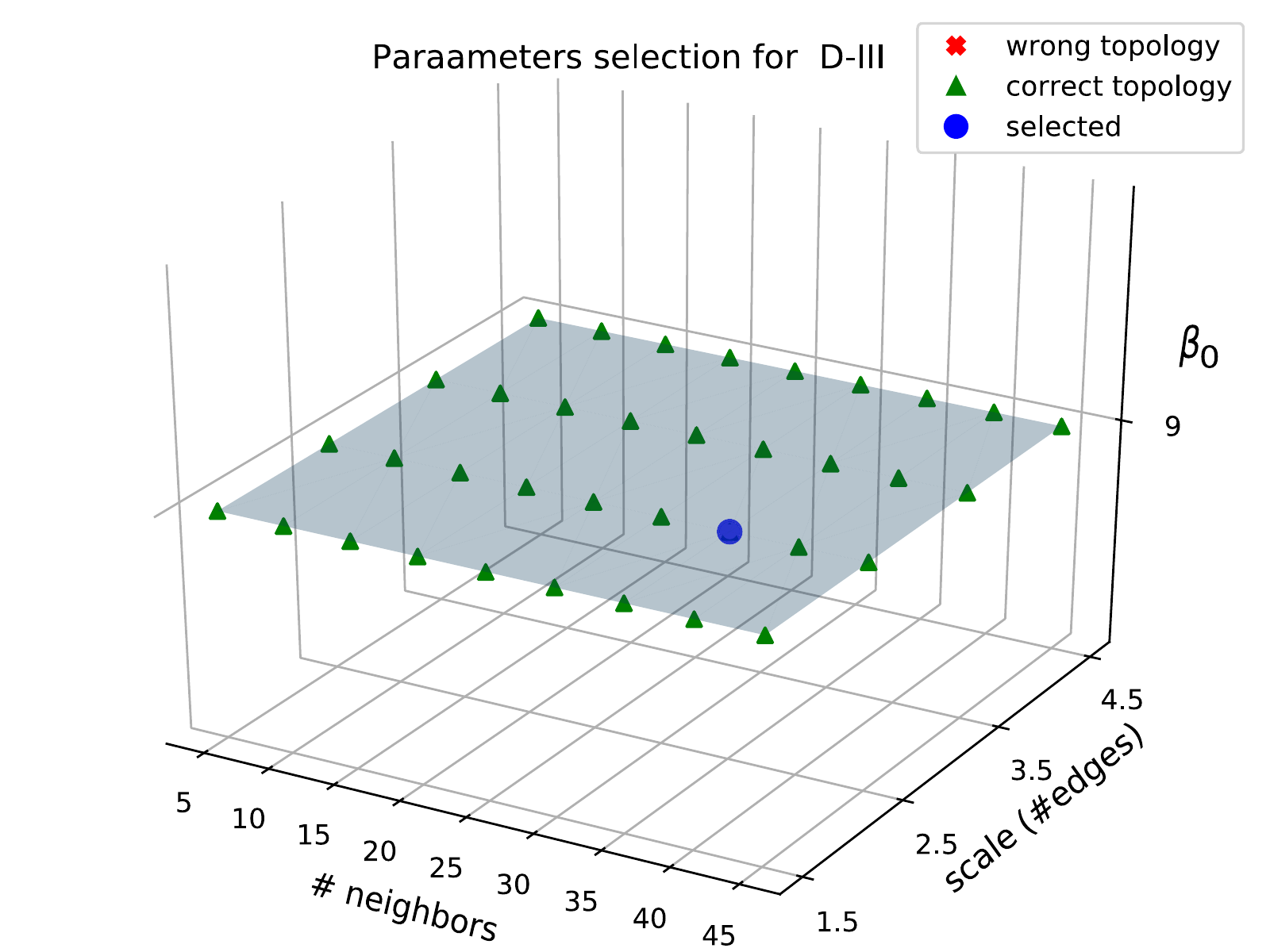}
\includegraphics[trim={2.0cm 0 0.3cm 0},clip,width=0.48\linewidth]{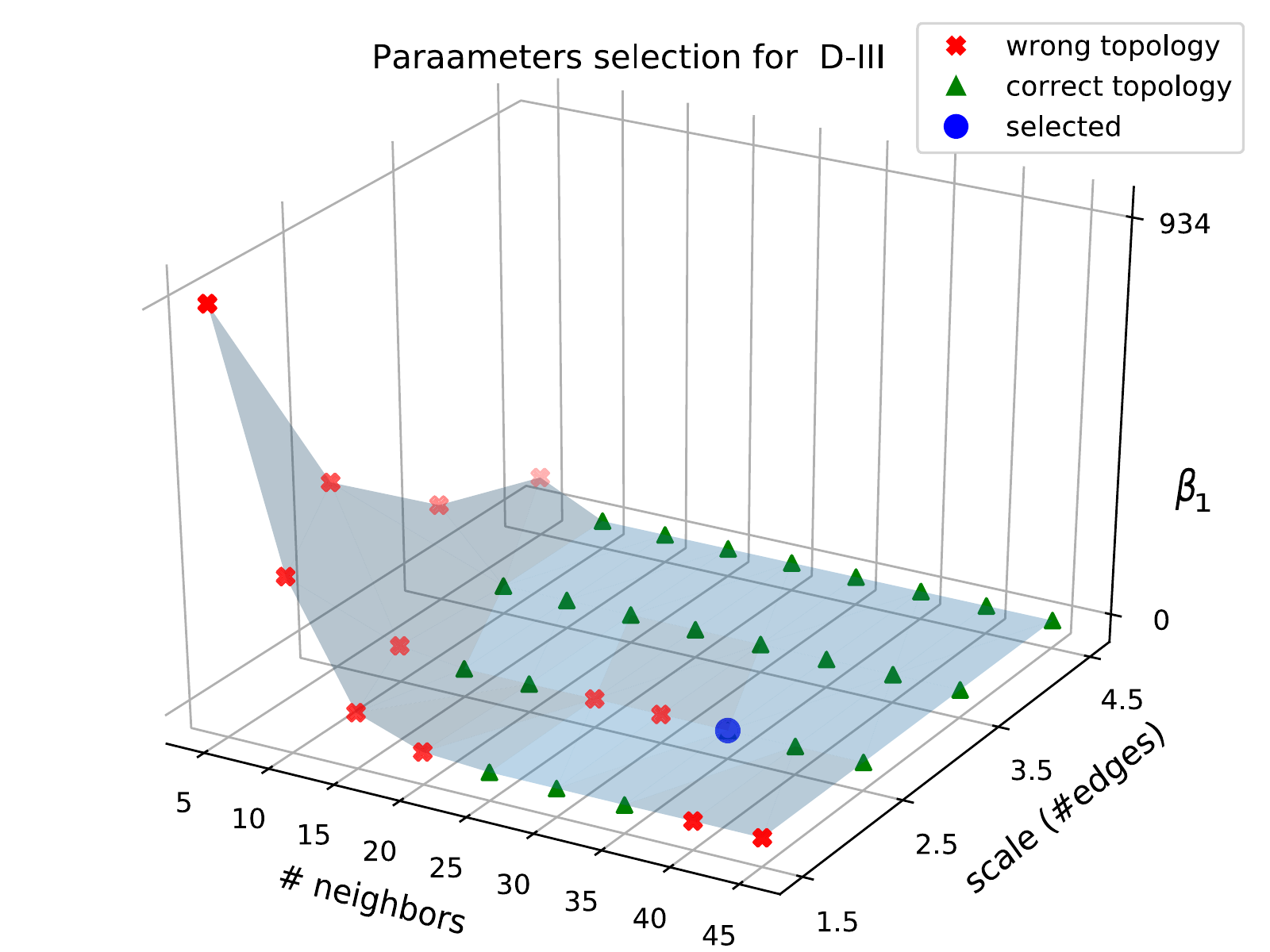}\hfill
\includegraphics[trim={2.0cm 0 0.3cm 0},clip,width=0.48\linewidth]{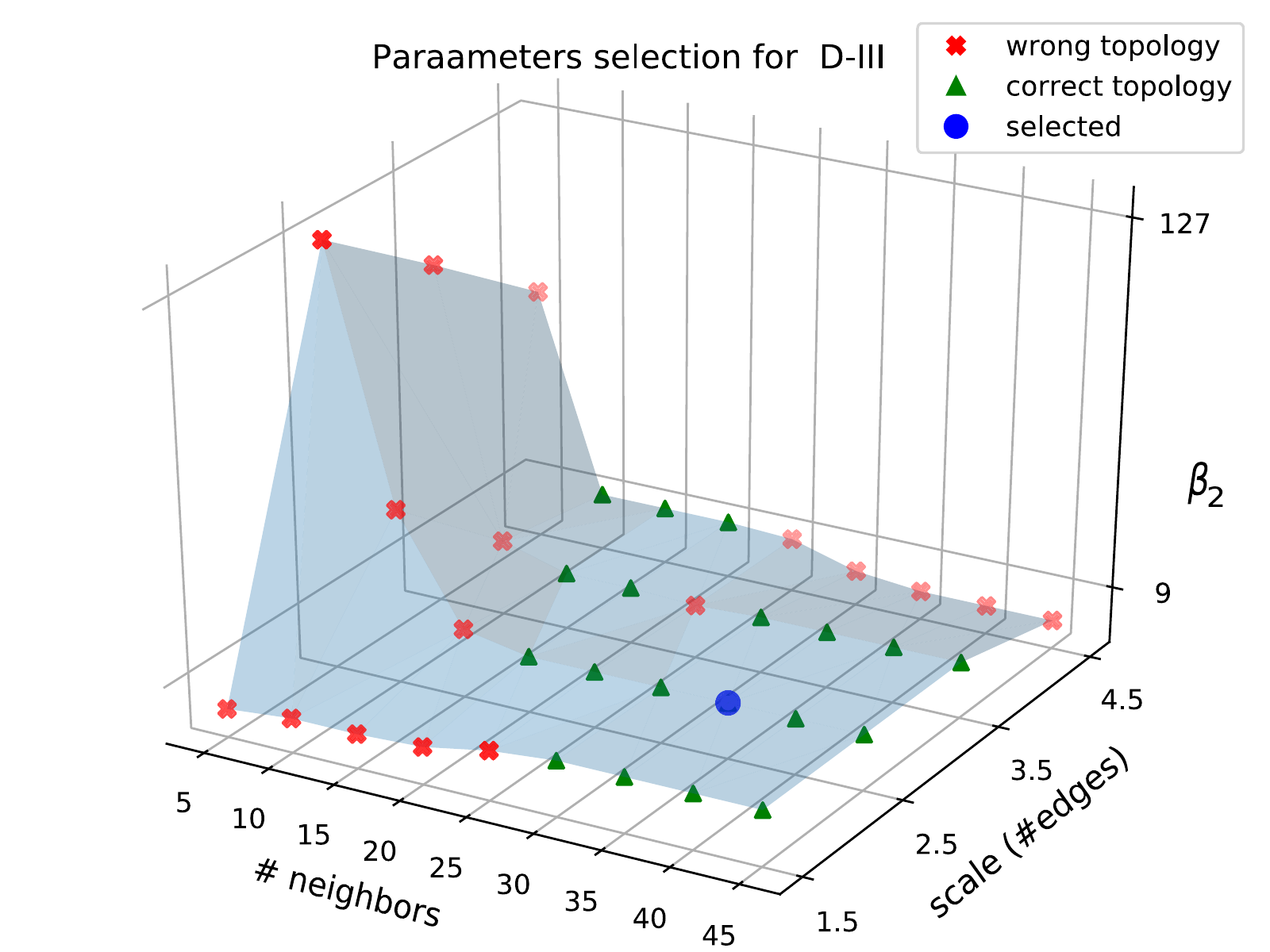}
\caption{For each combination of parameters $k$ and $\varepsilon$, we determine whether the homology of $\VR_{k,\varepsilon}(X)$ matches the homology of the manifold $M$ from which $X$ is sampled. We marked those values of $(k,\varepsilon)$ with correct homology in green and those with incorrect homology in red. Our choice of $(k,\varepsilon)$, naturally chosen among the correct ones, is marked in blue.}
\label{fig:hyperparams_landscape}
\end{figure}

The parameters chosen via the aforementioned procedure for our data sets are as follows. For D-I,  we have $k_* = 14$ neighbors, and  scale is set at $\varepsilon_* = 2.5$; recall that this means that  $x_0,x_1,\dots,x_n \in X$ form an $n$-simplex in $\VR_{14, 2.5}(X)$ whenever $\delta_{14}(x_i, x_j)\leq 2.5$ for all $i,j$. For both D-II and D-III, we have $k_* = 35$ and $\varepsilon_* = 2.5$.  Figure~\ref{fig:hyperparams_landscape} shows the Betti numbers for these three data sets over a range of values of $(k,\varepsilon)$: green (resp.\ red) dots indicate integral points on the $(k,\varepsilon)$-plane with correct (resp.\ incorrect) Betti numbers and the blue dot marks the $(k_*,\varepsilon_*)$ selected in each case.

Our homology and persistent homology computations are all performed using the \textsf{Eirene} package in Julia 0.6.4 \cite{henselmanghristl6}. To give readers an idea, the time required to compute a single Betti number from a point cloud $X$ ranges from a few tens of seconds, if $X\subseteq \mathbb{R}^5$ is the output of a five-neuron-wide layer, to at most 30 minutes, if $X \subseteq \mathbb{R}^{50}$ is the output of a $50$-neuron wide layer. On the other hand, the time taken for the persistent homology computations to obtain  $k_*$ and $\varepsilon_*$ is in excess of 80 minutes. These computations are run in parallel over 12 cores of an Intel i7-8750H-2.20GHz processor with 9,216KB cache and 32GB DDR4-2666MHz RAM. The jobs are fed in a queue, with a single core limited to 9GB of memory.

\subsection{Overview of our experiments}\label{sec:experiments}

All our experiments on simulated data (those on real data omits the first step) may be described at a high level as follows: (i) generate a manifold $M = M_a \cup M_b \subseteq \mathbb{R}^d$ with $M_a \cap M_b =\varnothing$; (ii) densely sample point cloud $X \subseteq M$ and let $X_i \coloneqq X \cap M_i$, $i =a,b$; (iii) train $l$-layer neural network  $\nu  : \mathbb{R}^d \to [0,1]$ on the labeled training set $X_a \cup X_b$ to classify points on $M$; (iv) compute homology of the output at the $j$th layer $\nu_j(X_i)$, $j =0,1,\dots,l,l+1$ and $i =a,b$. This allows us to track how the topology of $M_i$, by way of (persistent) homology of $X_i$, as it passes through the layers. Steps (i) and (ii) are described in Section~\ref{sec:data}, step (iii) in Section~\ref{sec:nn}, and step (iv) in Section~\ref{sec:comhom}. The neural network notations are as in Section~\ref{sec:problem}. Results will be described in Section~\ref{sec:results}.

In reality, the point cloud $X_i$ used in step (iv) is not the same as the training set $X_i$ used in step (iii) as it is considerably more expensive to compute homology (see Section~\ref{sec:comhom})  than to train a neural network to near zero generalization error (see Section~\ref{sec:nn}). As such the size of point clouds used for our homology computations are a fraction (about 1/4) that used to train our neural networks.
\enlargethispage{2\baselineskip}
\begin{table}[h]
\centering\footnotesize
\begin{tabular}[b]{lrr}
\hline
\hline
data set & training neural networks & homology computations\\
\hline
D-I & 7,800  &  2,600\\
D-II & 45,000 &  11,250 \\
D-III & 37,800 & 9,450 \\
\hline
\hline
\end{tabular}
\medskip
\caption{Comparison of sample sizes for computations in Sections~\ref{sec:nn} and \ref{sec:comhom}.}
\label{table:datasetsize}
\end{table}

\section{Results and discussions}\label{sec:results}

We present the results from our experiments to analyze how a well-trained neural network simplifies the topology of a data set. More importantly, we discuss what one may surmise from these results. We start with the three simulated data sets D-I, D-II, D-III in Section~\ref{sec:data} since the results are the most striking in this case. To validate that these observations indeed extend to real data, we repeat our experiments on four real-world data sets in Section~\ref{sec:real_data}.

\enlargethispage{\baselineskip}
\begin{figure}[H]
\begin{minipage}[h]{1.\textwidth}
\centering
\hspace{-1.5ex}
{\includegraphics[scale=0.31]{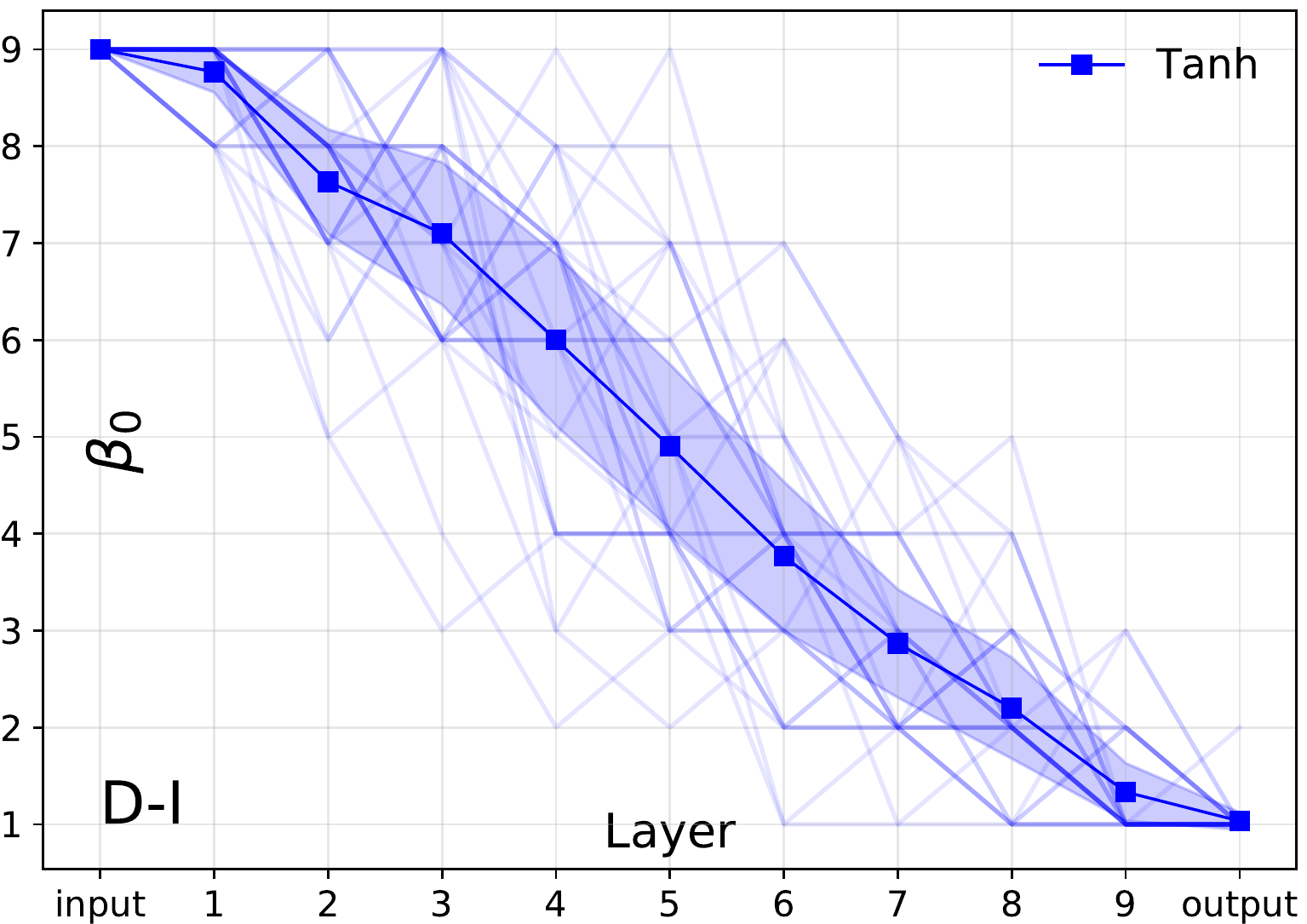}}
\hspace{0ex}
{\includegraphics[trim={0mm 0 0 0}, clip, scale=0.31]{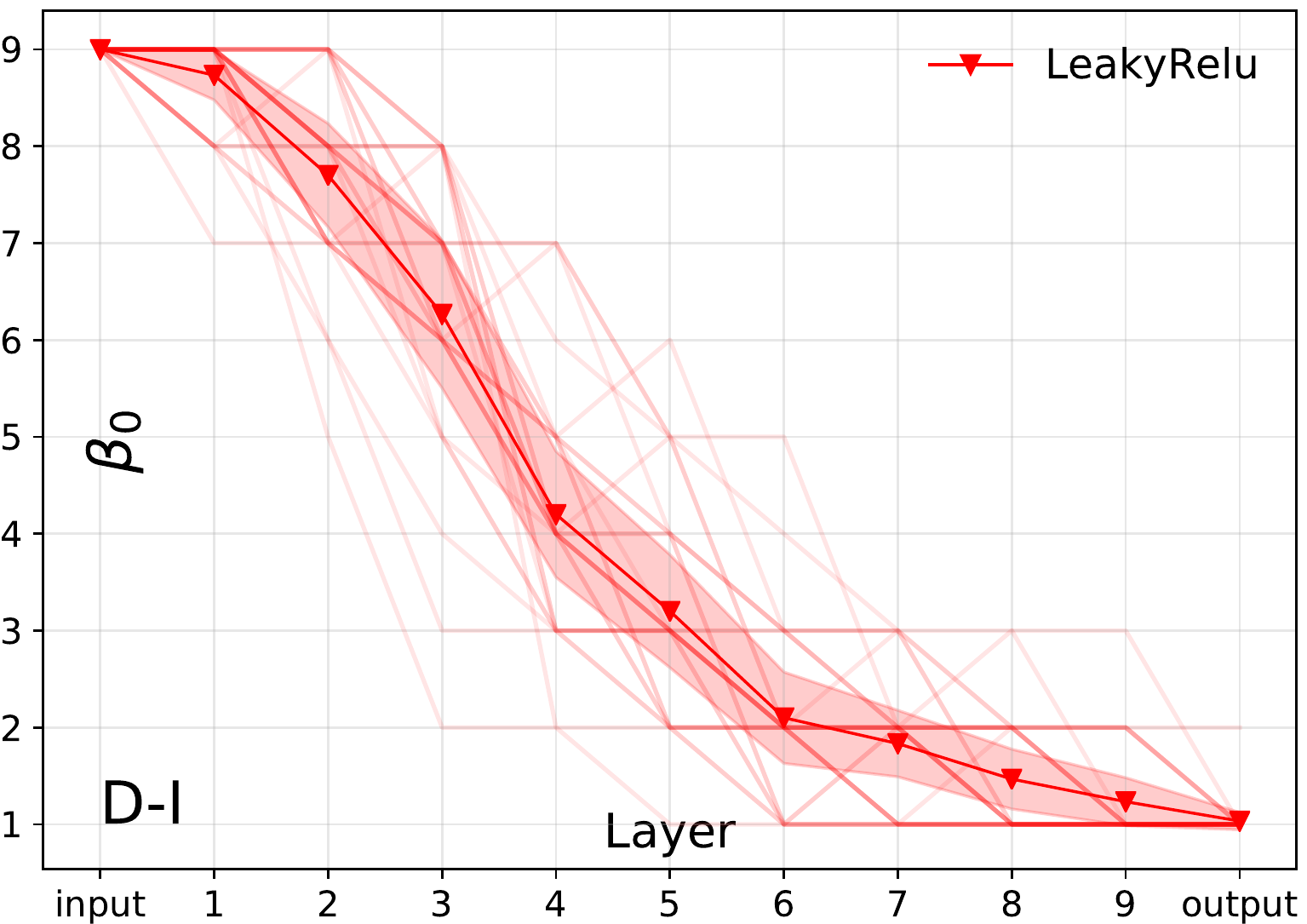}}
\hspace{0ex}
{\includegraphics[trim={0mm 0 0 0}, clip, scale=0.31]{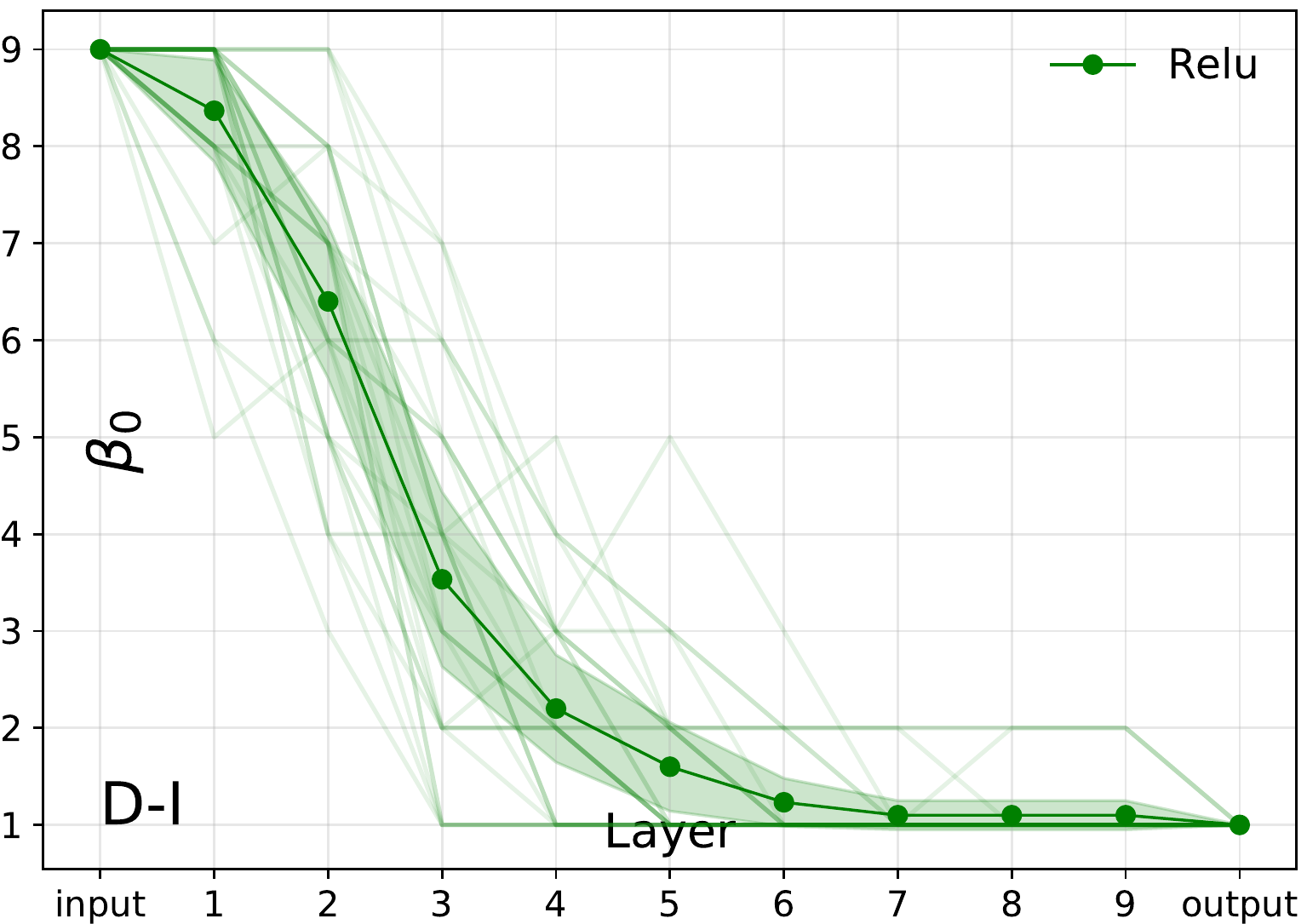}}
\end{minipage}
\begin{minipage}[c]{1.\textwidth}
\centering 
\includegraphics[width=0.96\textwidth]{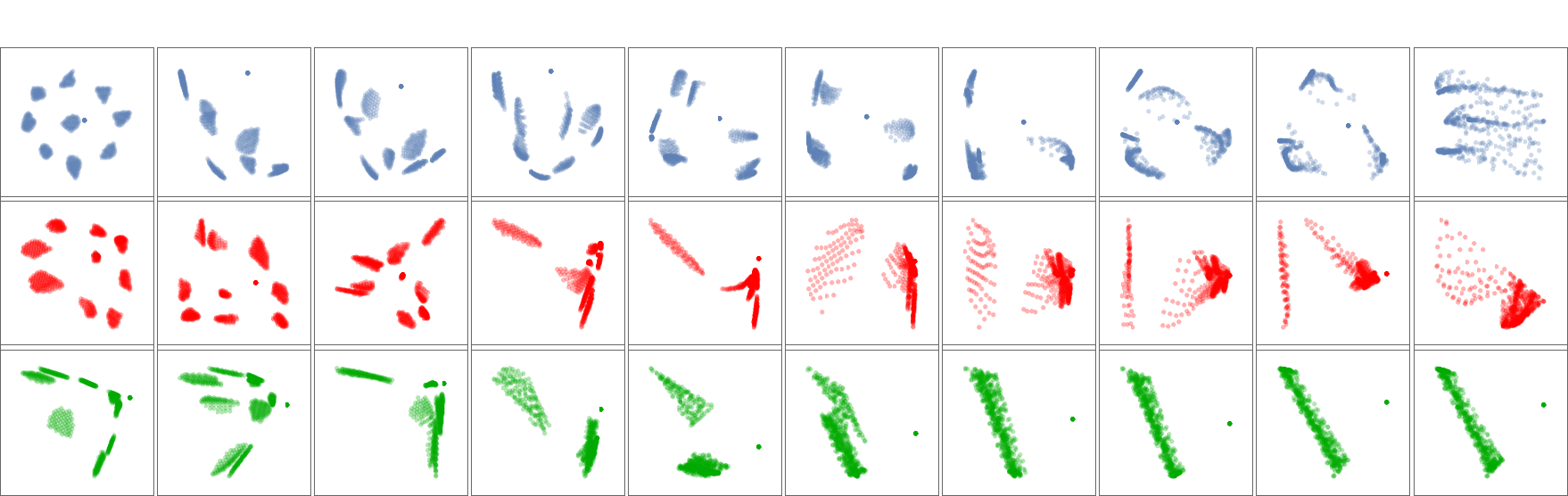}
\end{minipage}
\caption{\emph{Top}: Faint curves show individual profiles, dark curves show averaged profiles of $\beta_0\bigl(\nu_k(M_a)\bigr)$, $k=1, \dots, 10$, in data set D-I.  Shaded region is the region of $\pm$ half standard deviation about average curve.  Networks have different activations --- blue for $\tanh$, red for leaky ReLU, green for ReLU; but same architecture --- ten layers, two neurons in the first and the last layers, fifteen in the intervening layers. \emph{Bottom}: Projections of $\nu_k(M_a)$, $k=1, \dots, 10$, on the first two principal components, color-coded according to activations.} \label{fig:2d_dataset}
\end{figure}
\subsection{Topological simplification evident across training instances.}
\begin{figure}[h!]
\begin{minipage}[c]{1.\textwidth}
\centering
{\includegraphics[scale=0.45]{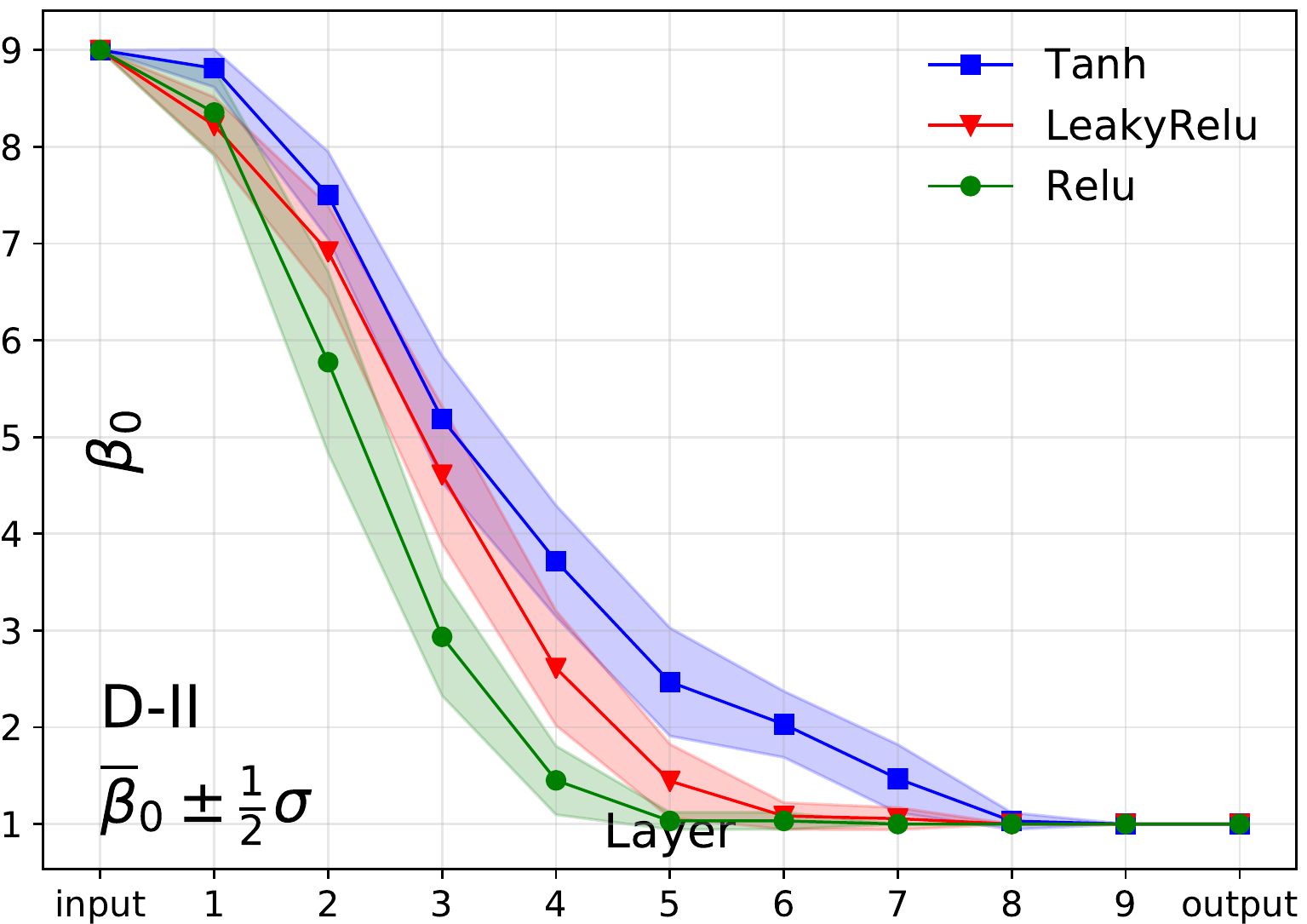}}
\hspace{5ex}
{\includegraphics[scale=0.45]{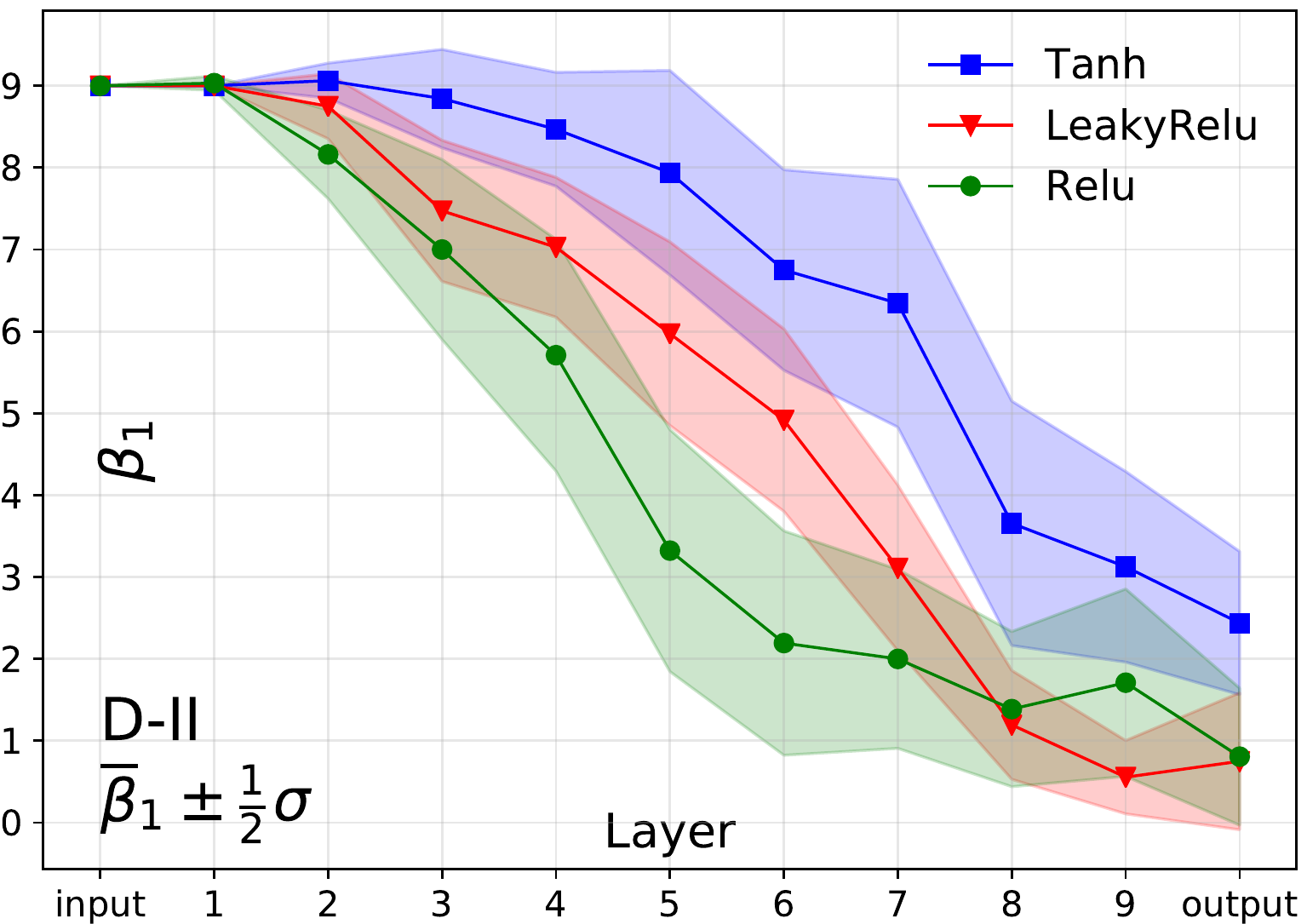}}
{\includegraphics[width=0.96\textwidth]{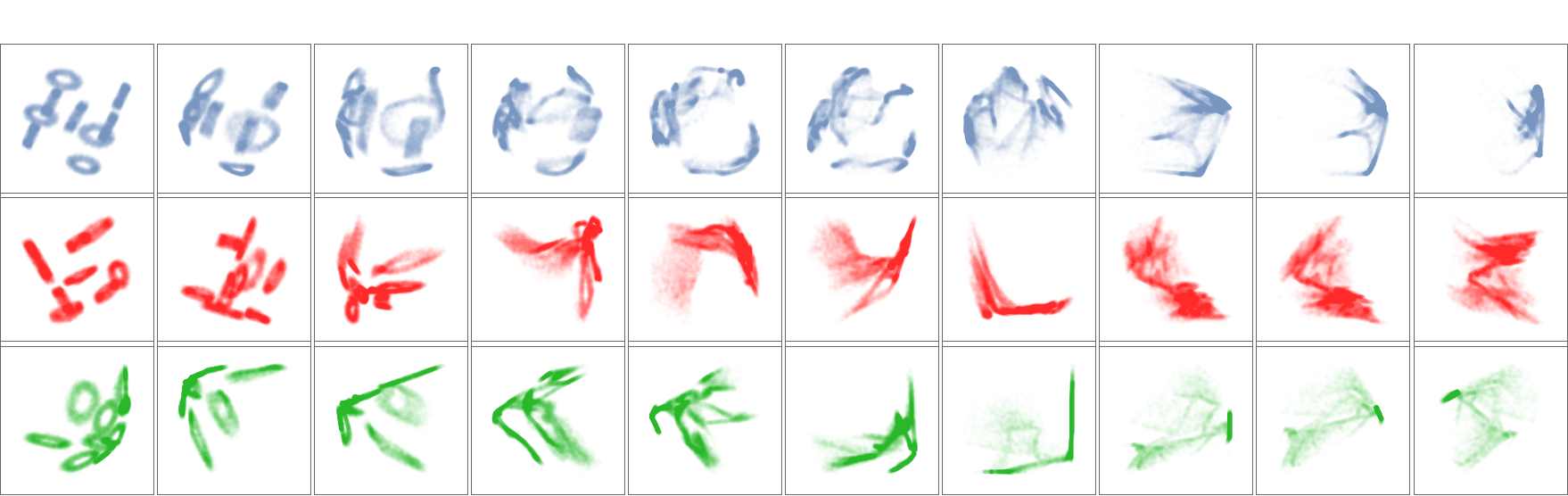}}
\end{minipage}
\caption{\emph{Top:} Profiles of $\beta_0\bigl(\nu_k(M_a)\bigr)$ and $\beta_1\bigl(\nu_k(M_a)\bigr)$ for data set D-II, $k =1,\dots,10$. Network's architecture: three-dimensional input, two-dimensional output, and fifteen neurons in intervening layers one to nine, with different activations.  \emph{Bottom}:  Projections of $\nu_k(M_a)$, $k=1, \dots, 10$, on the first two principal components.}
\label{fig:dataset_ii}
\end{figure}

Figure~\ref{fig:2d_dataset} records our simplest data set D-I, where  $M_a$ comprises nine contractible components and so higher Betti numbers are irrelevant (all zero). Here we present every curve corresponding to every neural network trained on D-I, recall that we do at least 30 runs for each experiment to account for the inherent randomness, and they all show consistent profiles --- a clear decay in $\beta_0$ across the layers although hyperbolic tangent activation (blue graph) shows larger variance in this decay than leaky ReLU (red graph) and ReLU (green graph).
The profiles shown in Figure~\ref{fig:2d_dataset} are representative of other experiments on higher Betti numbers and on other data sets. To avoid clutter, in the corresponding figures for D-II and D-III (Figures~\ref{fig:dataset_ii} and \ref{fig:dataset_iii}), we omit curves corresponding to the individual runs and show only the curve of their means (dark curve in middle) and the region of half standard deviation (shaded region). The bottom diagrams in Figure~\ref{fig:2d_dataset} show how $M_a$ changes from layer-to-layer by projecting onto its first two principal components (note that the intervening layers are in $\mathbb{R}^{15}$).

\subsection{Nonhomeomorphic activations induce rapid topology changes.}
\begin{figure}[h]
\begin{minipage}[c]{1.\textwidth}
\centering
{\includegraphics[scale=0.45]{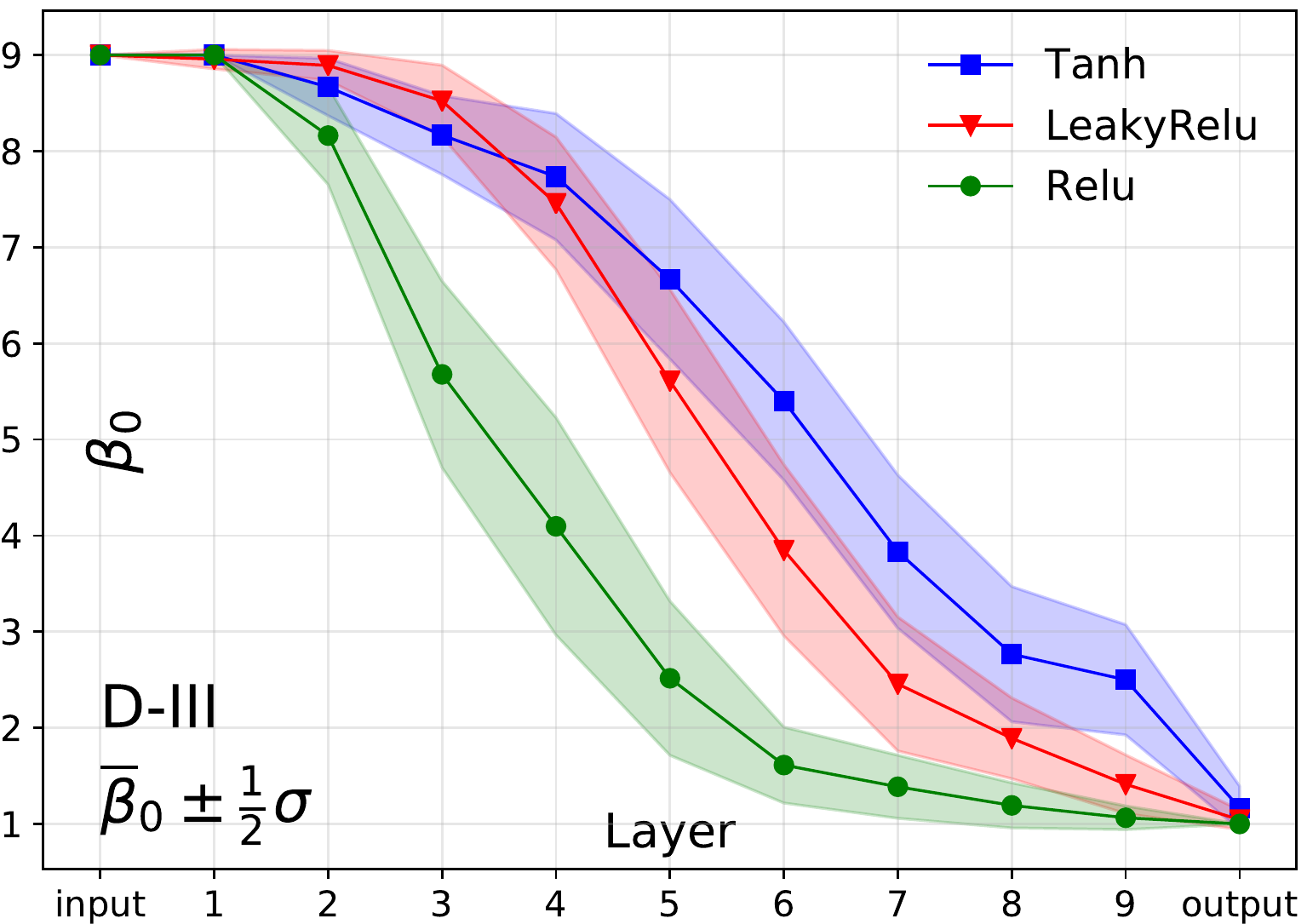}}
\hspace{5ex}
{\includegraphics[scale=0.45]{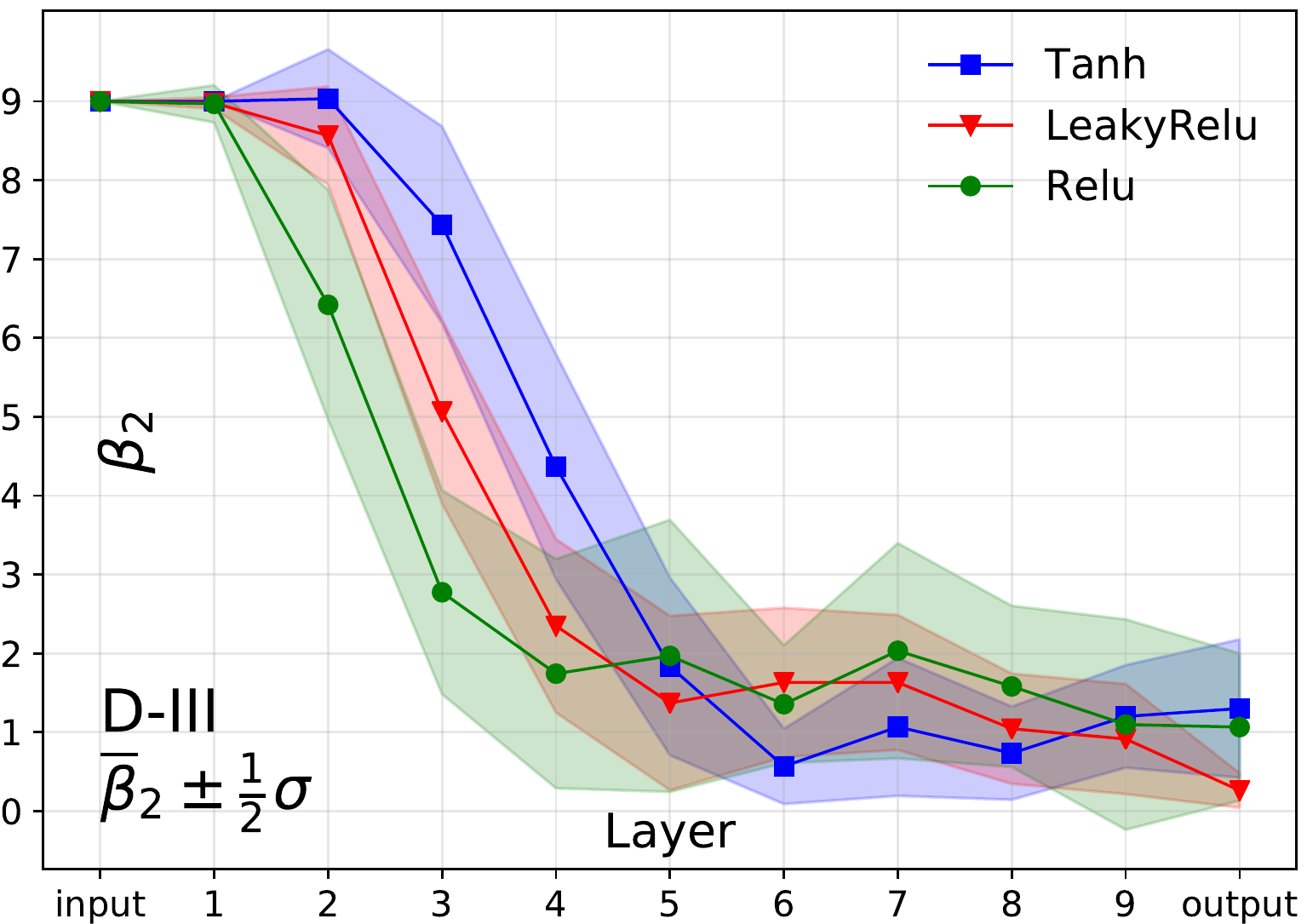}}
{\includegraphics[width=0.96\textwidth]{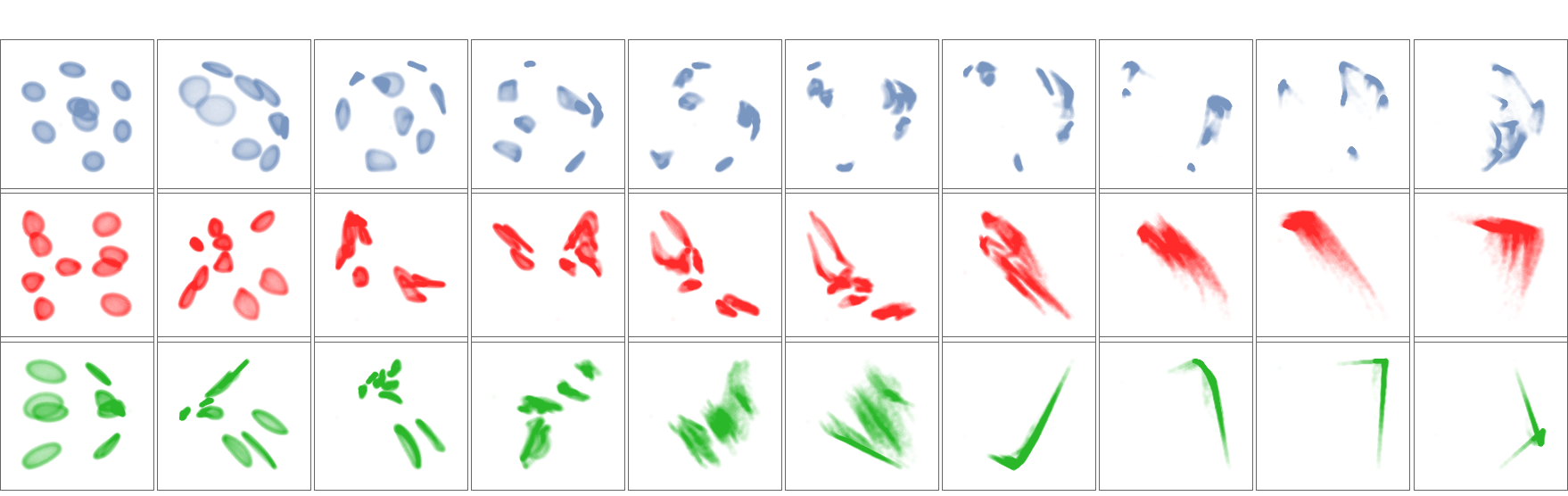}}
\end{minipage}
\caption{\emph{Top:} Profiles of $\beta_0\bigl(\nu_k(M_a)\bigr)$ and $\beta_2\bigl(\nu_k(M_a)\bigr)$ for data set D-III, $k =1,\dots,10$. Network's architecture: three-dimensional input, two-dimensional output, and fifteen neurons in intervening layers one to nine, with different activations.  \emph{Bottom}:  Projections of $\nu_k(M_a)$, $k=1, \dots, 10$, on first two principal components.}
\label{fig:dataset_iii}
\end{figure}

As the faint blue lines in Figure~\ref{fig:2d_dataset} reveal, hyperbolic tangent activation is less effective at reducing Betti numbers, occasionally even increasing them over layers. In all data sets, across all our experiments, the nonhomeomorphic  activation ReLU exhibits the most rapid reductions in all Betti numbers. The top halves of  Figures~\ref{fig:2d_dataset}, \ref{fig:dataset_ii} and \ref{fig:dataset_iii} show results for  a $10$-layer network (see caption for specifics). The different rates at which topological changes occur are also evident from the principal components projections in the bottom half of these figures.

\subsection{Efforts depend on topological features.}
Some topological features evidently require more layers to simplify than others. The hardest one is the interlocking tori in the data set D-II. The profile of $\beta_1\bigl(\nu_l(M_a)\bigr)$ in the graph on the right of Figure~\ref{fig:dataset_ii} shows that some loops survive across many layers, especially so when activated with hyperbolic tangent  (blue): both the (blue) principal components projections and the (blue) profile show that the loops persist considerably longer than any other features in any of the three data sets.

\subsection{Effects of width on topology change.}
\begin{figure}[h]
\centering
\includegraphics[trim={0 0 0 1.5}, clip, scale=0.31]{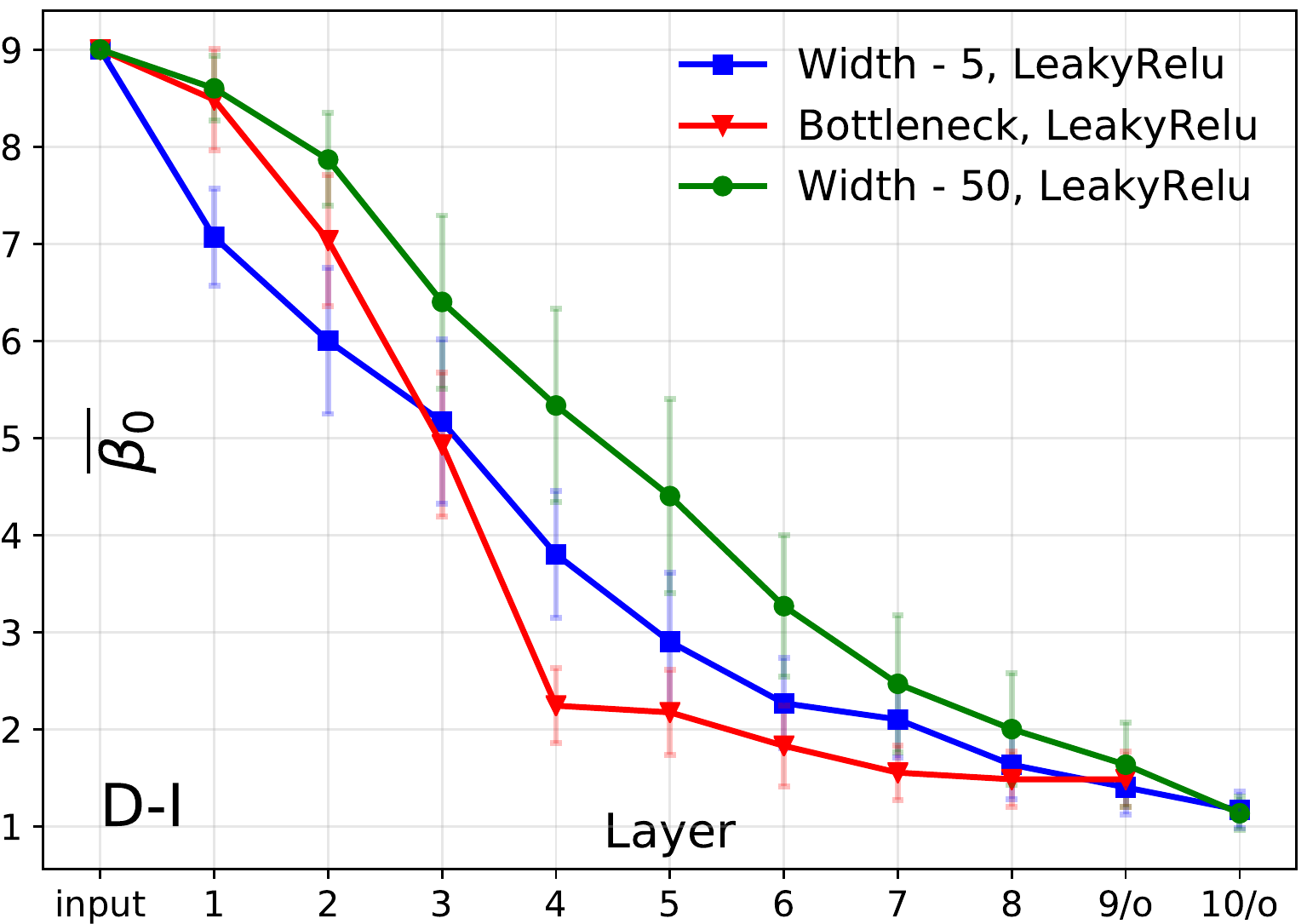}
\hspace{-1.0ex}
\includegraphics[trim={0 0 0 1.5}, clip, scale=0.31]{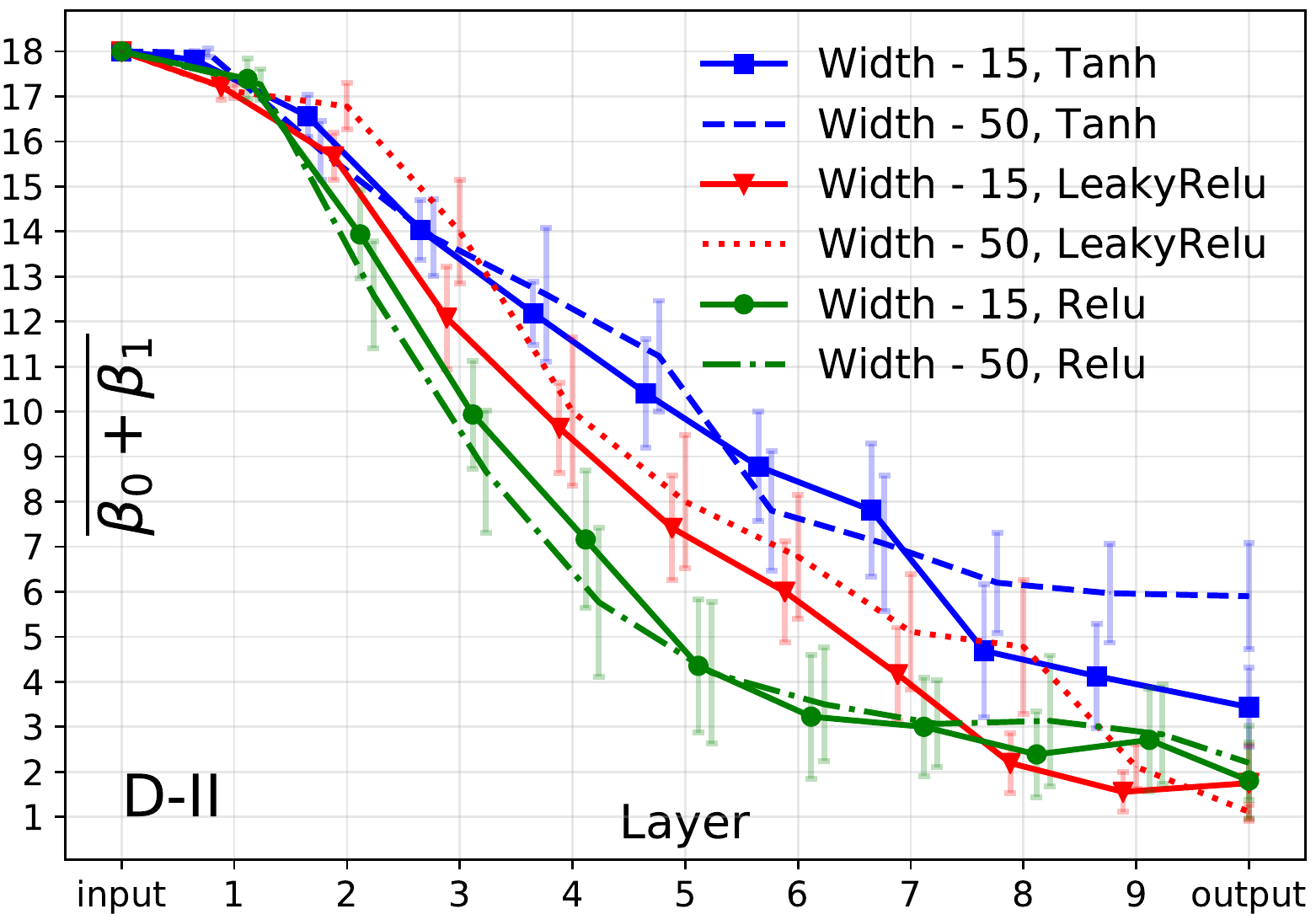}
\hspace{-1.0ex}
\includegraphics[trim={0 0 0 1.5}, clip, scale=0.31]{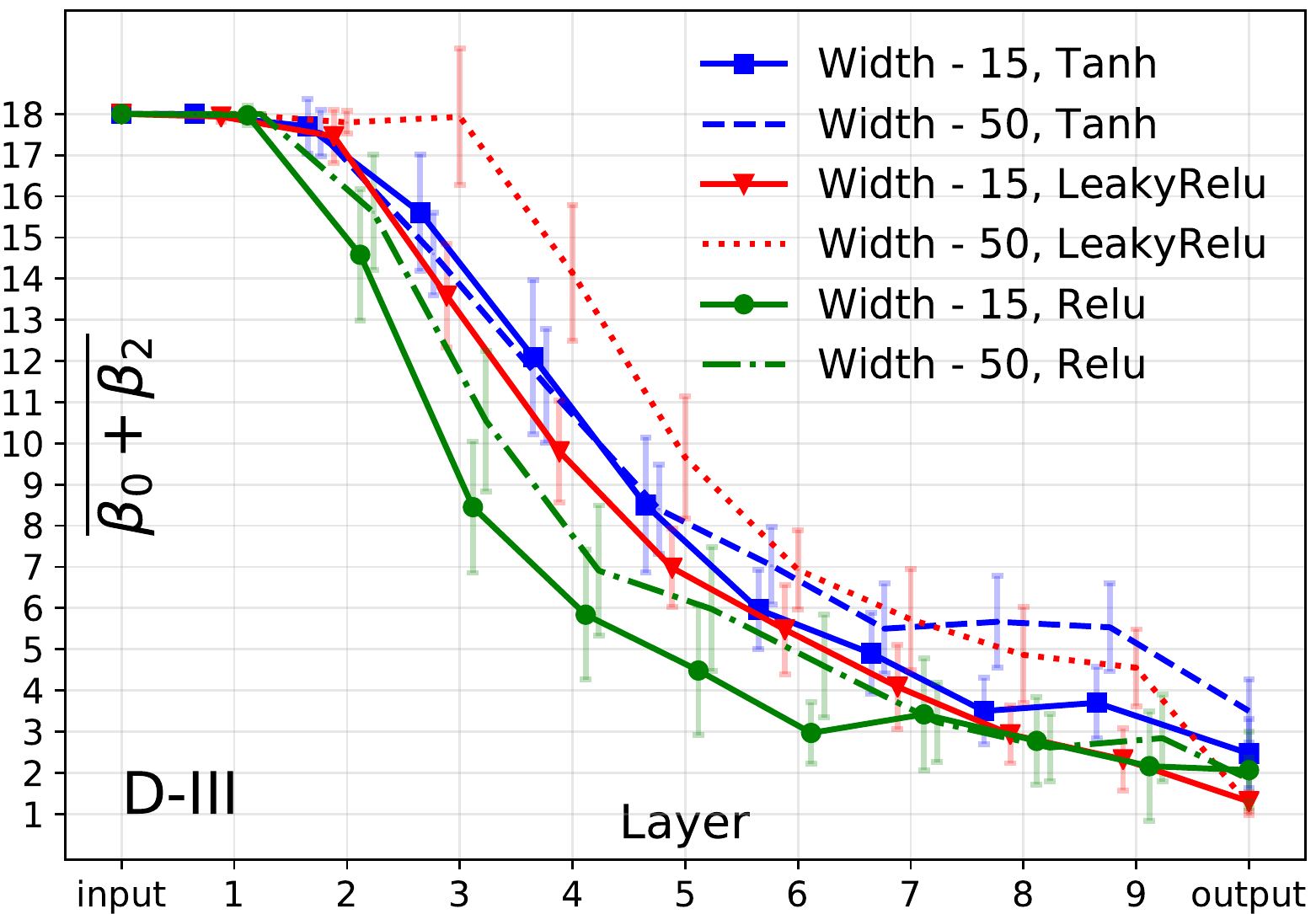}
\caption{Mean values of topological complexity $\TC\bigl(\nu_k(M_a)\bigr)$, $k=1, \dots, l$, for ten-layer deep networks of varying widths. Error bars indicate $\pm$ half standard deviation about the mean.}\label{fig:width}
\end{figure}

For the data set D-I, we compare three sets of ten-layer networks: (i) narrow networks with six neurons in each layer; (ii) `bottleneck' networks with 15, 15, 15, 15, 3, 15, 15, 15, 15 neurons respectively in layers one through nine --- notice the three neuron bottleneck layer; (iii) wide networks with fifty neurons in each layer. The left graph in Figure~\ref{fig:width} suggests that a bottleneck layer forces large topological changes, and a narrow network changes topology faster than a wider one. The other two graphs compare a 15-neuron wide network with a $50$-neuron wide one, both with ten layers, on data sets D-II and D-III  respectively. However, for the same choice of activation, the difference between them is negligible. Also, reducing the width below fifteen neurons makes training to high accuracy increasing more difficult, i.e., the percentage of successfully trained networks starts to drop.

\subsection{Effects of depth on topology change.}
\begin{figure}[h]
\centering
\hspace{-1.0ex}
\vspace{0.3cm}
\includegraphics[scale=0.32]{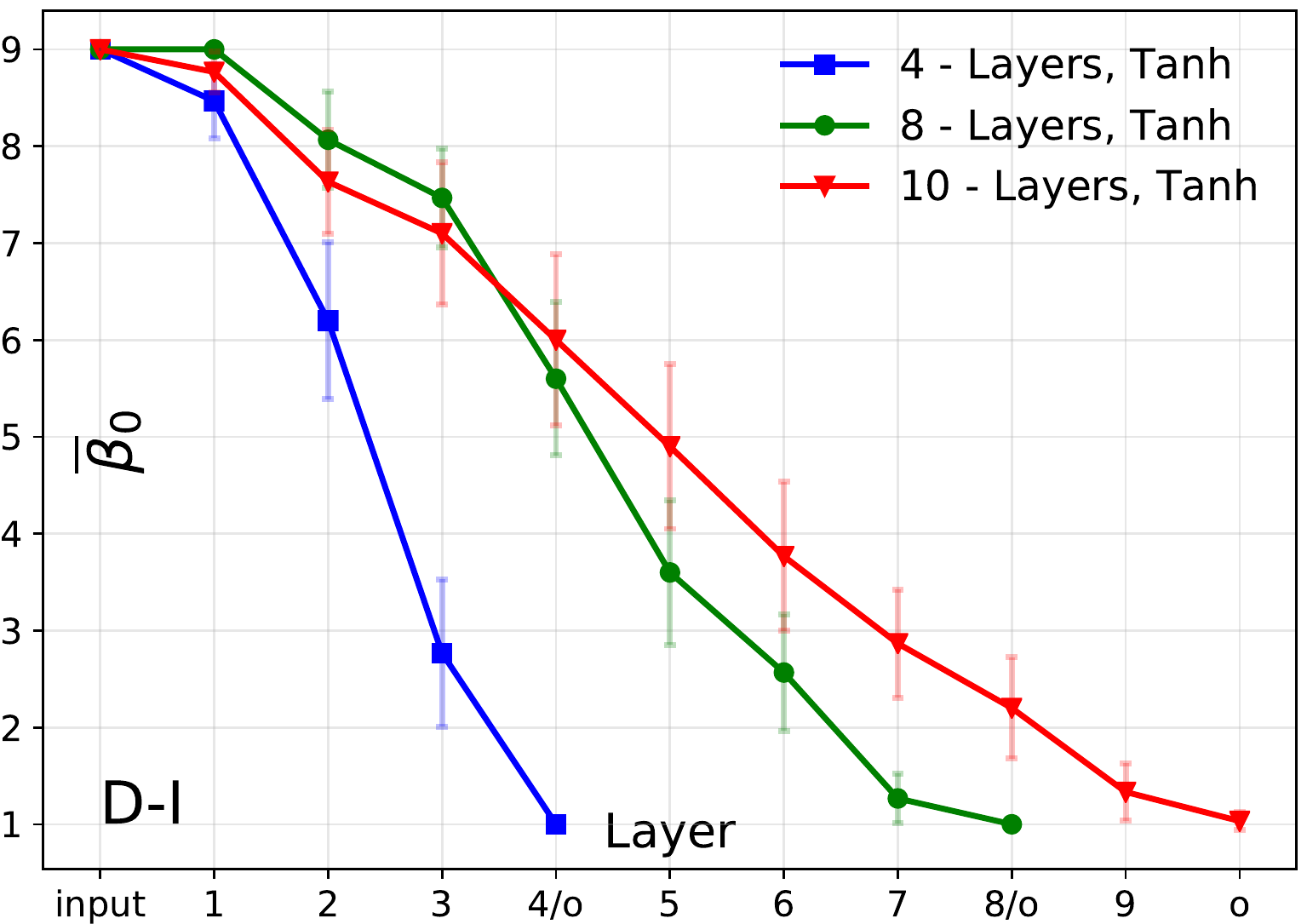}
\hspace{-1.3ex}
\includegraphics[trim={7mm 0 0 0}, clip, scale=0.32]{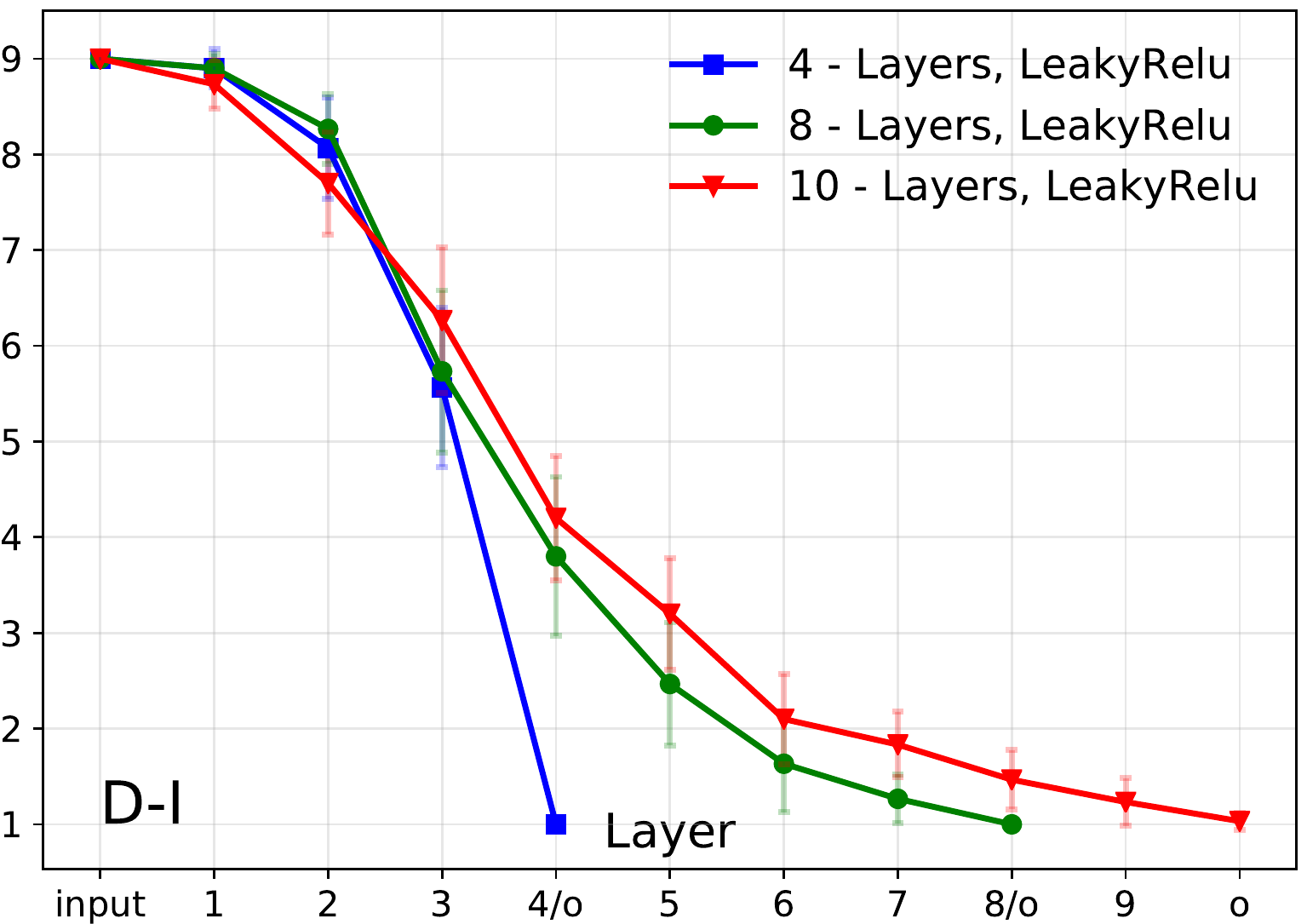}
\hspace{-1.3ex}
\includegraphics[trim={7mm 0 0 0}, clip, scale=0.32]{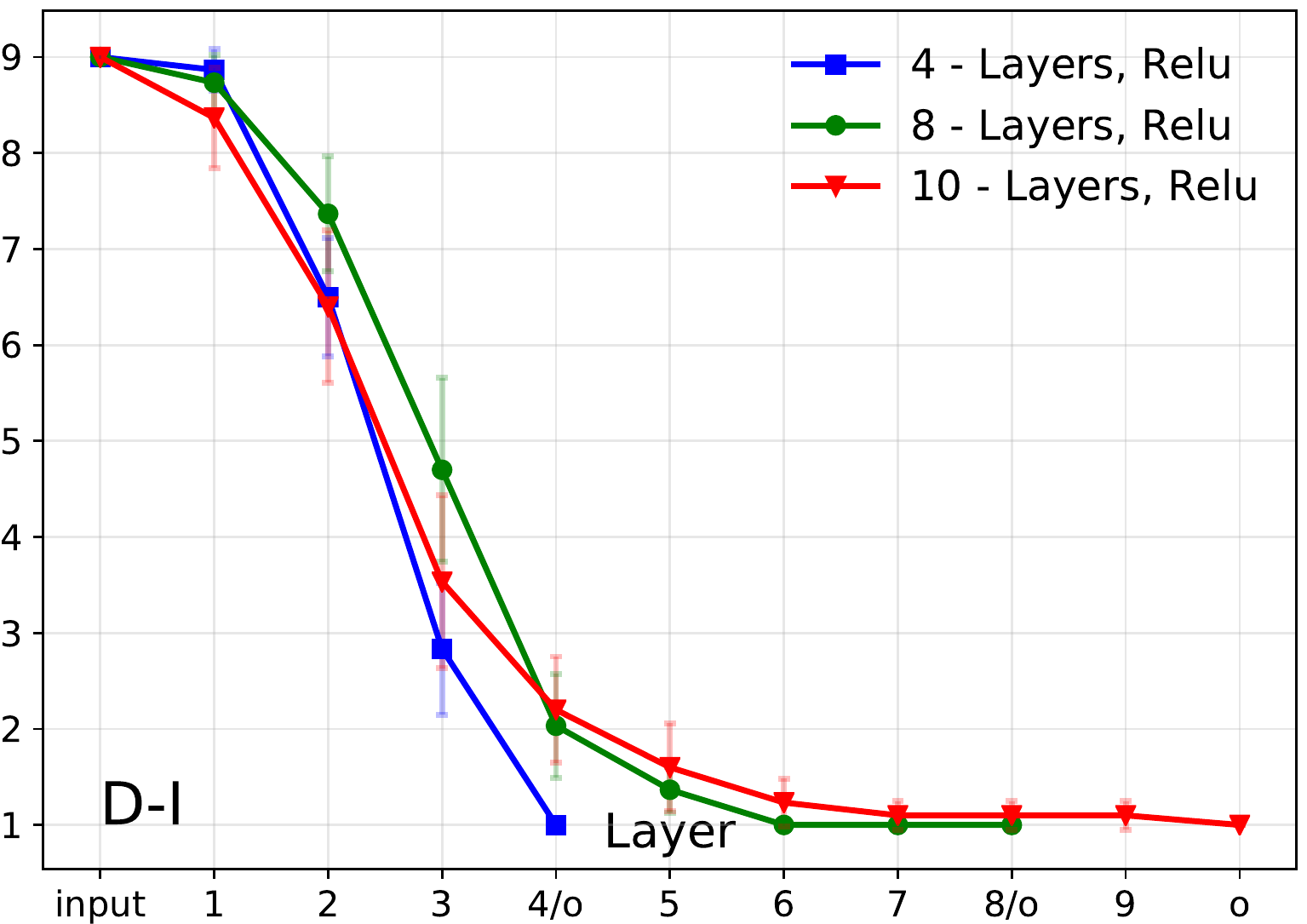}
\hspace{-1.5ex}
\vspace{0.3cm}
\includegraphics[scale=0.32]{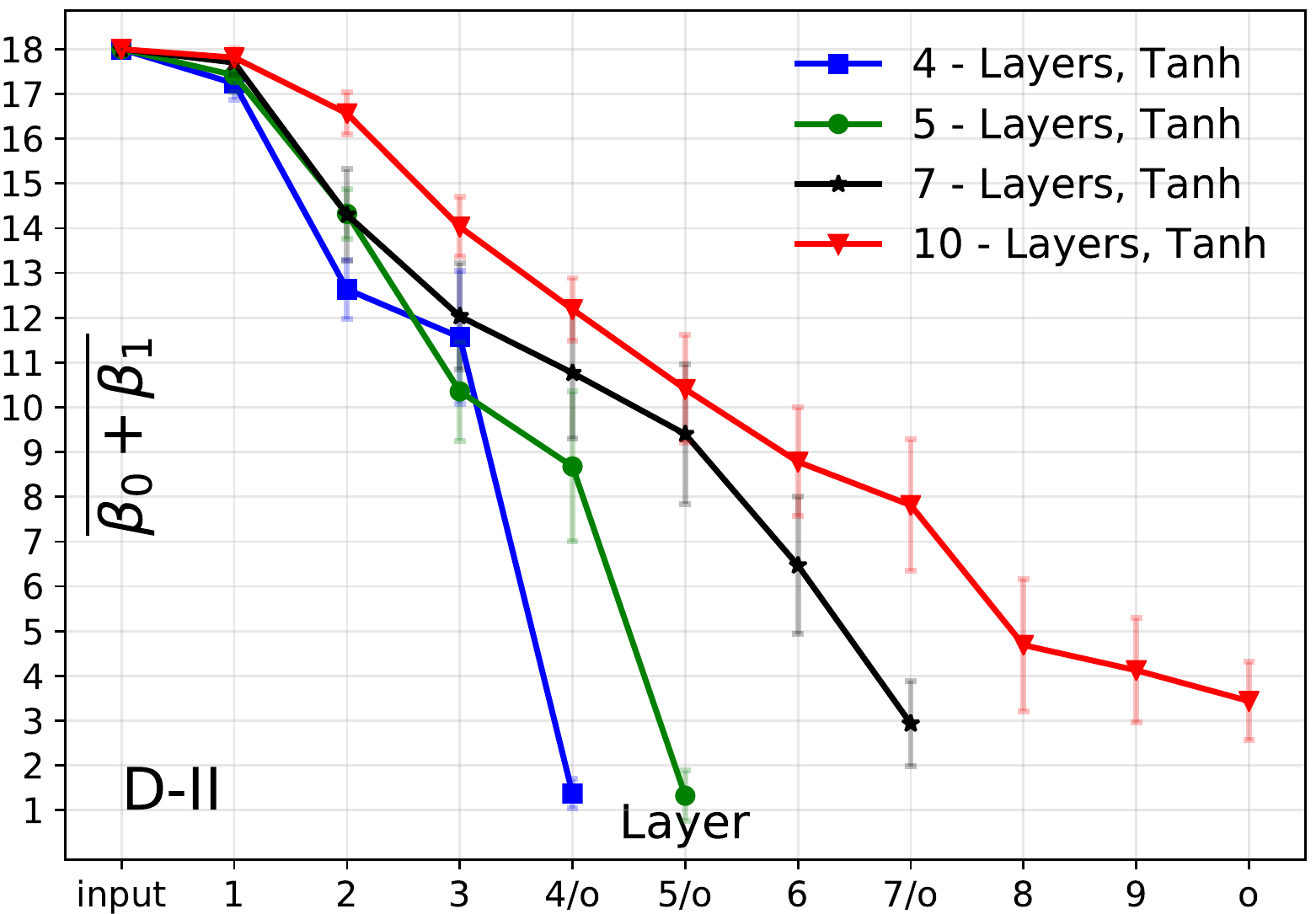}
\hspace{-1.3ex}
\includegraphics[trim={8mm 0 0 0}, clip, scale=0.32]{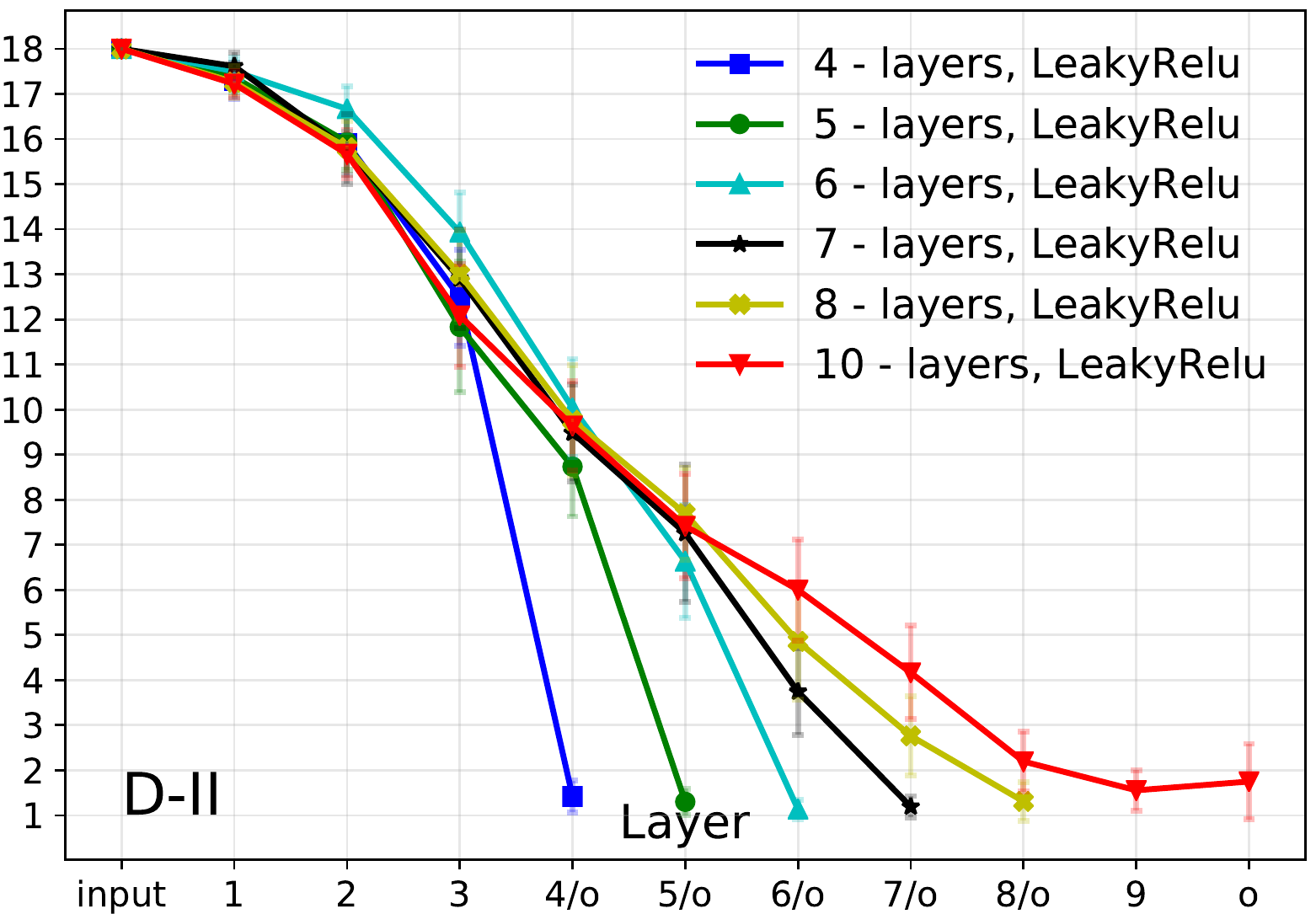}
\hspace{-1.3ex}
\includegraphics[trim={8mm 0 0 0}, clip, scale=0.32]{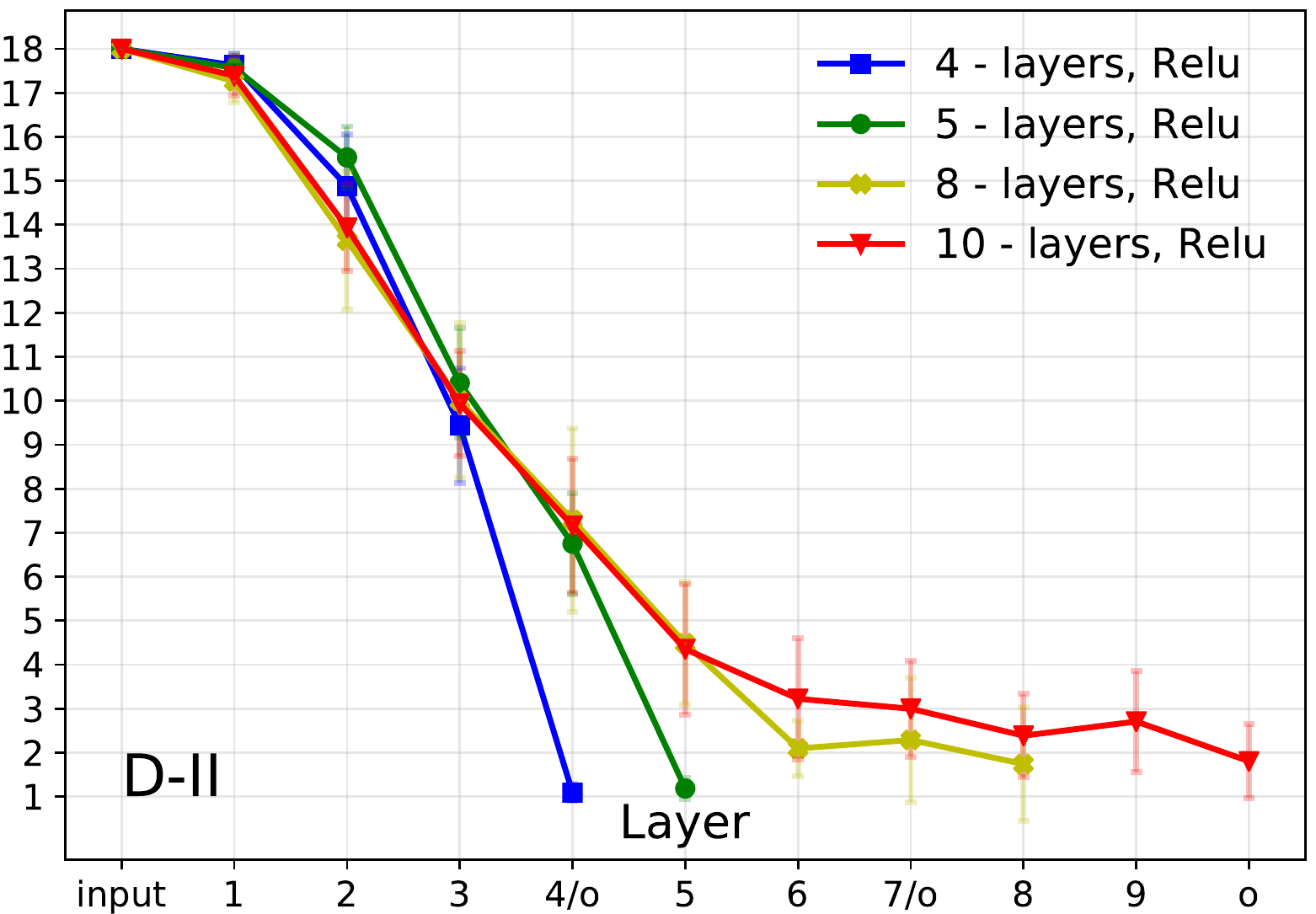}
\hspace{-1.5ex}
\includegraphics[scale=0.32]{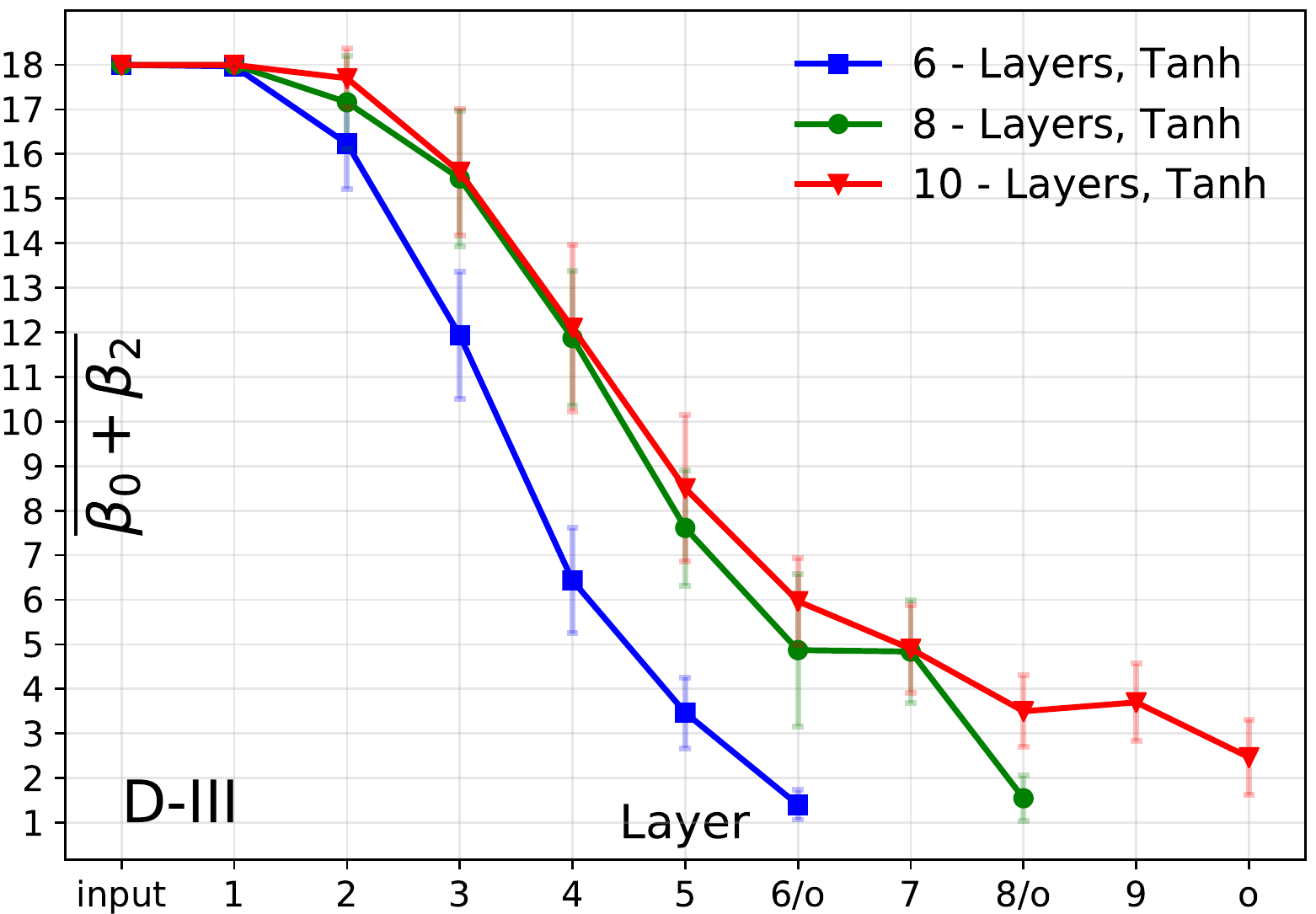}
\hspace{-1.3ex}
\vspace{0.3cm}
\includegraphics[trim={8mm 0 0 0}, clip, scale=0.32]{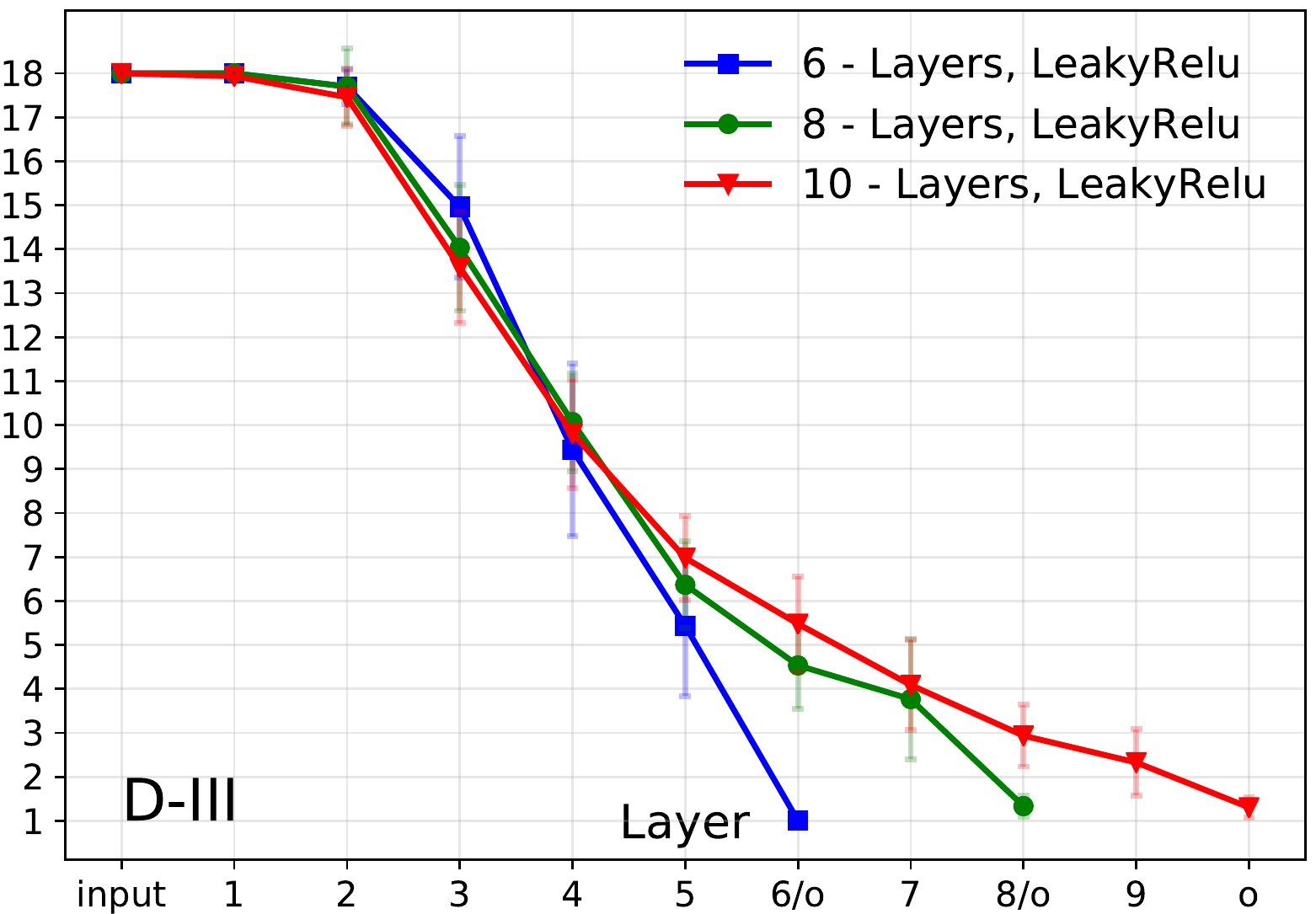}
\hspace{-1.3ex}
\includegraphics[trim={8mm 0 0 0}, clip, scale=0.32]{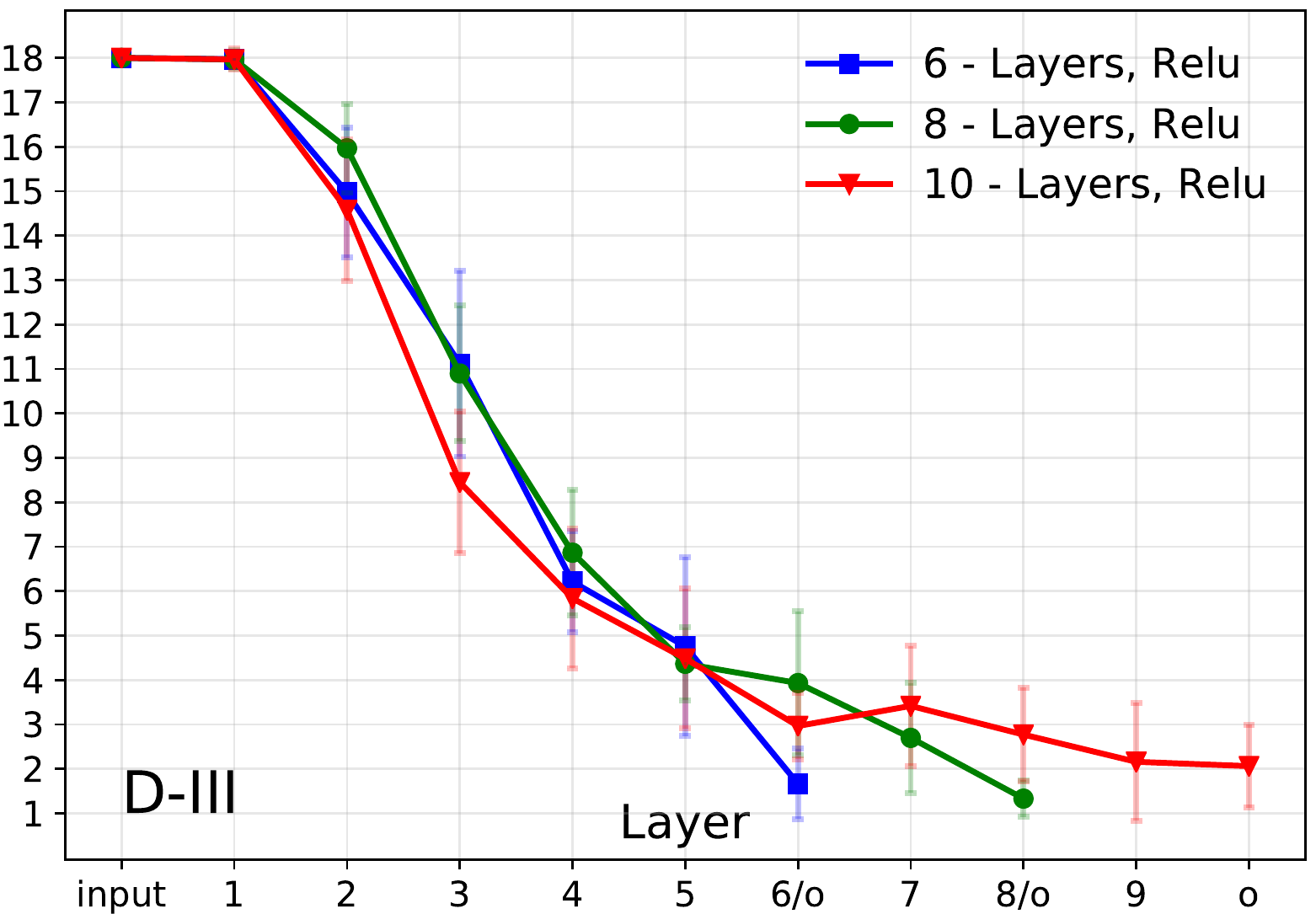}
\hspace{-1.5ex}
\caption{Mean values of topological complexity $\TC\bigl(\nu_k(M_a)\bigr)$, $k=1, \dots, l$, for fifteen-neuron-wide networks of  varying depths. Error bars indicate $\pm$ half standard deviation about the mean.}\label{fig:depth}
\end{figure}

Reducing the depth of a constant-width network beyond a certain threshold makes it increasingly difficult to train the network to high accuracy --- the percentage of successfully trained networks drops noticeably. Moreover, as the depth is reduced, the burden of changing topology does not spread evenly across all layers but becomes concentrated in the final layers. The initial layers do not appear to play a big role in changing topology, reducing depth simply makes the final layers `work harder' to produce larger reductions in Betti numbers. Figure~\ref{fig:depth} shows this effect. 

\section{Consistency with real-world data}\label{sec:real_data}

The results in Section~\ref{sec:results} are deduced from experiments on the simulated data sets D-I, D-II, D-III generated in Section~\ref{sec:data}. It is naturally to ask if these results remain valid on real data. In this section, we will see that they are, with some mild caveats. The key difference between real and simulated data is only in the amount of computational effort required to carry out our experiments --- they are much more expensive for real data sets.

We will validate our results on four real-world data sets from (i) MNIST Handwritten Digits \cite{mnist}, (ii) HTRU2 High Time-Resolution Universe Survey \cite{pulsardata2016}, (iii) UCI Banknotes Authentication \cite{banknotes2013}, (iv) UCI Sensorless Drive Diagnostic \cite{drive2013}. These data sets were chosen on the basis that they are real-valued and may be trained to high accuracy. The goal, as usual, is to observe how their topology changes as they pass through the layers of well-trained neural networks. To this end, our methodology in Section~\ref{sec:methodology} applies to these data sets with some modifications: 
\begin{itemize}
\item For real data, it is no longer possible to obtain the kind of near-perfectly trained neural networks in Section~\ref{sec:nn} that we could readily obtain with simulated data. As such, we adjust our expectations accordingly. By a well-trained neural network on a real data set, we mean one whose test accuracy ranges between $95$ to $98$ percent (recall that for simulated data, we required 99.99\% or better).

\item Unlike the simulated data sets in Section~\ref{sec:data}, we do not already know the topology of our real data sets and this has to be determined with persistent homology. More importantly, for real data, it is  no longer possible to set a single scale for observing topological changes across different layers, as  described in Section~\ref{sec:comhom} --- we have to compute persistent homology in every layer to track topological changes.
\end{itemize}
The complexity of real-world data and the need to calculate full persistent homology at every layer limits the number of experiments that we could run. As it will be prohibitively expensive to carry out extensive exploratory tests across a range of different architectures like what we did on simulated data (see Table~\ref{tab:experiments}), we will keep both  width (10 neurons) and depth (10 layers) fixed in this section. In any case, our experiments on real data sets are not intended to be exploratory but to corroborate the findings in Section~\ref{sec:results} that we deduced from simulated data. We seek confirmation on two findings in particular: the reduction in topological complexity through the layers and  the relative effectiveness of ReLU over $\tanh$ activations in achieving this. Note that while we will only present the persistence barcodes at the output of first, the middle, and the final layer, we computed persistence homology in every layer; interested readers may easily get those for other layers from our publicly available codes.

\subsection{MNIST Handwritten Digits \cite{mnist}.} 
\begin{figure}[ht]
\centering
\includegraphics[width=0.48\linewidth]{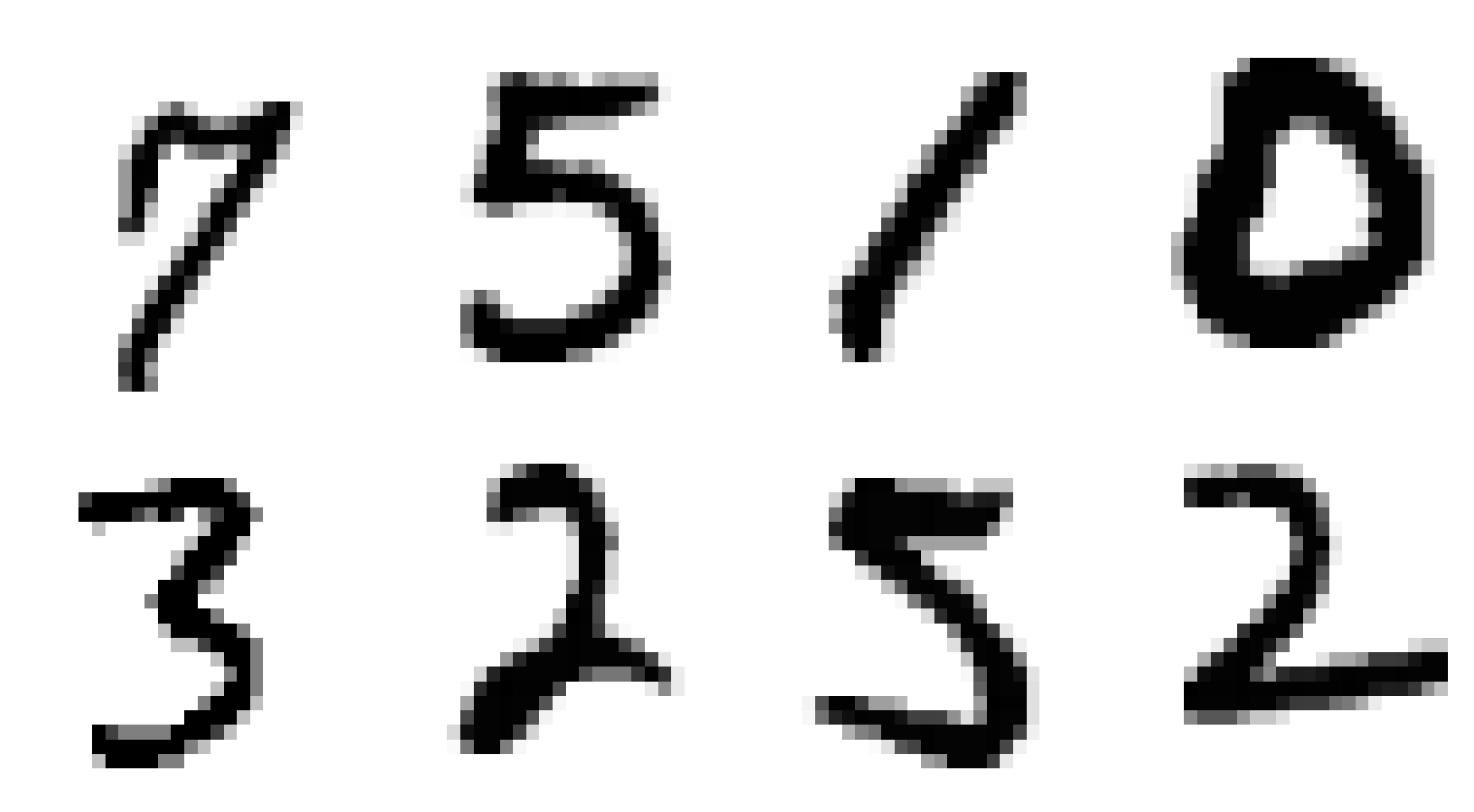}
\hspace{0.3cm}
\includegraphics[width=0.48\linewidth]{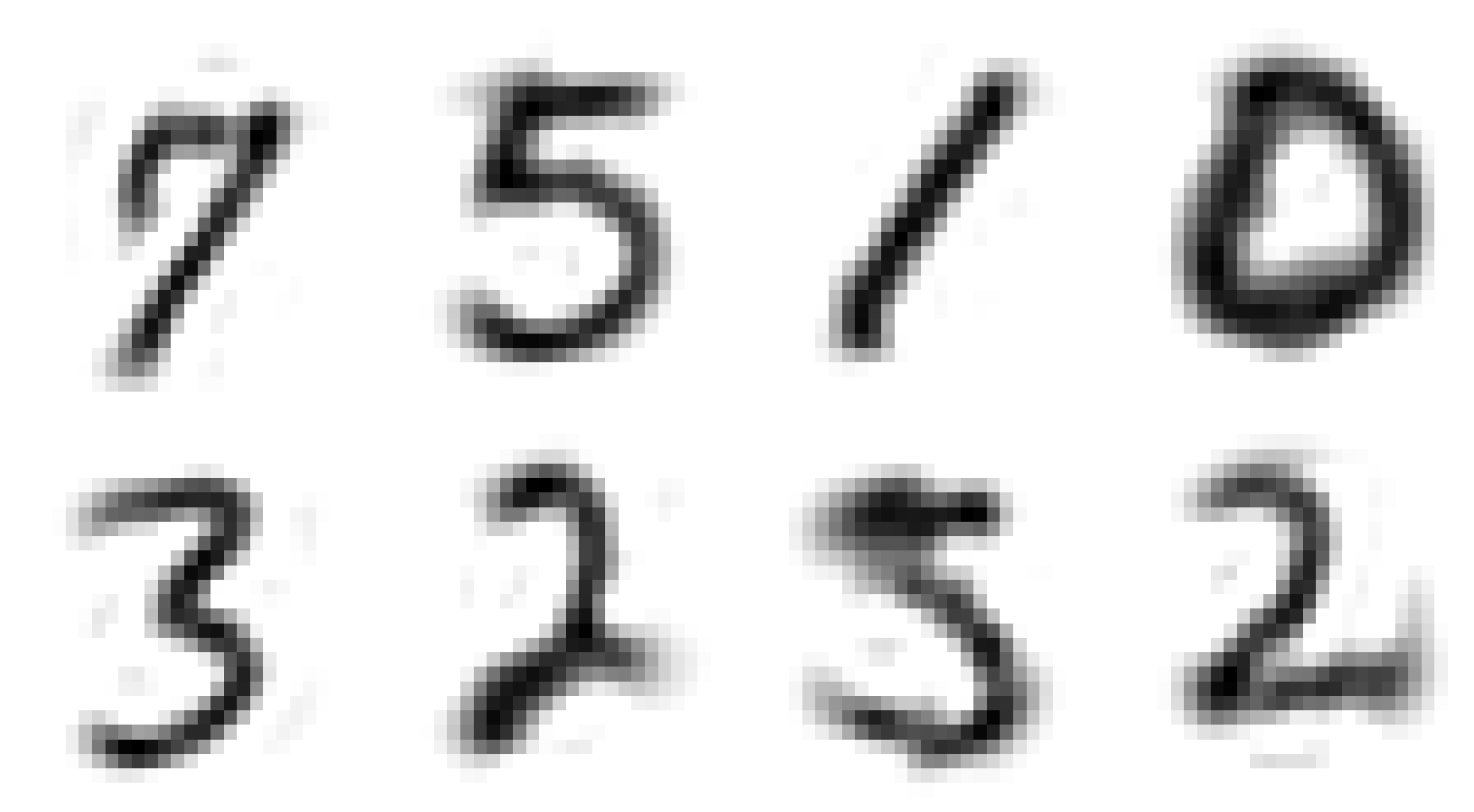}
\caption{\emph{Left}: Original MNIST handwritten digits. \emph{Right}: MNIST handwritten digits projected onto first fifty principal components.}
\label{fig:mnist}
\end{figure}

Each of the 70,000 images in the MNIST handwritten digits data set  is an image of size $28 \times 28$-pixels and collectively forms a point cloud on some manifold $M \subseteq \mathbb{R}^{784}$. Computing persistent homology for a $784$-dimensional point cloud is way beyond what our computing resource could handle and we first reduce dimension by projecting  onto its leading $50$ principal components. Nevertheless, the dimension-reduced images remain to be of reasonably high-quality; we show a comparison of a few original digits alongside their principal component projections onto  $\mathbb{R}^{50}$ in Figure~\ref{fig:mnist}. We will take the dimension-reduced point cloud $X$ as the starting point for our experiments and will loosely refer to $X \subseteq M \subseteq \mathbb{R}^{50}$ as the MNIST data set.

Since we would like to have a binary classification problem for consistency with all our other experiments, instead of regarding the MNIST data as a classification problem with ten classes, we reduce it to a problem of classifying a chosen (any) digit $a$ versus all non-$a$ digits. So our manifold is $M =M_a \cup M_b$, where $M_a$ is  the `manifold all handwritten $a$' and $M_b$ is the union of the `manifolds of all other non-$a$ handwritten digits.' Following our methodology in Section~\ref{sec:methodology}, we train a neural network with ten layers and ten neuron in each layer on a labelled point cloud $X_a = X \cap M_a$ and $X_b = X \cap M_b$ to classify points in $M_a$ and $M_b$; then with persistent homology we analyze how $M_a$ is transformed after each layer in the network. While we fix the depth and widths, we vary the activation among $\tanh$, leaky ReLU, and ReLU. We used 60,000 samples to train and the remaining 10,000 samples to  test our neural networks; we use one third of the test set for our persistent homology calculations.

\begin{table}[h]
\centering 
\renewcommand{\arraystretch}{1.2}
\footnotesize
\begin{tabular}[b]{c|c|cccccp{0.3cm}ccccc}
activation&  scale $\varepsilon$  &    $k = 0$  & $1$  & $ 2$ & $3$ & $4$ & $5$ & $6$ & $7$ & $8$ & $9$ &  $10$  \\
\hline
\multirow{3}{*}{{\rotatebox[origin=c]{90}{$\tanh$}}
}  & $1.5$ & $525$ & $408$ & $356$ & $266$ & $233$ & $145$ & $156$ & $88$  & $30$ & $20$ & $9$ \\
&  $2.5$ & $6$ & $5$ & $2$ & $1$ & $3$ & $14$ & $12$ & $8$ & $1$ & $4$ & $4$  \\
&  $3.5$ & $1$ & $1$ & $1$ & $1$ & $1$ & $1$ & $1$ & $1$ & $1$ & $1$ & $3$ \\
\hline
\multirow{3}{*}{{\rotatebox[origin=c]{90}{leaky}}
{\rotatebox[origin=c]{90}{ReLU}}
}  &  $1.5$ & $525$ & $340$ & $182$ & $108$ & $38$ & $16$ & $10$ & $8$ & $1$ & $1$ & $1$ \\
&  $2.5$ & $6$ & $6$ & $6$ & $5$ & $1$ & $1$ & $2$ & $1$ & $1$ & $1$ & $1$  \\
&  $3.5$ & $1$ & $1$ & $1$ & $1$ & $1$ & $1$ & $1$ & $1$ & $1$ & $1$ & $1$ \\
\hline
\multirow{3}{*}{{\rotatebox[origin=c]{90}{ReLU}}} &  $1.5$ & $525$ & $199$ & $106$ & $27$ & $13$ & $6$ & $1$ & $1$  & $1$ & $1$ & $1$ \\
&  $2.5$ & $6$ & $2$ & $6$ & $6$ & $2$ & $1$ & $1$ & $1$ & $1$ & $1$ & $1$  \\
&  $3.5$ & $1$ & $1$ & $1$ & $1$ & $1$ & $1$ & $1$ & $1$ & $1$ & $1$ & $1$ \\
\end{tabular}
\bigskip
\caption{Topological complexity $\omega(M_a) = \beta_0\bigl(\nu_k(M_a)\bigr)+\beta_1\bigl(\nu_k(M_a)\bigr)+\beta_2\bigl(\nu_k(M_a)\bigr)$ at layers $k =0,1,\dots,10$ with $M_a$ the `manifold of handwritten $a$'. Network has $50$-dimensional input,
$2$-dimensional output, and is $10$-dimensional in intermediate layers. For each of three activation types, we show homology at three scales $\varepsilon=1.5$, $2.5$, $3.5$.}
\label{tbl:mnist_all}
\end{table}
The results of our experiments for the digit $a = 0$ are shown in Table~\ref{tbl:mnist_all}. What we see in the table corroborates our earlier findings on simulated data sets --- an unmistakable reduction in topological complexity through the layers, with ReLU activation reducing topological complexity most rapidly when compared to the other two activations. With $\tanh$ activation, reduction in topological complexity is not only much slower but the network fails to reduce $M_a$ to a topological disk, despite having ten layers.
The persistence barcodes diagram for the MNIST data set is much larger than those for the next three data sets but it is not much more informative than our summary statistics in Table~\ref{tbl:mnist_all}. As such we do not show it here although it can just as readily be generated from our program.

\subsection{HTRU2 High Time Resolution Universe Survey \cite{pulsardata2016}.} This data set consists of statistics of radio source signals from 17,898 stars, measured during the High Time Resolution Universe Survey (HTRU2) experiment to identify pulsars. For our purpose, it suffices to know that pulsars are stars that produce radio emission measurable on earth. In the HTRU2 data set, each recorded radio emission is described by eight continuous variable: four are statistics of the radio signal called `integrated profile' and the other four are statistics of the `DM-SNR curve' that tracks frequency components of the signal versus its arrival time. The radio sources are labeled by $a$ or $b$ according to whether the source is a pulsar or not. We show a small portion of this data set in Table \ref{tbl:pulsars}.  

\begin{table}[h]
\centering 
\renewcommand{\arraystretch}{1.2}
\footnotesize
\begin{tabular}{r|d{3.4}|d{3.4}|d{3.4}|d{3.4}|d{3.4}}
Star \# & \multicolumn{1}{c|}{1}    & \multicolumn{1}{c|}{2}  & \multicolumn{1}{c|}{3}  & \multicolumn{1}{c|}{4}  & \multicolumn{1}{c}{5}\\ 
\hline
Mean  (integral profile) & 140.5625 & 102.5078  &  103.0156
& 136.7500
& 99.3672\\
Standard Deviation (integral profile) & 55.6838 & 45.5499& 39.3416
& 57.1784
& 41.5722\\
Excess Kurtosis (integral profile)
& -0.2346 & 0.2829& 0.3233
& -0.0684
&0.4653\\ 
Skewness (integral profile) & -0.6996 &0.4199& 1.0511
& -0.6362
& 4.1541\\ 
Mean (DM-SNR) & 3.1998& 1.3587   & 3.1212
& 3.6423  &  1.6773\\ 
Standard Deviation (DM-SNR)  & 19.1104 & 13.0790 & 21.7447
& 20.9593
& 61.7190\\ 
Excess Kurtosis (DM-SNR) & 7.9755 & 13.3121 & 7.7358
& 6.8965
& 2.2088\\ 
Skewness (DM-SNR) & 74.2422 & 212.5970 & 63.1720
& 53.5937
&  127.3930\\
Pulsar `$a$' or not `$b$' & \multicolumn{1}{c|}{$b$}  & \multicolumn{1}{c|}{$a$}  & \multicolumn{1}{c|}{$b$}  & \multicolumn{1}{c|}{$b$} & \multicolumn{1}{c}{$b$}
\end{tabular}
\bigskip
\caption{Five entries from HTRU2. The first eight rows are statistics of the radio signal from that star. The last row indicates whether the respective star is a pulsar `$a$'  or not `$b$'.}
\label{tbl:pulsars}
\end{table}

\begin{figure}[H]
\centering
\includegraphics[width=0.33\linewidth]{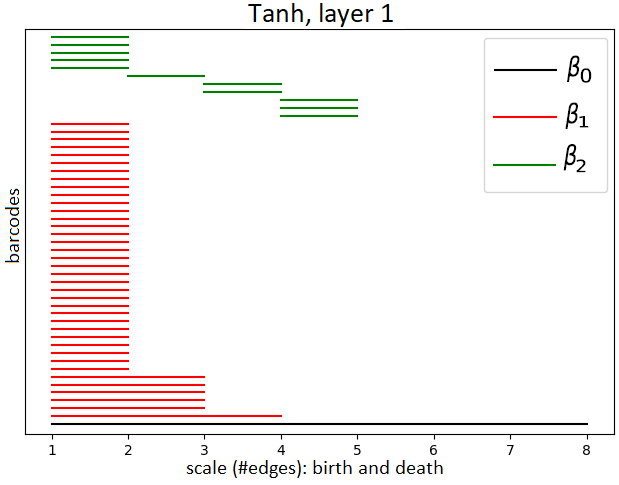}
\hspace{-1.0ex}
\includegraphics[width=0.33\linewidth]{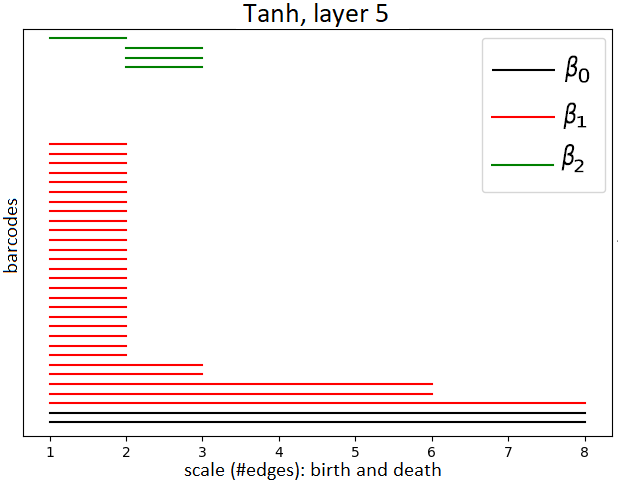}
\hspace{-1.0ex}
\includegraphics[width=0.33\linewidth]{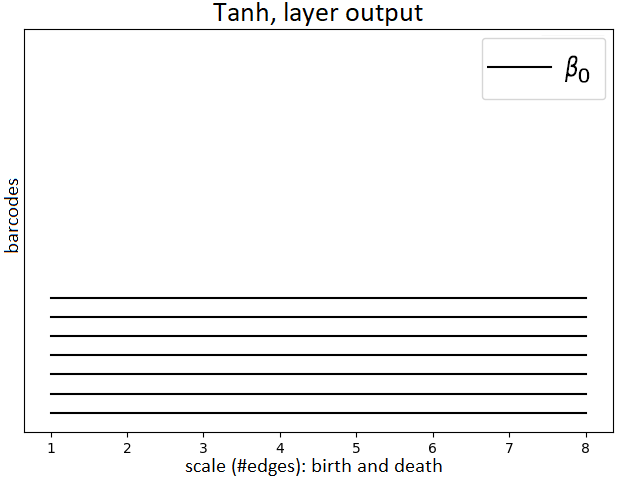}
\includegraphics[trim=150 100 150 170, clip, width=0.31\linewidth]{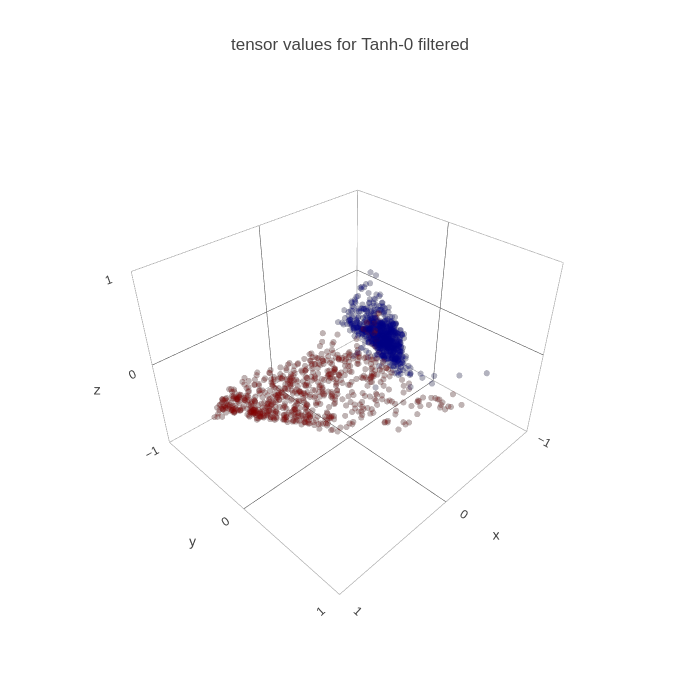}
%left bottom right top
\includegraphics[trim=150 100 155 160, clip, width=0.31\linewidth]{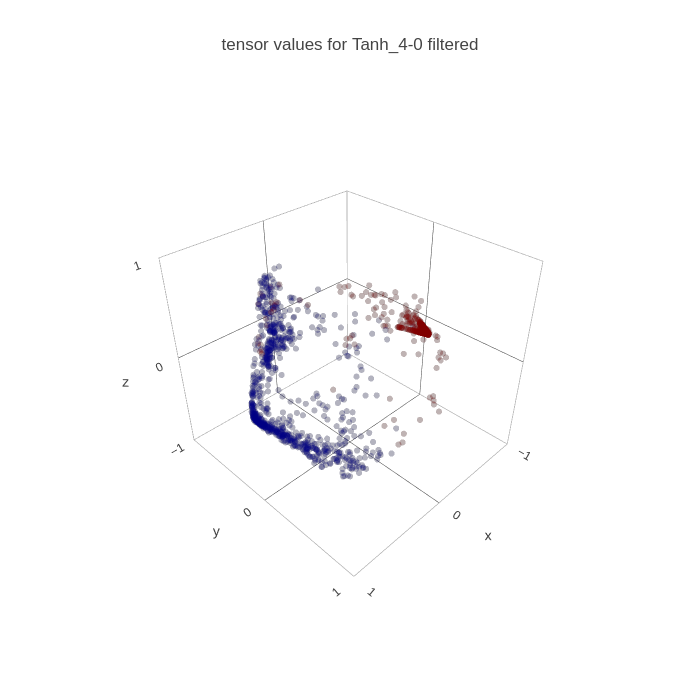}
\includegraphics[trim=150 100 150 180, clip, width=0.31\linewidth]{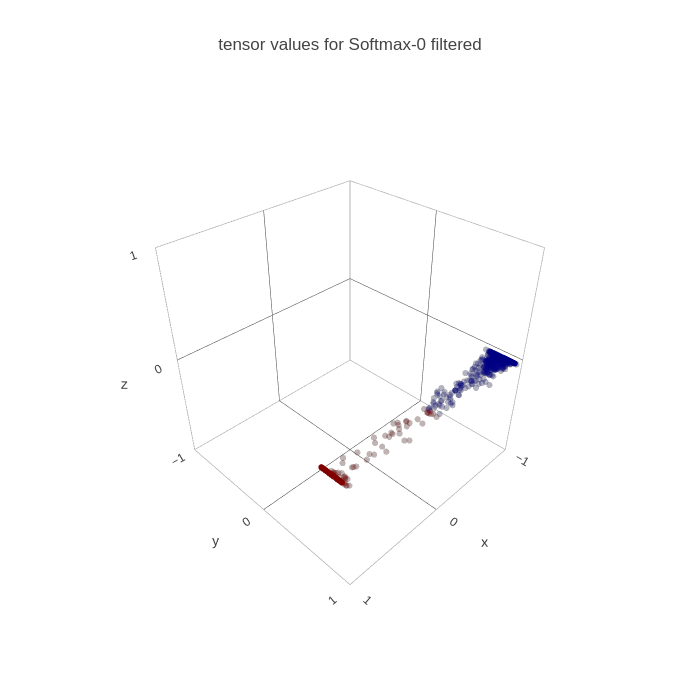}
\includegraphics[width=0.33\linewidth]{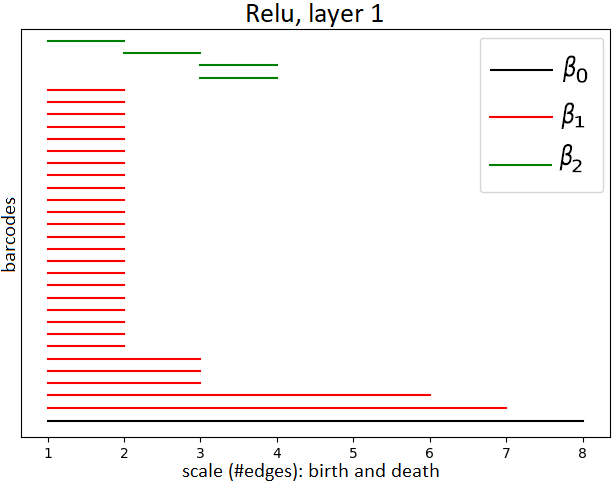}
\hspace{-1.0ex}
\includegraphics[width=0.33\linewidth]{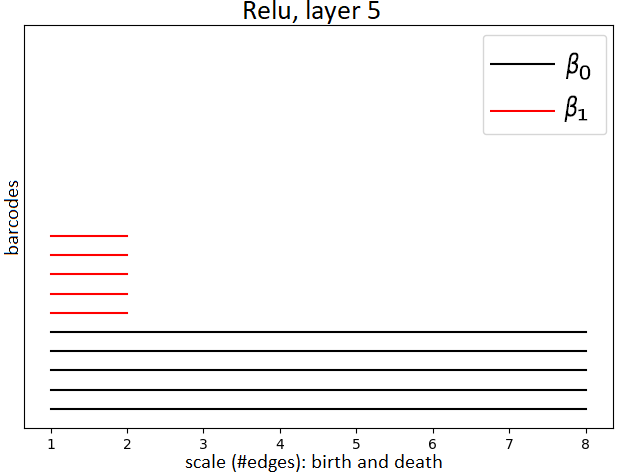}
\hspace{-1.0ex}
\includegraphics[width=0.33\linewidth]{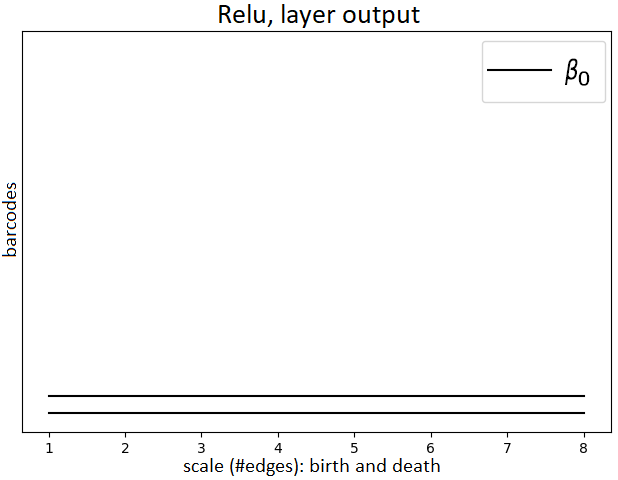}
\includegraphics[trim=150 100 160 170, clip, width=0.31\linewidth]{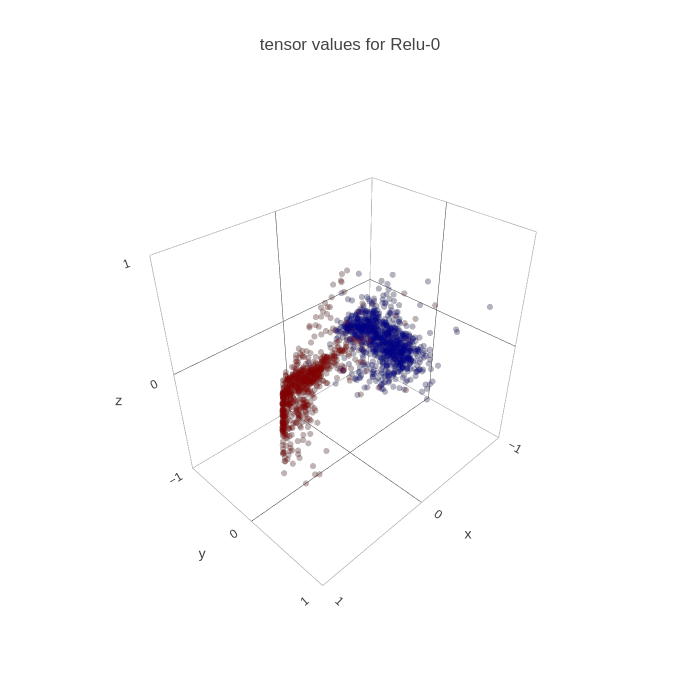}
%left bottom right top
\includegraphics[trim=155 100 163 160, clip, width=0.31\linewidth]{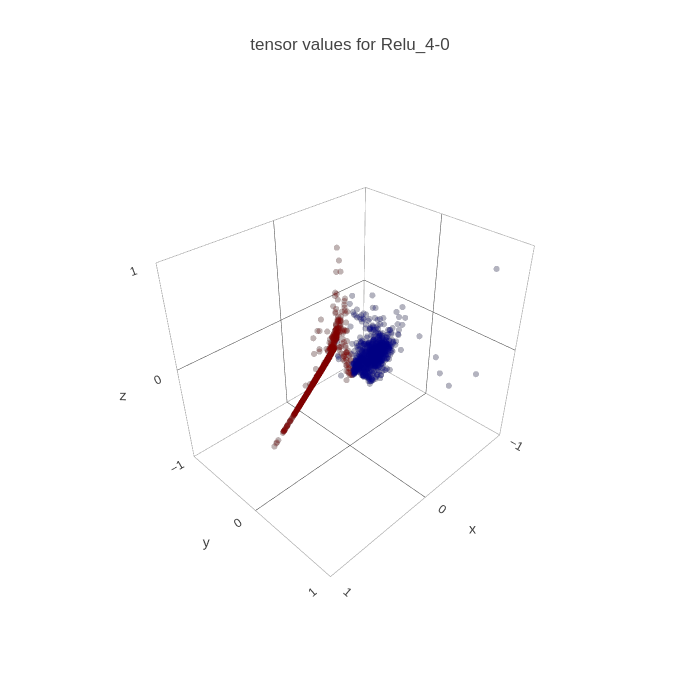}
\includegraphics[trim=152 102 170 186, clip, width=0.31\linewidth]{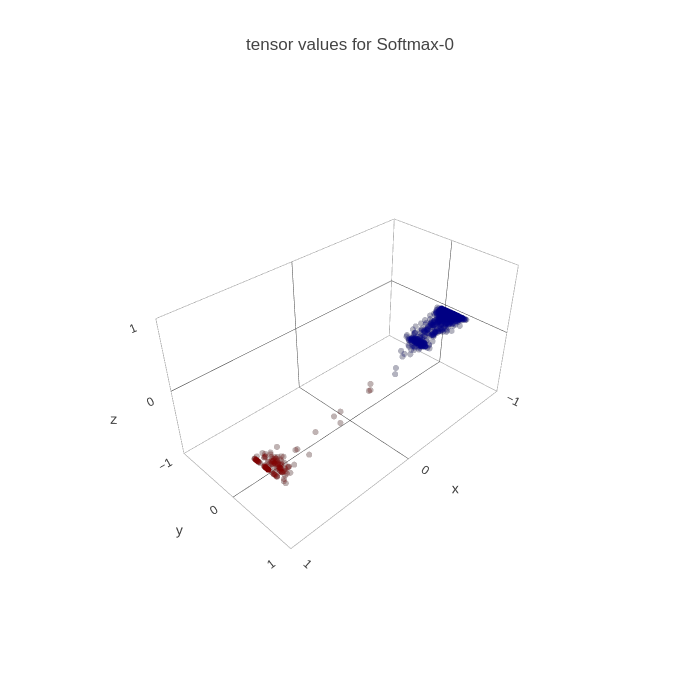}
\medskip
\caption{Persistence barcode diagrams for neural networks trained on HTRU2 show topology changes in the `pulsar manifold' $M_a$ as it passes through layers activated with $\tanh$ (\emph{top}) and ReLU (\emph{bottom}).  Scatter plots show principal component projections of $M_a$ (red) and the `non-pulsar manifold' $M_b$ (blue) at the corresponding layers.}
\label{fig:ph_pulsar_dataset}
\end{figure}

We take a 3,278 subsample of the HTRU2 data set so that we have an equal number of pulsar and non-pulsars. This is a point cloud $X \subseteq M \subseteq \mathbb{R}^8$ with $M = M_a \cup M_b$ a union of the `pulsar manifold' $M_a$ and `non-pulsar manifold' $M_b$; the point clouds $X_a = X \cap M_a$ and $X_b = X \cap M_b$ each has 1,639 points. We use 80\% of this balanced data $X$ for training the neural networks and the remaining 20\% for testing. For persistent homology computations, we use the test set but we first passed it through a local outlier removal algorithm in \cite{LOF2000} for denoising. Again, our neural networks have ten layers with ten neurons in each layer and are activated with either ReLU or $\tanh$.

In Figure~\ref{fig:ph_pulsar_dataset}, we show the persistence barcode diagrams for $\nu_k(X_a)$ in the first, middle, and last layer, i.e., $k =1,5,10$. The scatter plots below the barcode diagrams show,  for $k =1,5,10$, the projections of $\nu_k(X_a)$ (red) and $\nu_k(X_b)$ (blue) onto the three leading principal components. These persistent barcodes tell the same story for the HTRU2 data as Table~\ref{tbl:mnist_all} does for the MNIST data and Figures~\ref{fig:2d_dataset}, \ref{fig:dataset_ii}, \ref{fig:dataset_iii} do for the simulated data D-I, D-II, D-III: Topology is simplified as the data passes through the layers; and ReLU does a better job  than $\tanh$ activation at simplifying topological complexity.

\subsection{UCI Banknotes Authentication \cite{banknotes2013}.} 
\begin{figure}[h]
\centering
\includegraphics[width=0.8\linewidth]{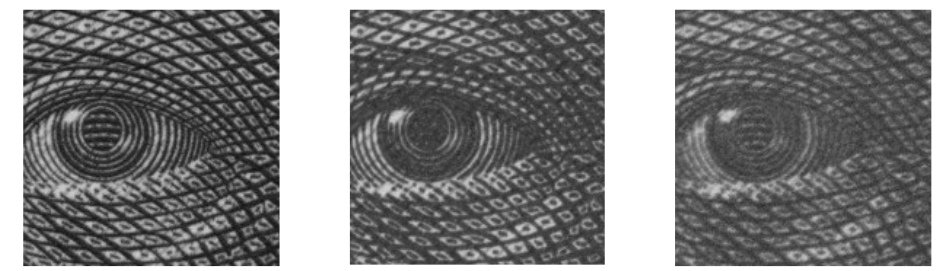}
\caption{Texture sample for genuine banknote (\emph{left}),  high-quality forgery (\emph{middle}) and  low-quality forgery (\emph{right}). The figures are taken from \cite{banknotes2013}.}
\label{fig:banknotes}
\end{figure}

This data set is derived from $400 \times 400$-pixels gray scale images of 1,372 genuine and forged banknotes; small patches ranging in sizes from $96 \times 96$ to $128 \times 128$-pixels are extracted from the images and wavelet-transformed. Figure~\ref{fig:banknotes} shows three of these small extracted patches. 

The UCI Banknotes data set does not contain any images but is simply a list of four statistics computed from the wavelet coefficients of the image patches, together with a label indicating whether the patch is from a genuine `$a$' or forged `$b$' banknote. We show five of these entries in Table~\ref{tbl:bbanknotes}; the full data set comprises 1,372 entries like these.
\begin{table}[h]
\centering 
\renewcommand{\arraystretch}{1.2}\newcolumntype{C}[1]{>{\centering\arraybackslash}p{1cm}|}
\footnotesize
\begin{tabular}{r|d{2.4}|d{2.4}|d{2.4}|d{2.4}|d{2.4}|d{2.4}}
Banknote \# & \multicolumn{1}{c|}{1}  & \multicolumn{1}{c|}{2} & \multicolumn{1}{c|}{3} & \multicolumn{1}{c|}{4} & \multicolumn{1}{c|}{5} & \multicolumn{1}{c}{6} \\ 
\hline
Variance (wavelet coef.) &  -1.3971 & 4.5459 &  3.8660
& 3.4566
& 0.3292 & 0.3901\\
Skewness (wavelet coef.)  & 3.3191 & 8.1674&-2.6383
& 9.5228
& -4.4552  & -0.1428 \\
Kurtosis (wavelet coef.)
& -1.3927 & -2.4586 & 1.9242
& -4.0112
& 4.5718 & -0.0319\\ 
Entropy (wavelet coef.) & -1.9948 &-1.4621& 0.1065
& -3.5944
& -0.9888 &  0.3508\\ 
Genuine `$a$' or forged `$b$' & \multicolumn{1}{c|}{$a$} & \multicolumn{1}{c|}{$b$}  & \multicolumn{1}{c|}{$b$}  & \multicolumn{1}{c|}{$b$}  & \multicolumn{1}{c|}{$b$}  & \multicolumn{1}{c}{$a$}
\end{tabular}
\bigskip
\caption{Five entries from the UCI Banknotes data set. The first four rows are statistics of the wavelet coefficients of banknotes image patches. The last row indicates whether the respective banknote is genuine `$a$' or forged `$b$'.}
\label{tbl:bbanknotes}
\end{table}
\begin{figure}[H]
\centering
\includegraphics[width=0.33\linewidth]{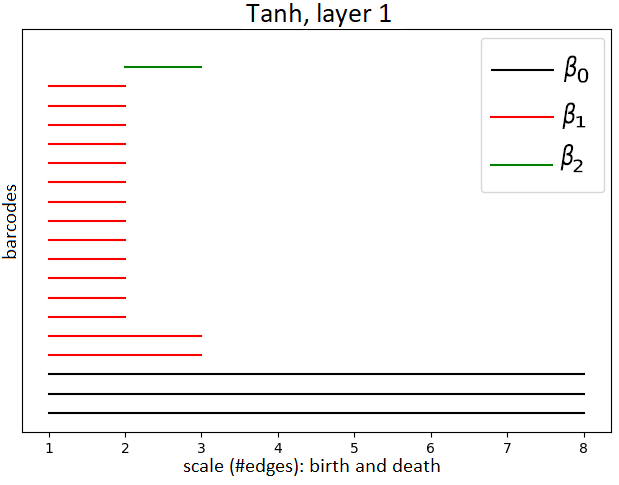}
\hspace{-1.0ex}
\includegraphics[width=0.33\linewidth]{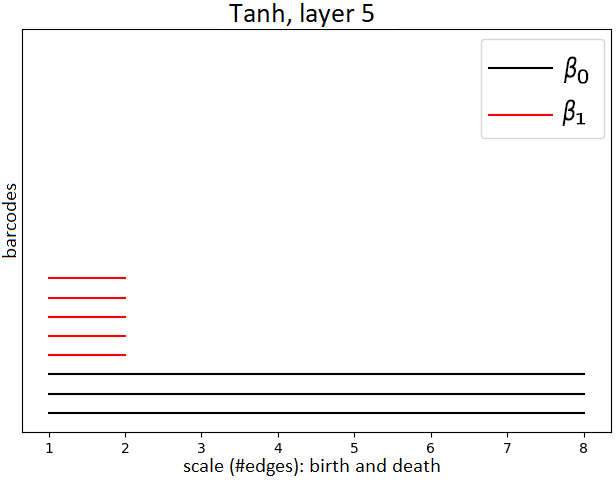}
\hspace{-1.0ex}
\includegraphics[width=0.33\linewidth]{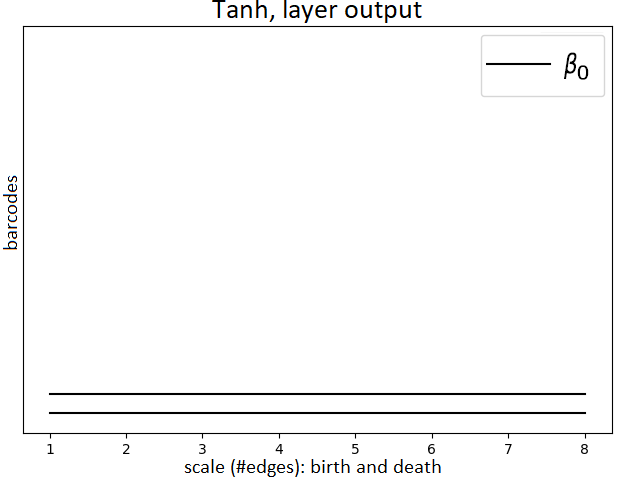}
\includegraphics[trim=160 100 160 170, clip, width=0.31\linewidth]{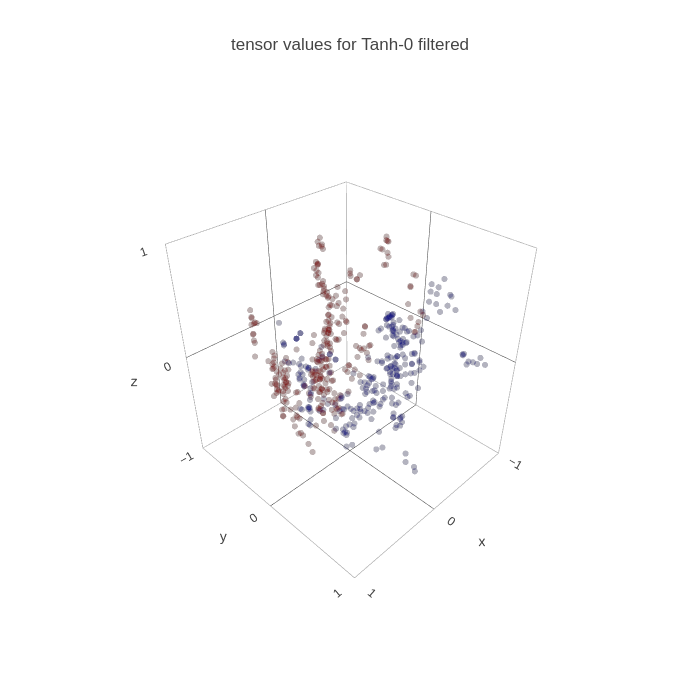}
%left bottom right top
\includegraphics[trim=158 100 150 195, clip, width=0.31\linewidth]{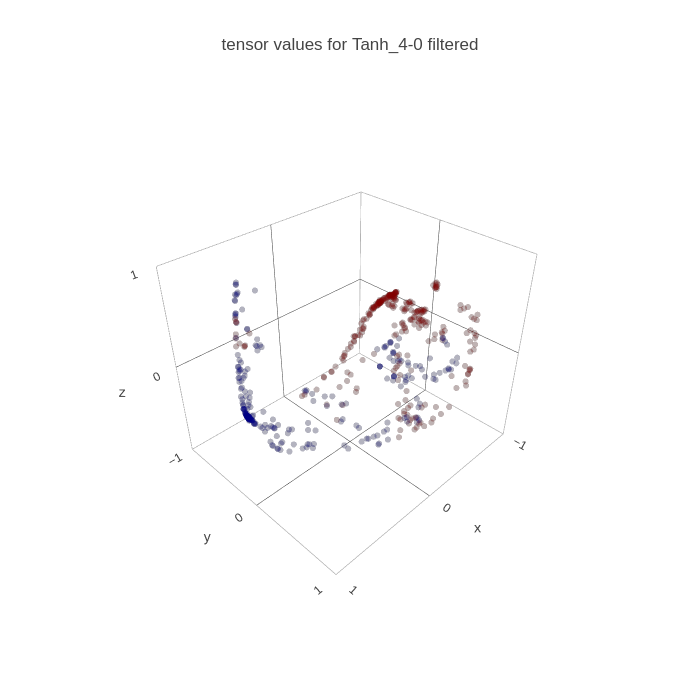}
\includegraphics[trim=160 150 190 220, clip, width=0.31\linewidth]{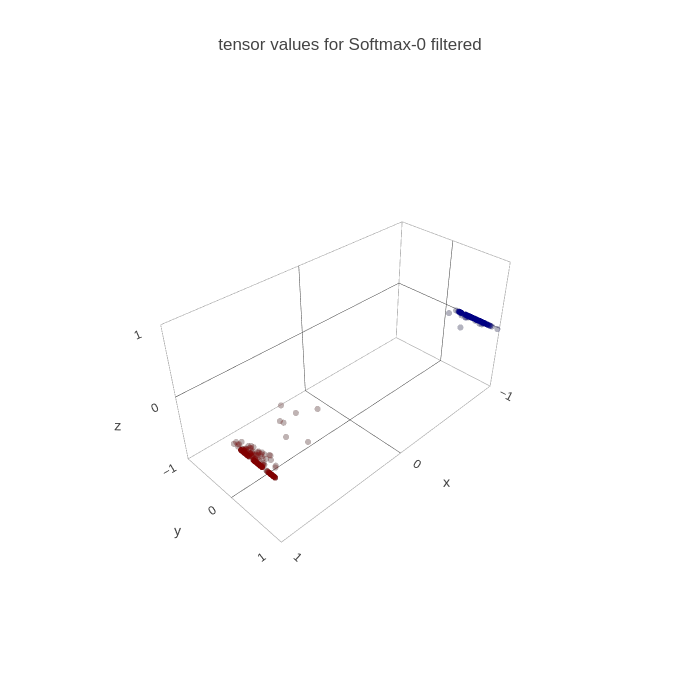}
\includegraphics[width=0.33\linewidth]{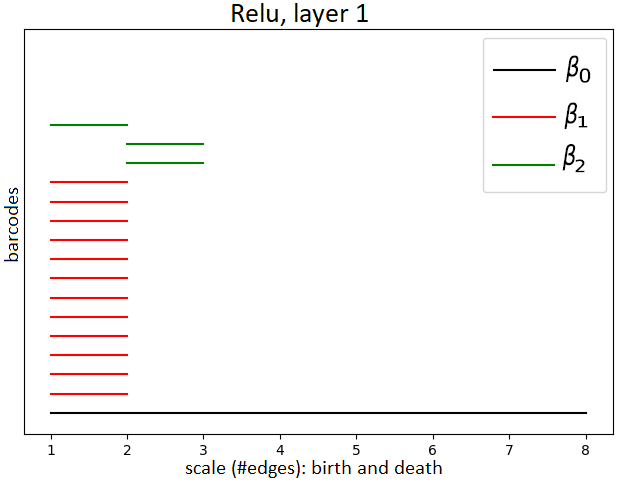}
\hspace{-1.0ex}
\includegraphics[width=0.33\linewidth]{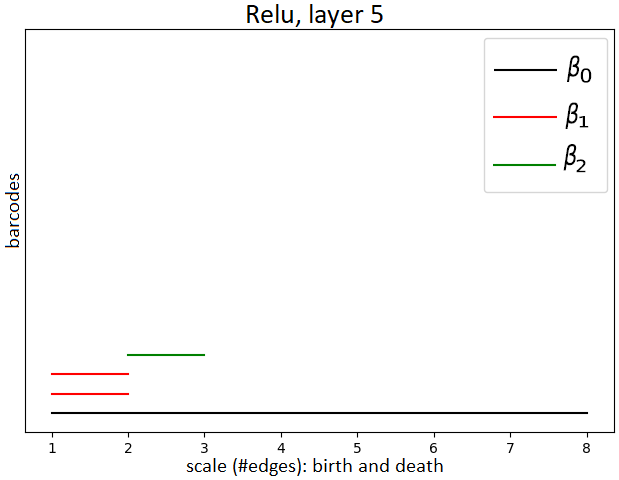}
\hspace{-1.0ex}
\includegraphics[width=0.33\linewidth]{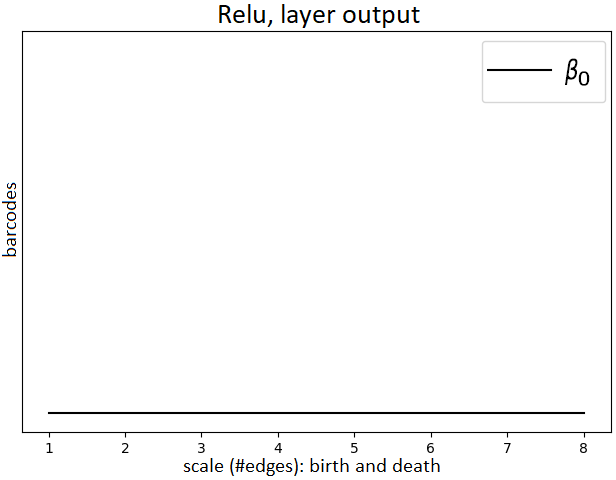}
\includegraphics[trim=150 100 160 170, clip, width=0.31\linewidth]{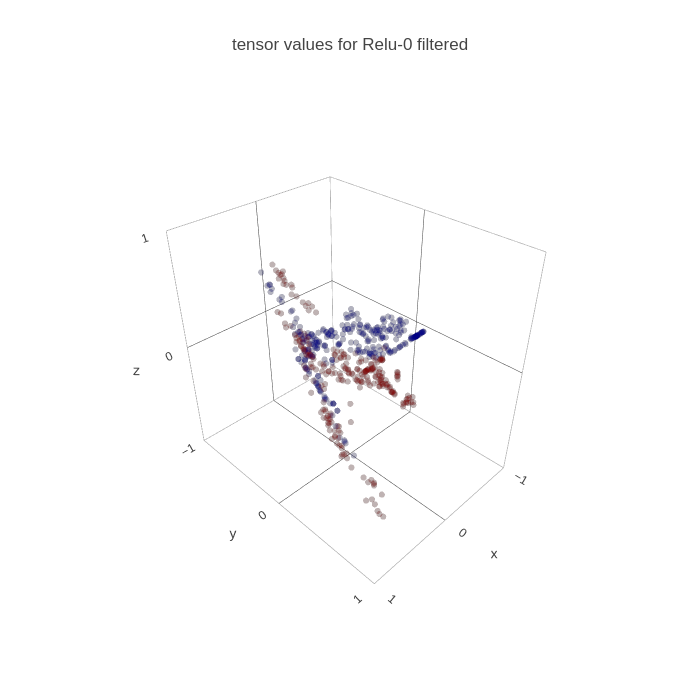}
%left bottom right top
\includegraphics[trim=150 100 160 160, clip, width=0.31\linewidth]{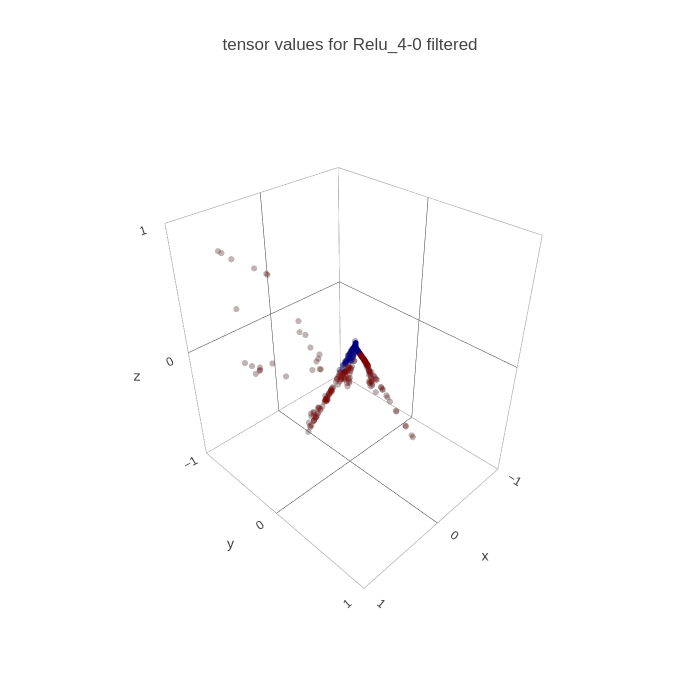}
\includegraphics[trim=180 150 230 220, clip, width=0.31\linewidth]{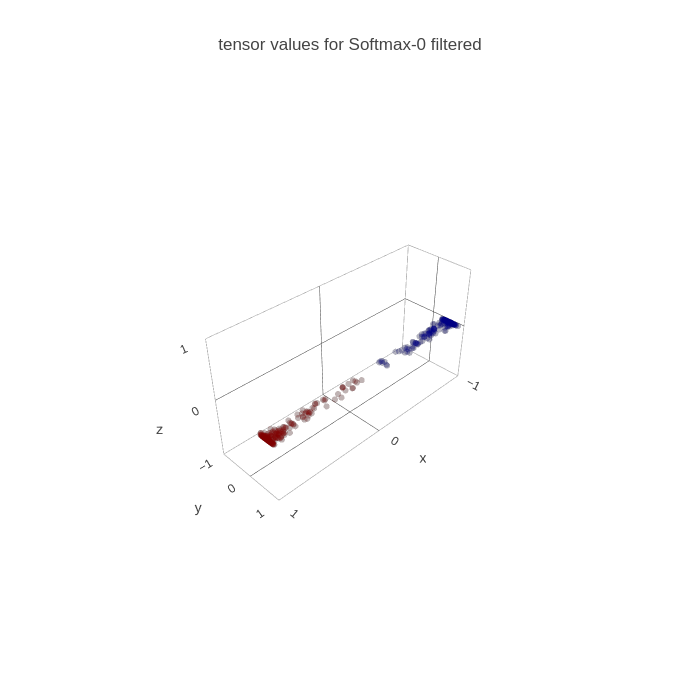}
\caption{Persistence barcode diagrams for neural networks trained on UCI Banknotes data show topology changes in the `manifold of genuine banknotes' $M_a$ as it passes through layers activated with $\tanh$ (\emph{top}) and ReLU (\emph{bottom}).  Scatter plots show principal component projections of $M_a$ (red) and the `manifold of forged banknotes' $M_b$ (blue) at the corresponding layers.}
\label{fig:ph_banknotes_autentication_dataset}
\end{figure}
As with the HTRU2 data set, we subsample 1,200 entries from the UCI Banknotes data set so that we have an equal number of genuine and forged samples; we use 80\% of this data set for training and 20\% for testing; and for persistent homology computations, we preprocess the data with the outliers removal algorithm in \cite{LOF2000}.  
For our purpose, the UCI Banknotes data set is a point cloud $X \subseteq M \subseteq \mathbb{R}^4$ with $M = M_a \cup M_b$ a union of the `manifold of genuine banknotes' $M_a$ and `manifold of forged banknotes' $M_b$; the point clouds $X_a = X \cap M_a$ and $X_b = X \cap M_b$ each has 600 points.

When $X_a$ and $X_b$ are passed through well-trained neural networks (ten layers, ten neurons in each layer, ReLU or $\tanh$-activated), we obtained results consistent with all earlier experiments. The persistence barcodes diagrams in Figure~\ref{fig:ph_banknotes_autentication_dataset} show that Betti numbers $\beta_1$ and $\beta_2$ are reduced to zero for both activations, $\beta_0$ successfully reduces to one when ReLU-activated but is stuck at two when $\tanh$-activated. Also, reduction of Betti numbers happens more rapidly with ReLU-activation. These observations are also reflected in the respective principal components scatter plots below each persistence barcodes diagram.

\subsection{UCI Sensorless Drive Diagnostic \cite{drive2013}.} This data set concerns a printed circuit board that operates a specific type of drive motor. The goal is to classify twelve types of common defects in the drive motor based on $49$ measurements of electric currents at various locations on the printed circuit board. Table~\ref{tbl:drive} shows a sample of five such entries in this data set, which has a total of 58,509 such entries.

\begin{table}[h]
\centering 
\renewcommand{\arraystretch}{1.2}
\footnotesize
\begin{tabular}{r|r|r|r|r|r}
sample \# & \multicolumn{1}{c|}{1}   & \multicolumn{1}{c|}{2} & \multicolumn{1}{c|}{3}  & \multicolumn{1}{c|}{4} & \multicolumn{1}{c}{5}  \\ 
\hline
Elect.\ curr.\ 1 & $-3.015\times 10^{-7}$ & $-2.952\times  10^{-6}$  &  $-2.952\times 10^{-6}$ 
& $-4.961\times 10^{-6}$ 
& $-6.501\times 10^{-6}$ 
\\
Elect.\ curr.\ 2 & $8.260\times 10^{-6}$ & $-5.248\times  10^{-6}$ & $-3.184\times 10^{-6}$ 
& $-2.088\times 10^{-6}$ 
& $-6.208\times 10^{-6}$
\\
Elect.\ curr.\ 3
&  $-1.152\times 10^{-5}$ & $3.342\times 10^{-6}$ & $-1.592\times 10^{-5}$
& $-1.366\times 10^{-5}$
&  $4.644\times 10^{-6}$ \\ 
Elect.\ curr.\ 4 &  $-2.310\times 10^{-6}$  & $-6.056\times  10^{-6}$& $-1.208\times 10^{-6}$
& $4.661\times 10^{-7}$
& $-2.749\times 10^{-6}$ 
\\ 
Elect.\ curr.\ 5
& $-1.439\times 10^{-6}$ &  $2.779\times 10^{-6}$& $-1.575\times 10^{-6}$ 
& $2.369\times 10^{-6}$
& $-4.167\times 10^{-6}$ \\ 
Elect.\ curr.\ 6 &  $-2.123\times 10^{-5}$  & $-3.752\times  10^{-6}$& $1.7394\times  10^{-5}$ 
& $3.785\times 10^{-6}$
& $-3.347\times 10^{-5}$ 
\\
$\vdots$ $\quad\quad\vdots \quad~~$  &  \multicolumn{1}{c|}{ $\vdots\quad\quad\vdots$ } & \multicolumn{1}{c|}{ $\vdots\quad\quad\vdots$ } & \multicolumn{1}{c|}{$\vdots\quad\quad\vdots$  }
&\multicolumn{1}{c|}{$\vdots\quad\quad\vdots$  }
& \multicolumn{1}{c}{ $\vdots\quad\quad\vdots$  } 
\\ 
Elect.\ curr.\ 49
& $-1.500\times 10^{0{~~}}$ &  $-1.501\times 10^{0{~~}}$ & $-1.496\times 10^{0{~~}}$
& $-1.497\times 10^{0{~~}}$
& $-1.500\times 10^{0{~~}}$ \\
Failure types & \multicolumn{1}{c|}{$a$} & \multicolumn{1}{c|}{$a$} & \multicolumn{1}{c|}{$a$} & \multicolumn{1}{c|}{$b$} &  \multicolumn{1}{c}{$b$}
\end{tabular}
\bigskip
\caption{Five entries from the UCI Drive data set. Last row indicates whether the failure is of type `$a$' or one of the other eleven types, all of which are indicated as `$b$'.}
\label{tbl:drive}
\end{table}

\begin{figure}[h]
\centering
\includegraphics[width=0.33\linewidth]{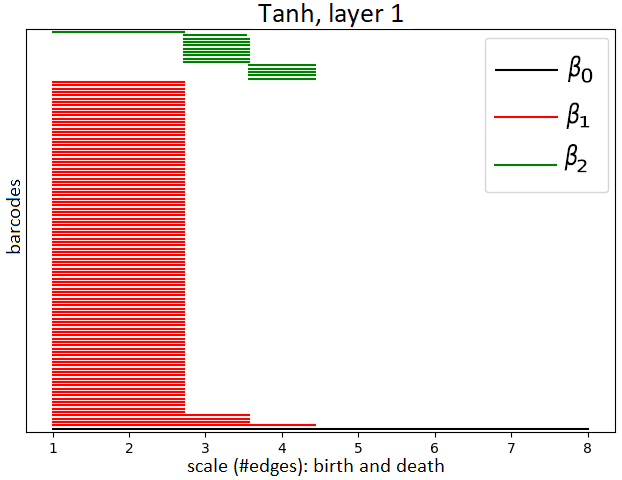}
\hspace{-1.0ex}
\includegraphics[width=0.33\linewidth]{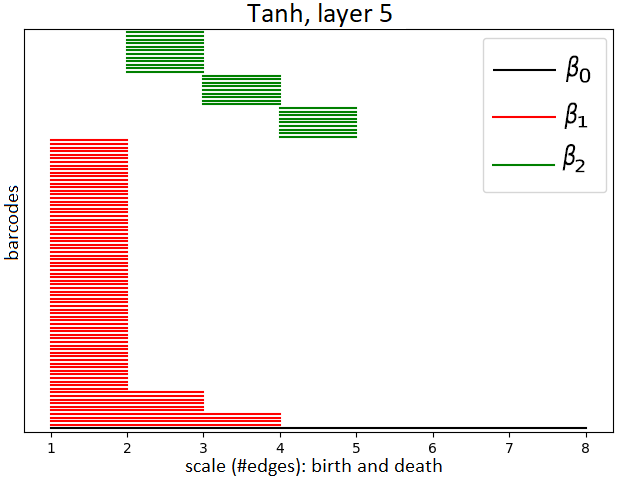}
\hspace{-1.0ex}
\includegraphics[width=0.33\linewidth]{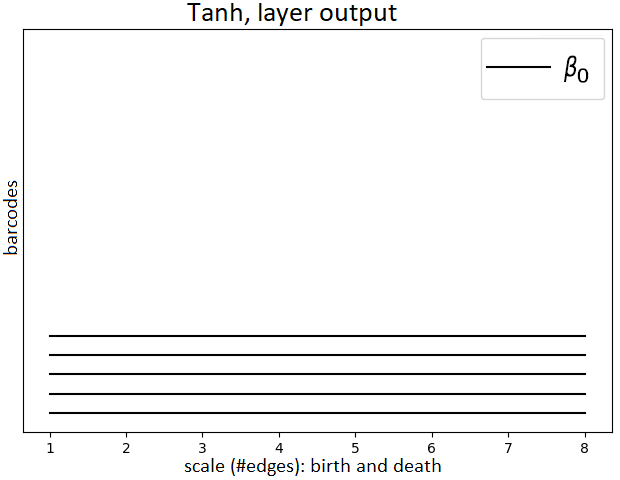}
\includegraphics[trim=200 150 150 170, clip, width=0.31\linewidth]{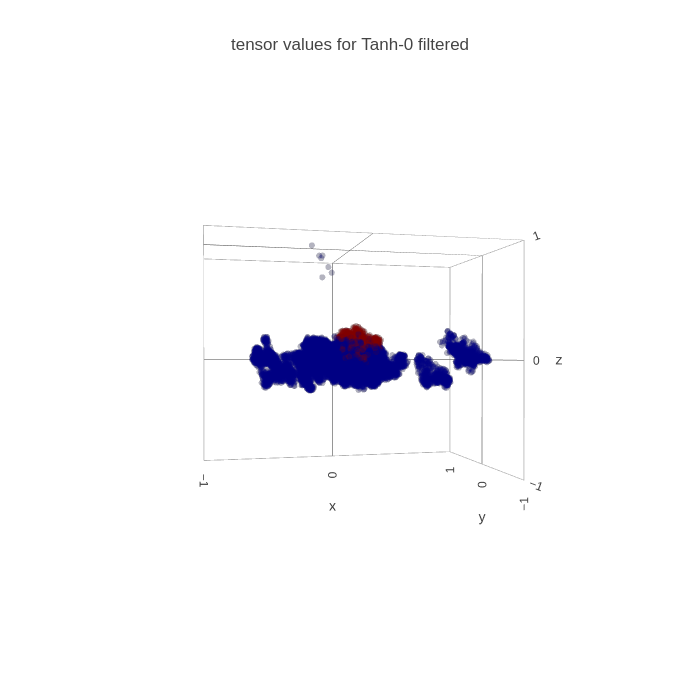}
%left bottom right top
\includegraphics[trim=150 100 150 180, clip, width=0.31\linewidth]{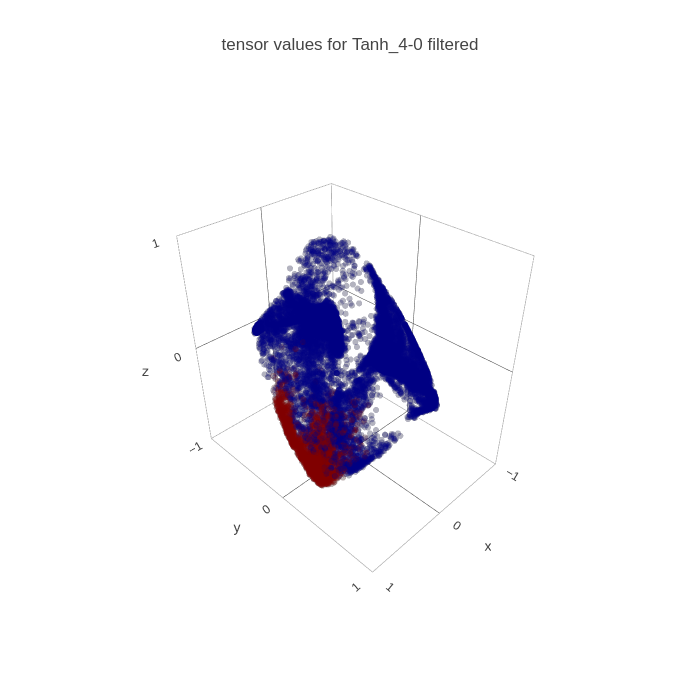}
\includegraphics[trim=150 100 150 190, clip, width=0.31\linewidth]{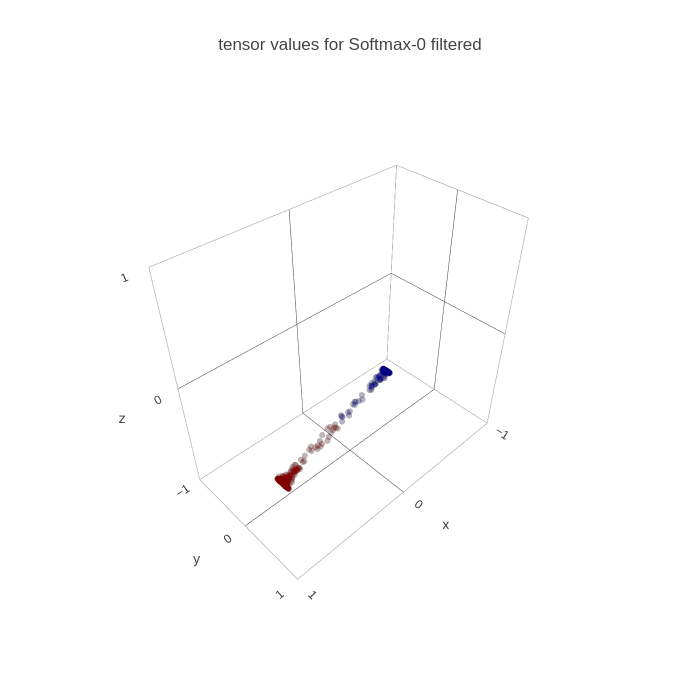}
\includegraphics[width=0.33\linewidth]{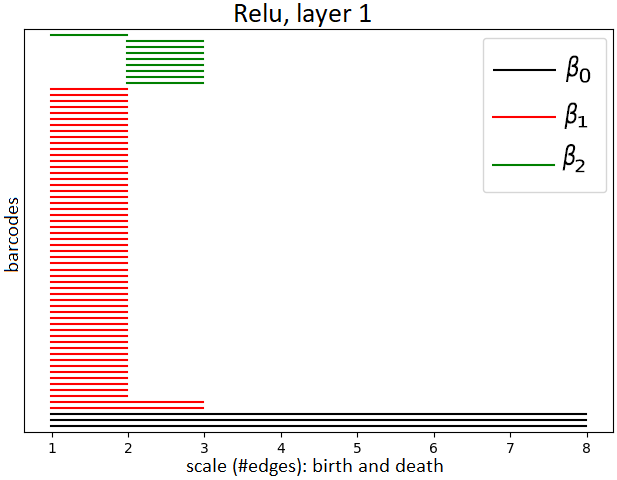}
\hspace{-1.0ex}
\includegraphics[width=0.33\linewidth]{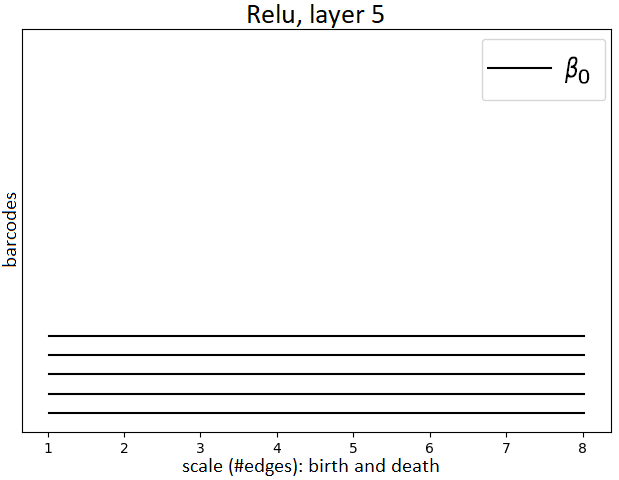}
\hspace{-1.0ex}
\includegraphics[width=0.33\linewidth]{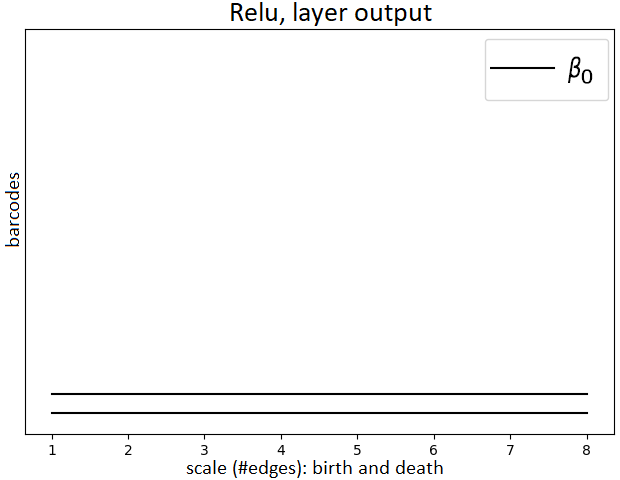}
\includegraphics[trim=150 110 160 220, clip, width=0.31\linewidth]{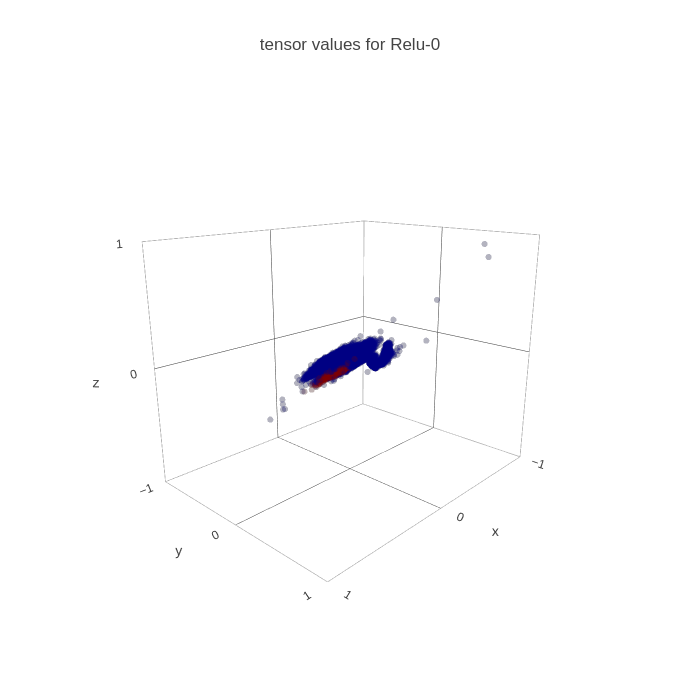}
%left bottom right top
\includegraphics[trim=160 130 200 150, clip, width=0.31\linewidth]{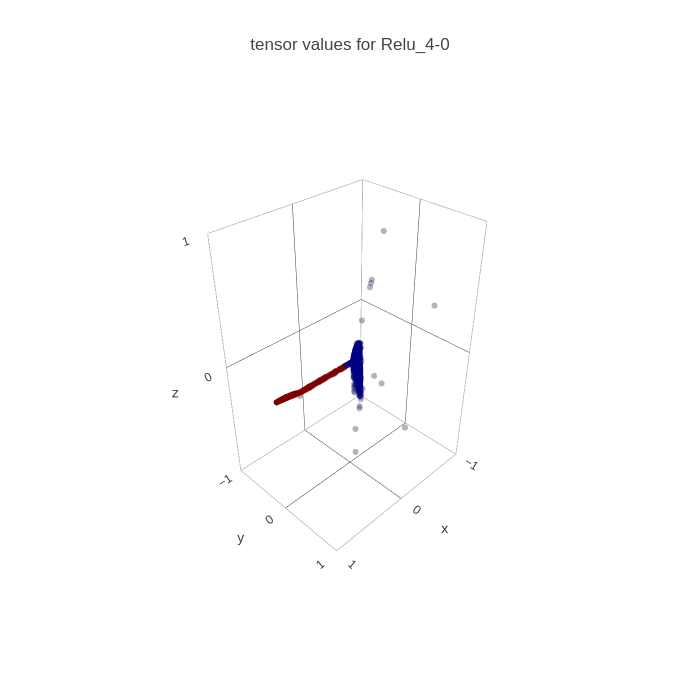}
\includegraphics[trim=140 120 200 160, clip, width=0.31\linewidth]{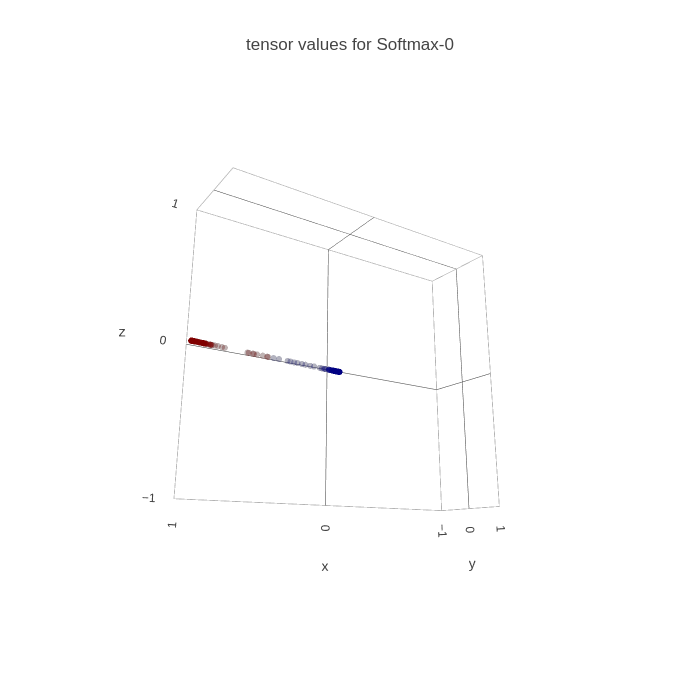}
\caption{Persistence barcode diagrams for neural networks trained on UCI Drive data show topology changes in the `manifold of type $a$ defects' $M_a$ as it passes through layers activated with $\tanh$ (\emph{previous page}) and ReLU (\emph{above}).  Scatter plots show principal component projections of $M_a$ (red) and the `manifolds of all other types of defects' $M_b$ (blue) at the corresponding layers.}
\label{fig:ph_drive_dataset}
\end{figure}

As in the case of the MNIST data set, instead of regarding the UCI Drive data as a classification problem with twelve classes, we reduce it to a binary classification problem of classifying a type $a$ defect versus all other eleven types of defects. So our manifold is $M =M_a \cup M_b$ where $M_a$ is  the `manifold of type $a$ defects' and $M_b$ is the union of the `manifolds of all other types of defects.'
Of the 58,509 entries in the UCI Drive data, we choose a random subset of 10,600 as our point cloud $X \subseteq M \subseteq \mathbb{R}^{49}$. The rest of the experiments is exactly as in the previous two cases (HTRU2 and UCI Banknotes data). The results are shown in Figure~\ref{fig:ph_drive_dataset} and they are fully consistent with the results in Figures~\ref{fig:ph_pulsar_dataset} and \ref{fig:ph_banknotes_autentication_dataset}, supporting the same conclusions we drew from all previously examined data sets.

\section{Concluding discussions}\label{sec:final_remarks}

Our findings support the view that deep neural networks operate by transforming topology, gradually simplifying topologically complicated data shapes and arrangements in the input space until it becomes linearly separable in the output space.  We proffered some new insights on the roles of the deep layers and of rectified activations, namely, that they are mechanisms that aid topological changes.
As this article is an empirical study intended to provide evidence for the above point-of-view, we did not investigate the actual mechanics of how a ReLU-activated neural network carries out topological changes. We conclude our article with a few speculative words about the `topology changing mechanism' of neural networks, mainly to serve as pointers for future work.

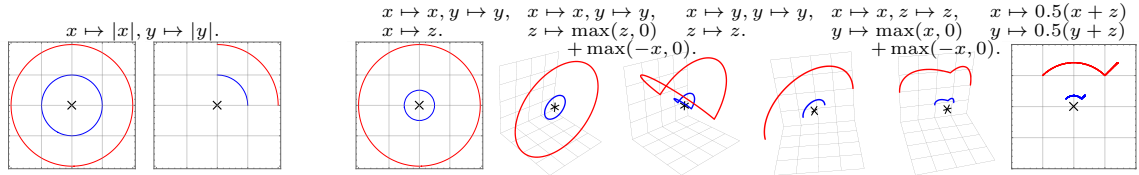
\begin{figure}[ht]
\hspace*{-5ex}
\begin{picture}(100,60)
\put(-5,0){\includegraphics[width=0.10955\linewidth]{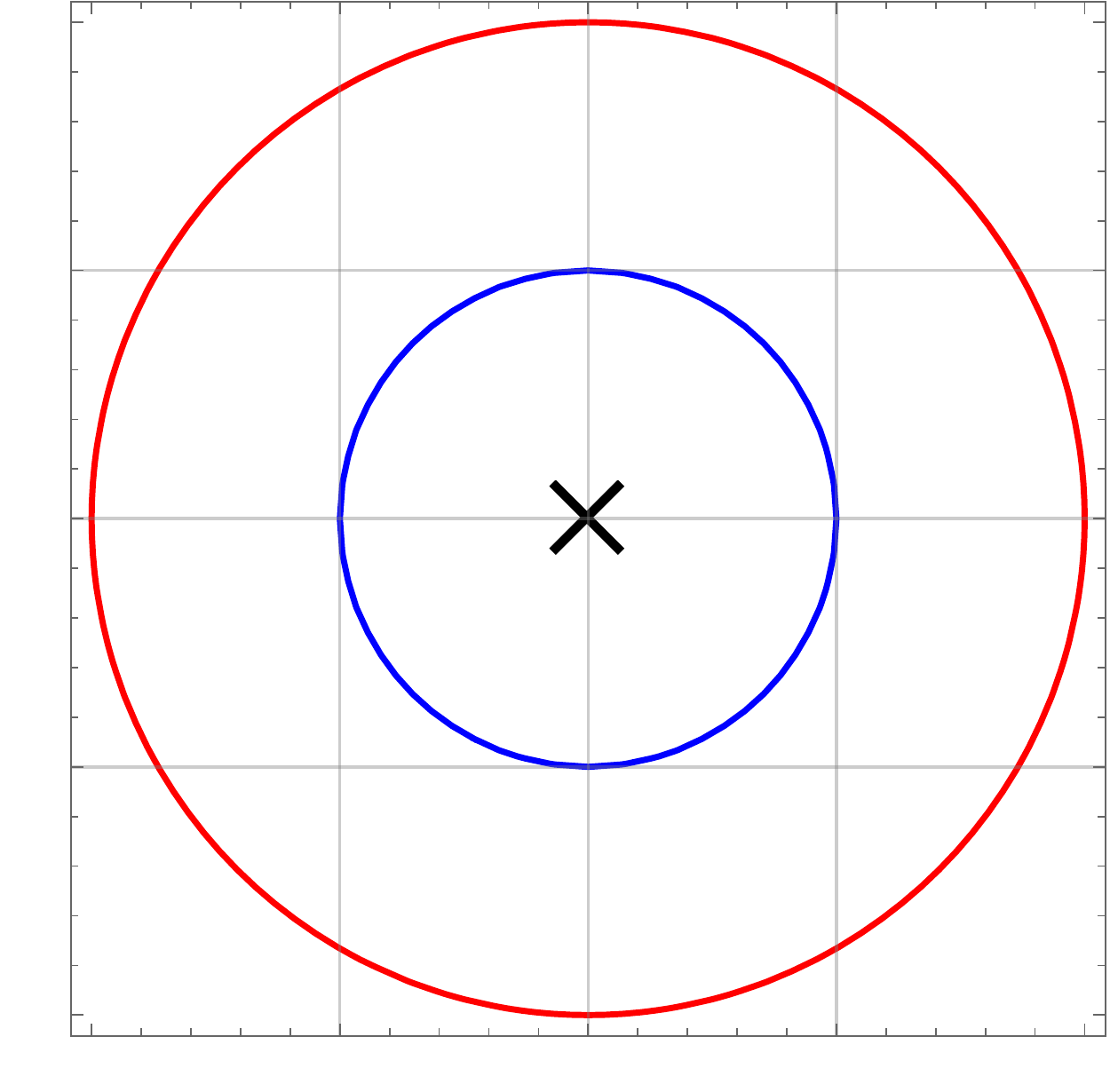}}
\put(50,0){\includegraphics[width=0.10955\linewidth]{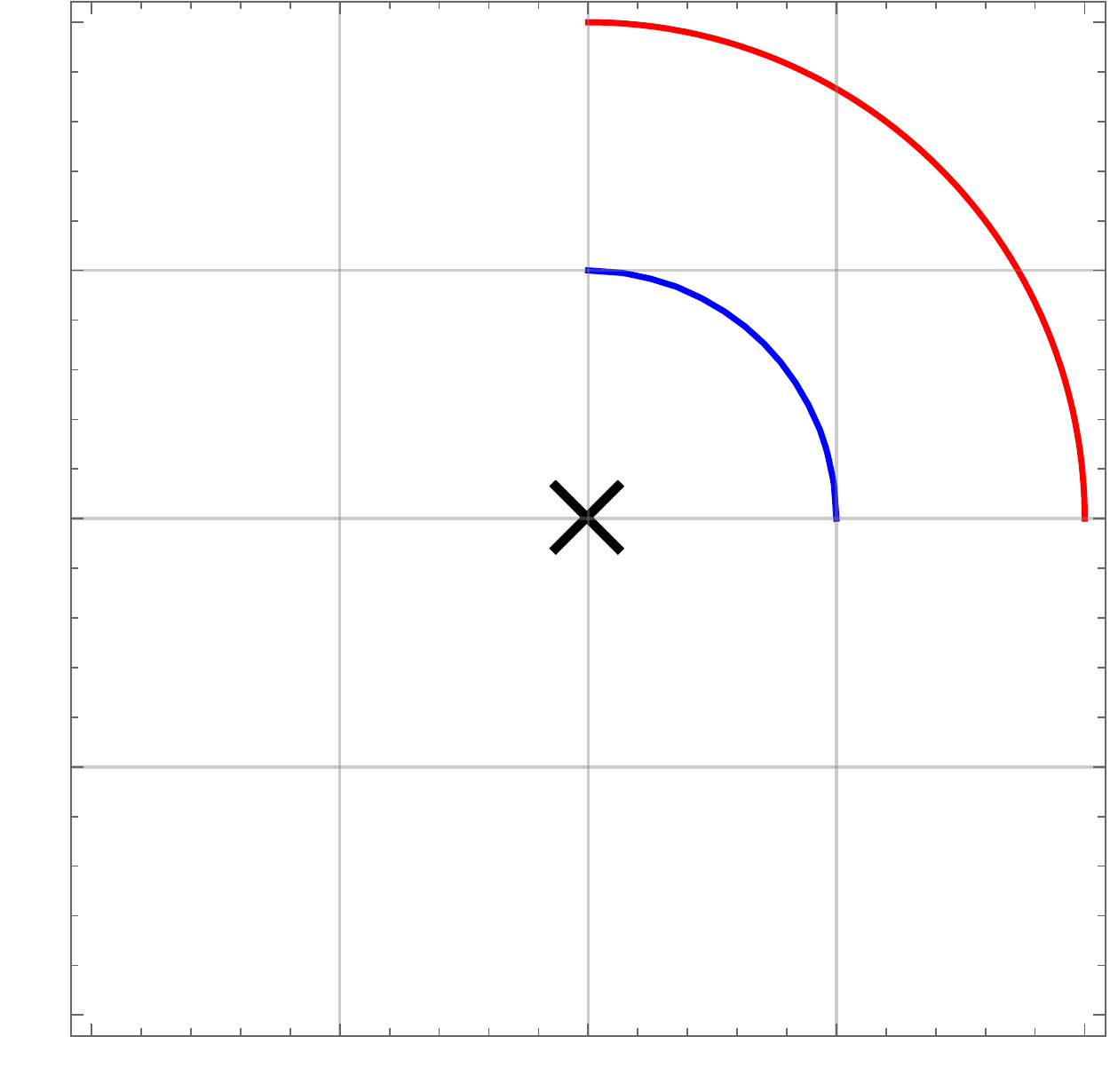}}
\put(20,53){\tiny $x \mapsto |x|, y \mapsto |y|.$}
\end{picture}      
\hspace*{10ex}
\begin{picture}(250,20)
\put(-28,0){\includegraphics[width=0.10955\linewidth]{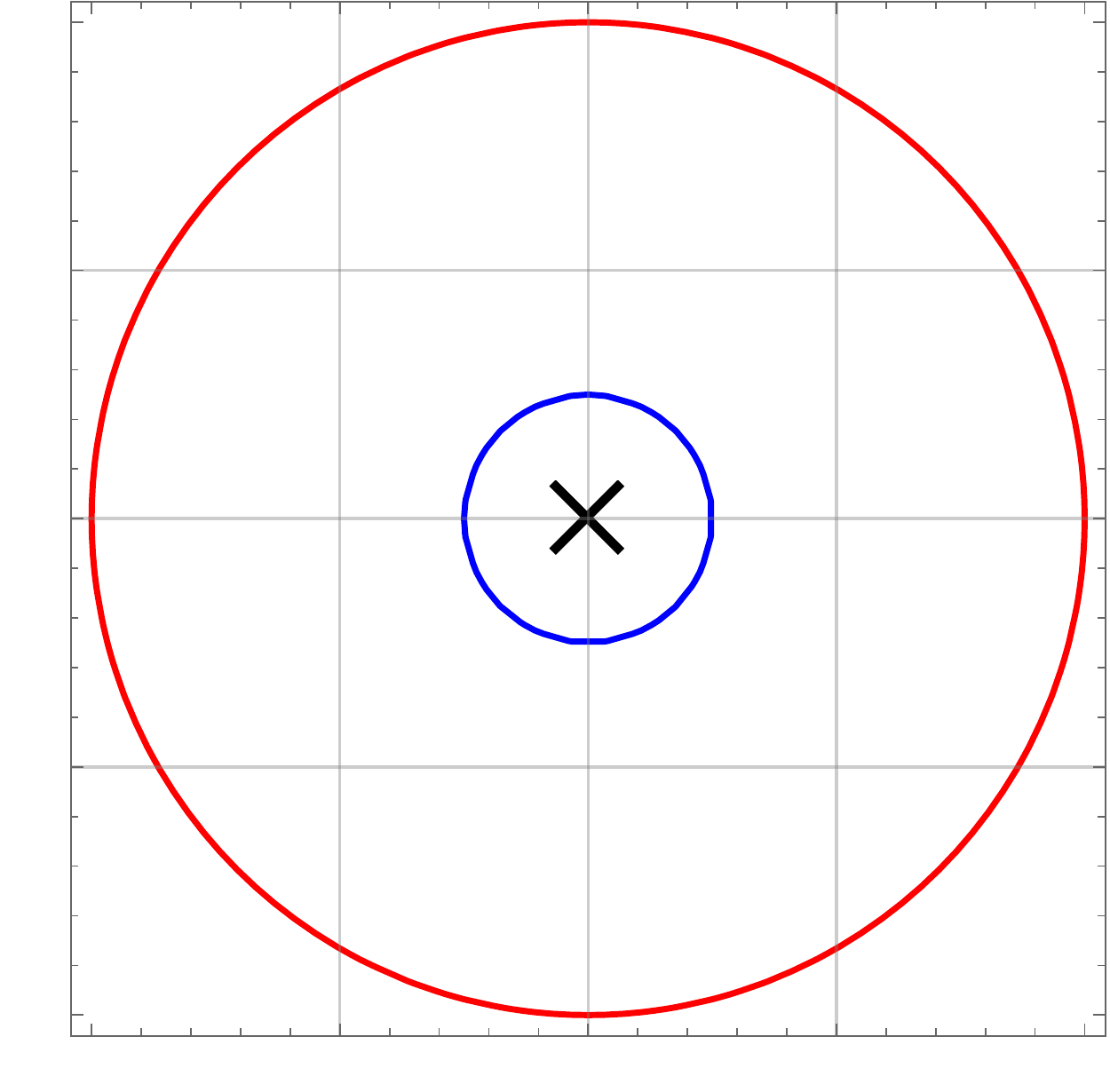}}
\put(25,0){\includegraphics[width=0.10\linewidth]{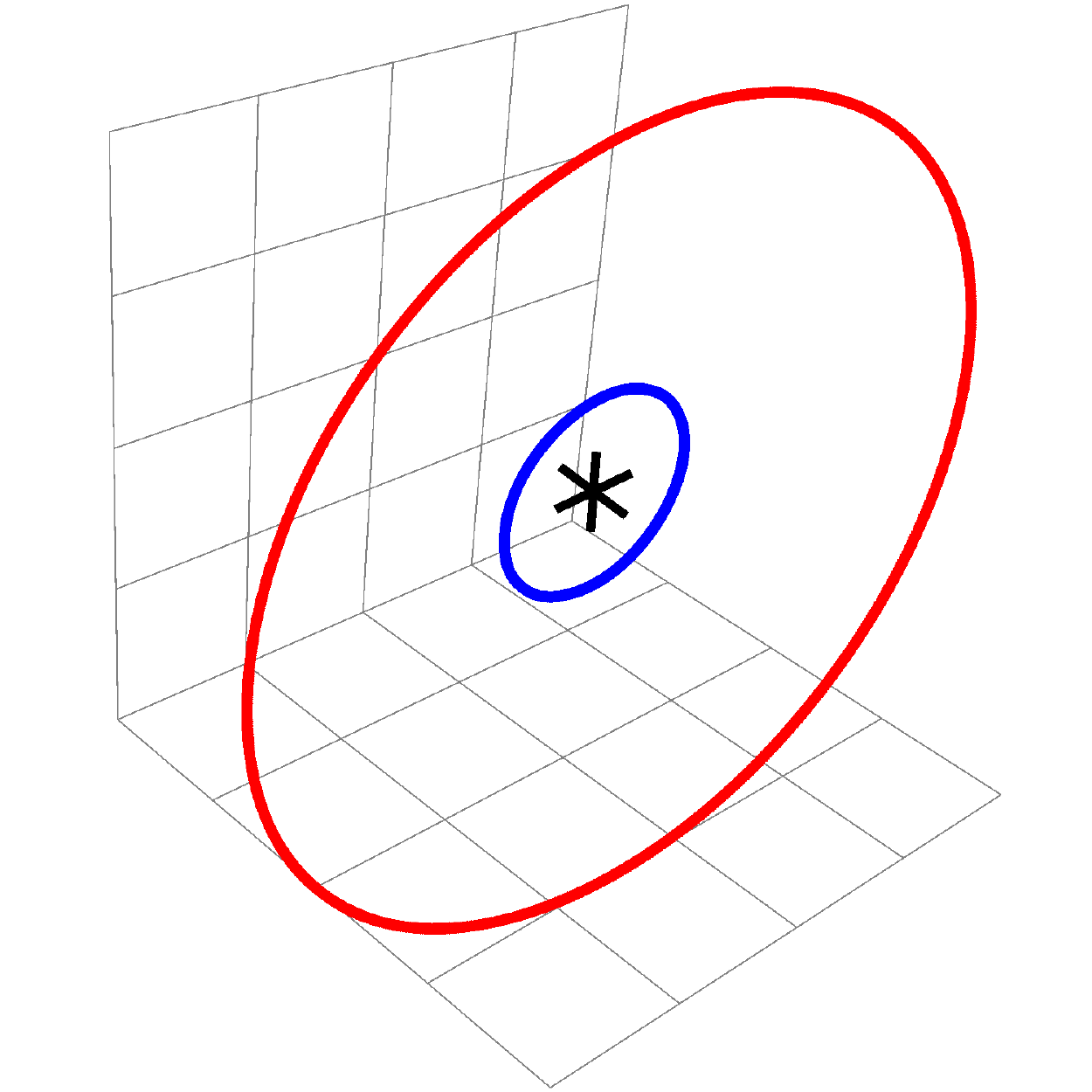}}
\put(75,0){\includegraphics[width=0.10\linewidth]{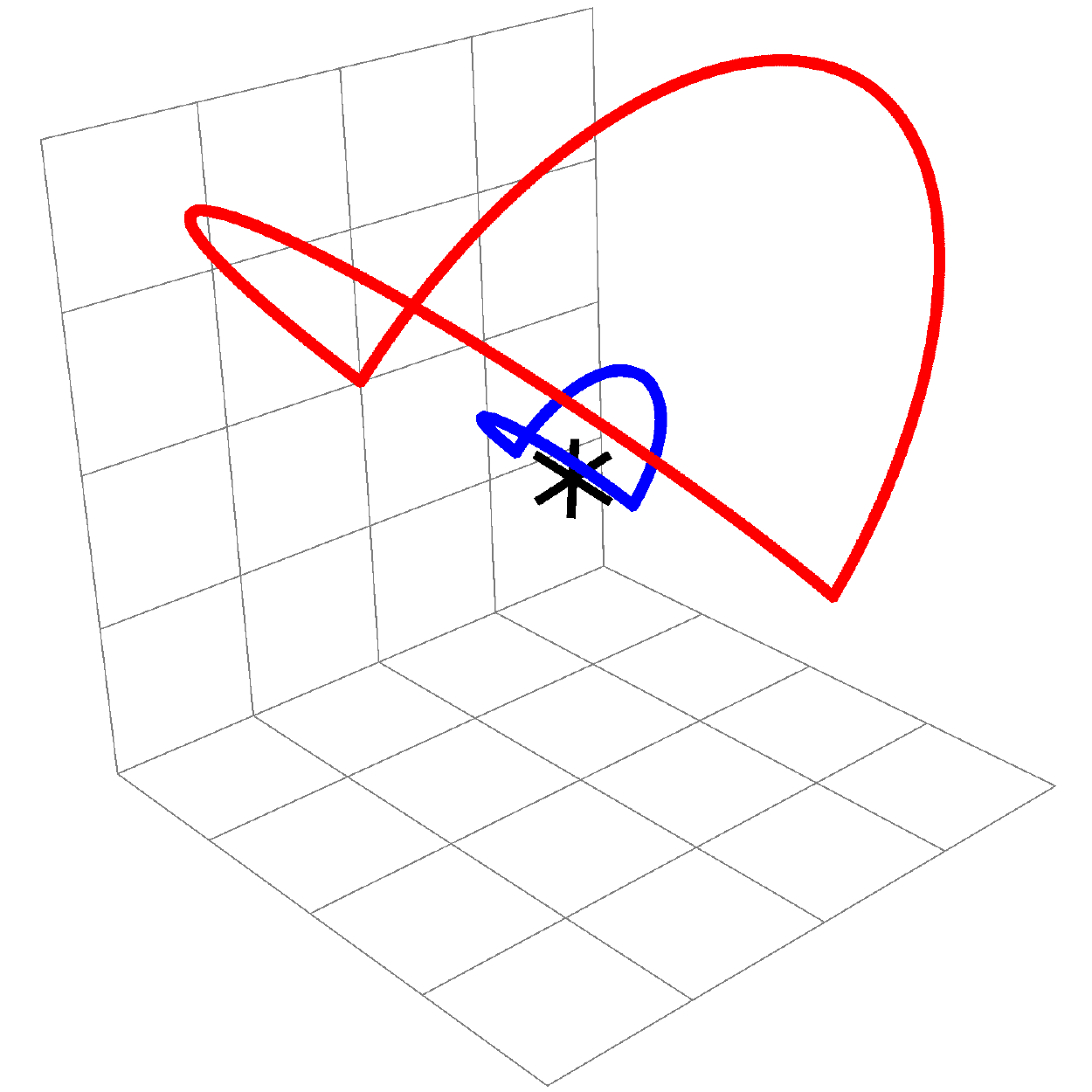}}
\put(125,0){\includegraphics[width=0.10\linewidth]{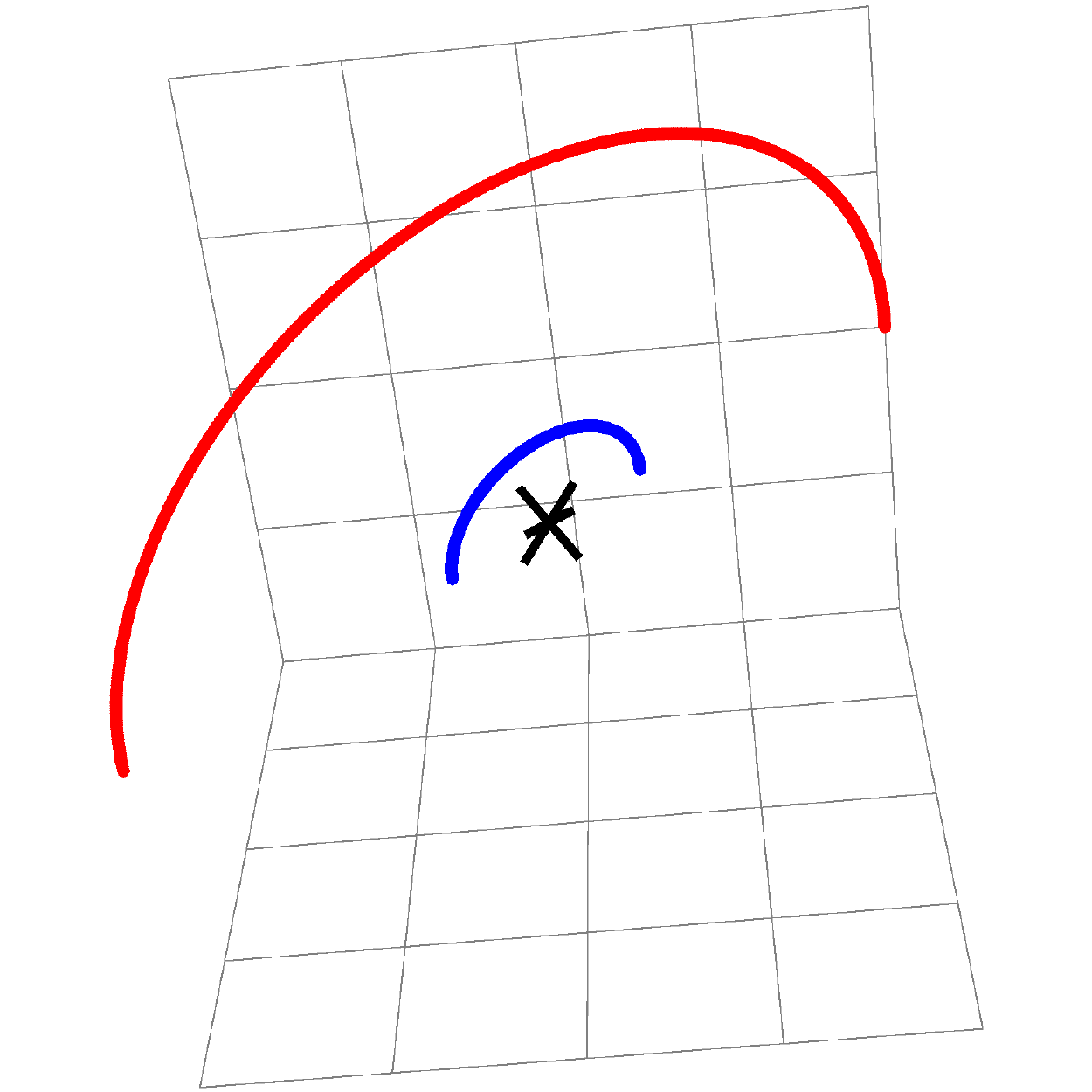}}
\put(175,0){\includegraphics[width=0.10\linewidth]{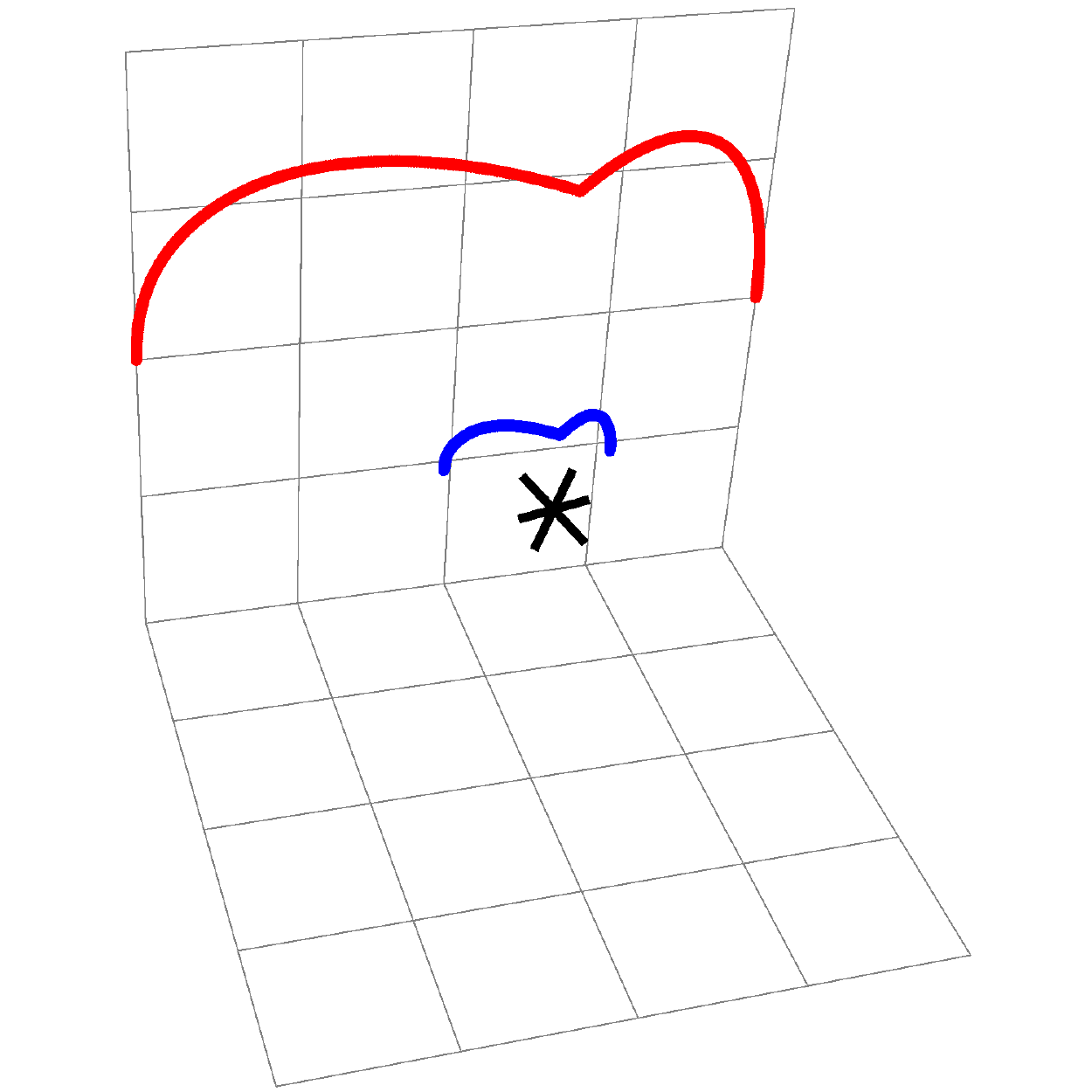}}

\put(220,0){\includegraphics[width=0.10955\linewidth]{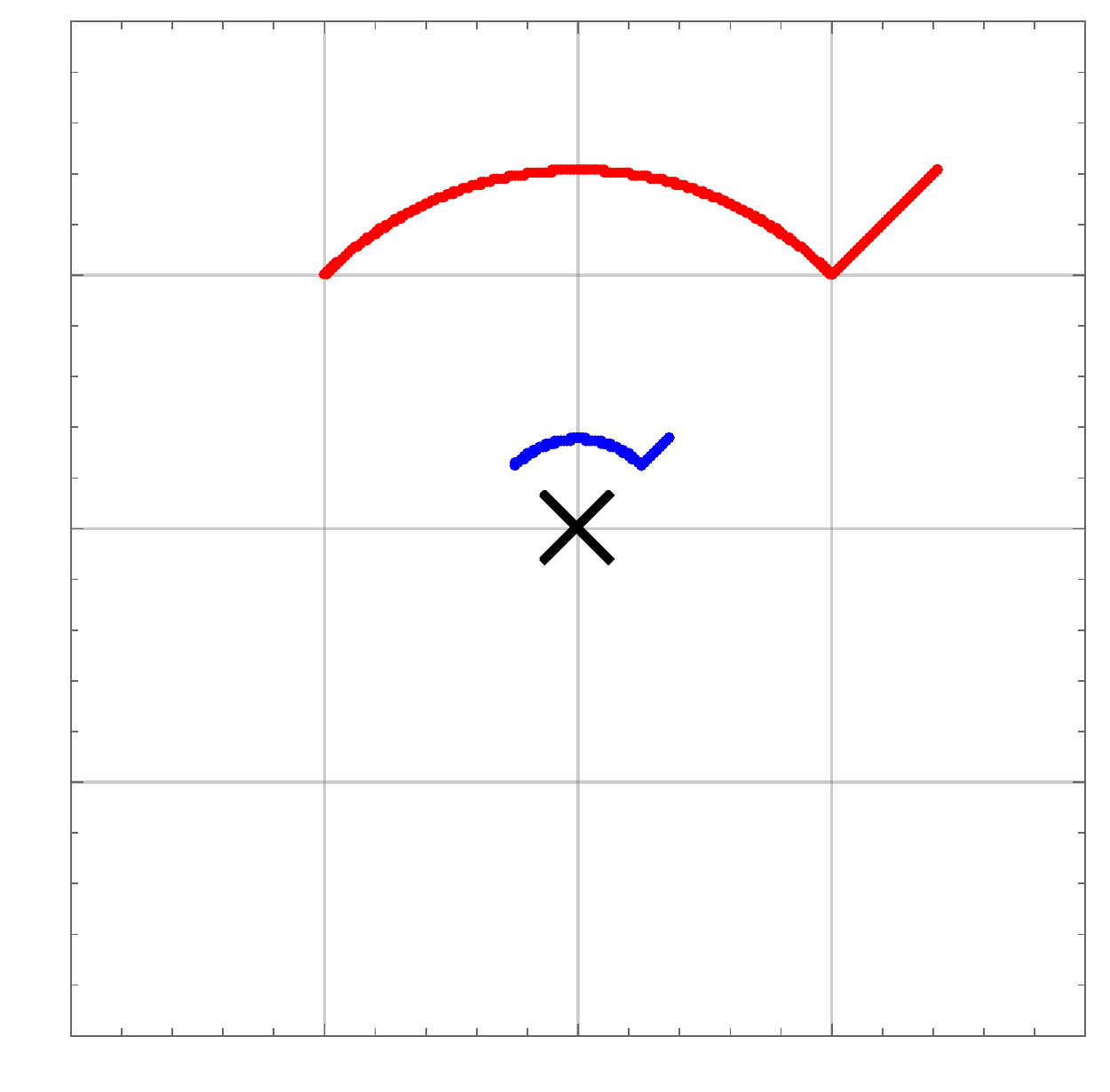}}
\put(-15,60){\tiny $x \mapsto x, y \mapsto y,$}
\put(-15,53){\tiny $x \mapsto z.$}

\put(40,53){\tiny $z \mapsto \max(z,0)$}
\put(55,46){\tiny $+\max(-x,0).$} 
\put(40,60){\tiny $x \mapsto x,y \mapsto y,$}  

\put(100,60){\tiny $x \mapsto y, y \mapsto y,$}
\put(100,53){\tiny $z \mapsto z.$}

\put(155,53){\tiny $y \mapsto \max(x,0)$}
\put(170,46){\tiny $+\max(-x,0).$} 
\put(155,60){\tiny $x \mapsto x, z\mapsto z,$}  

\put(215,53){\tiny $y \mapsto 0.5(y+z)$}
\put(215,60){\tiny $x \mapsto 0.5(x+z)$}  
\end{picture}  
\caption{\emph{Left}: Topology change with many-to-one maps: two neurons activated with the absolute value function can disentangle two concentric circles in a single step, transforming them into linearly separable sets. \emph{Right}: Topology change by embedding into higher dimensions and performing the disentangling operations therein.}
\label{fig:non_diffeo}
\end{figure}

Consider the concentric red and blue circles on the left of Figure~\ref{fig:non_diffeo}, two one-dimensional manifolds embedded in $\mathbb{R}^2$. By the Jordan Curve Theorem, there is no homeomorphism $\mathbb{R}^2 \to \mathbb{R}^2$ that will transform the two circles into two sets that can be separated by a hyperplane in $\mathbb{R}^2$. Nevertheless it is easy to achieve this with a \emph{many-to-one map} like $(x,y) \mapsto  (\lvert x\rvert,\lvert y\rvert)$ as shown in the left of Figure~\ref{fig:non_diffeo} that allows one to `fold' a set. An alternative way to achieve this is with an \emph{embedding into higher dimensional space} like in the right Figure~\ref{fig:non_diffeo} where we did  $\mathbb{R}^2 \to \mathbb{R}^3  \to \mathbb{R}^3 \to \mathbb{R}^3 \to \mathbb{R}^3\to \mathbb{R}^2$ with operations within $\mathbb{R}^3$ that disentangles the red and blue circles. 
Our speculation is that in a neural network, (i) the ReLU activation is a many-to-one map that can `fold' a space; (ii) the excess width\footnote{Width in excess of input dimension.} of the intermediate layers affords a higher dimensional \emph{space} in which to transform the data; (iii) the depth on the other hand plays the role of \emph{time}, every additional layer affords additional time to transform the data.  To elaborate on the last point (iii), note that since we are limited to affine transformations and ReLU activation, a substantial change to the topology of a space may require a longer sequence of these operations, and by `time', we simply mean the length of this sequence.

\begin{figure}[h]
\includegraphics[width=\linewidth]{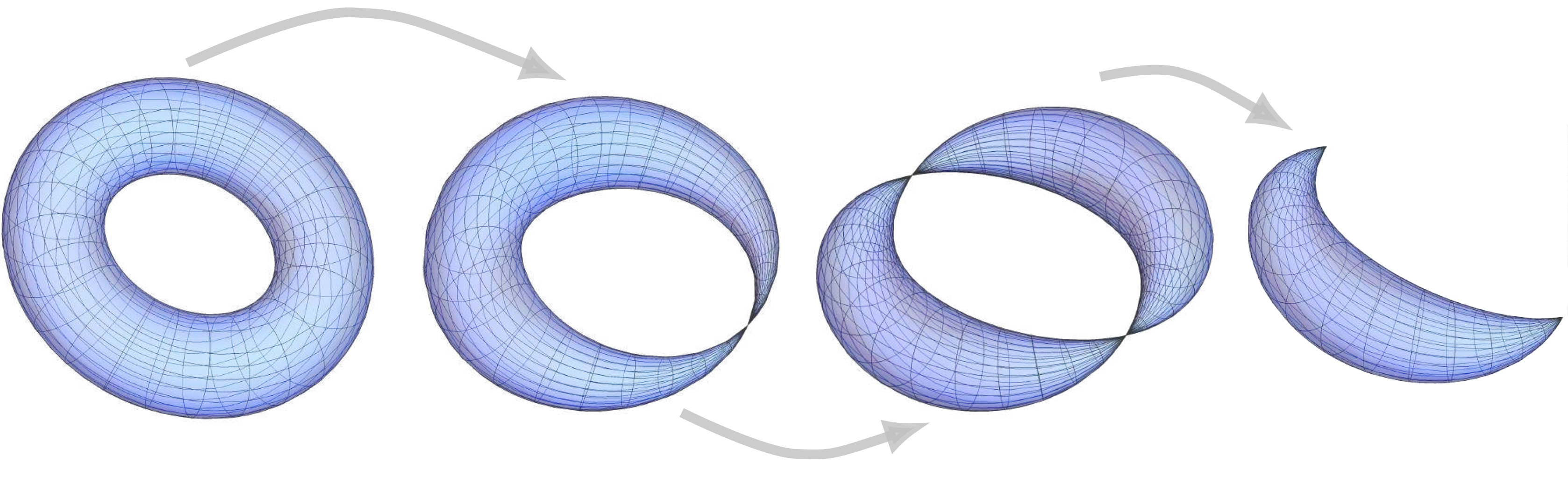}
\caption{Donut to croissant: torus $\to$ pinched torus $\to$ doubly-pinched torus $\to$ sphere.
Betti numbers: $(1,2,1) \to (1,1,1) \to (1,1,2) \to (1,0,1)$.}
\label{fig:torus}
\end{figure}

Adding to the first point (i), since an affine map takes the form $x \mapsto U \Sigma V x + b$, it provides the capability to translate (by the vector $b$), rotate/reflect (by the orthogonal matrices $U$ and $V$), and stretch/shrink (by the nonnegative diagonal matrix $\Sigma$);  ReLU-activation adds folding to the arsenal --- an important capability. For example,  to transform the surface of a donut (torus) into the surface of a croissant (sphere) as in Figure~\ref{fig:torus}, the first two operations may be achieved with appropriate affine maps but the last one requires that we fold the doubly-pinched torus into a croissant surface.

\subsection*{Acknowledgment} We thank David Bindel for very helpful discussions on floating point arithmetic. We would like to acknowledge support for this project from the National Science Foundation (NSF grant IIS 1546413), the Defense Advanced Research Projects Agency (DARPA D15AP00109 and a Director's Fellowship), and the University of Chicago (Chicago--Vienna Faculty Grant and Eckhardt Faculty Fund).

\end{document}